\documentclass[11pt,twocolumn]{article}
\usepackage[utf8]{inputenc}
\usepackage[T1]{fontenc}
\usepackage[margin=1in]{geometry}
\usepackage{mathtools}
\usepackage{amssymb}
\usepackage{graphicx}
\usepackage{array}
\usepackage{longtable}
\usepackage{booktabs}
\usepackage{float}
\usepackage{caption}
\usepackage{enumitem}
\usepackage{textcomp}
\usepackage[hidelinks]{hyperref}

\floatstyle{ruled}
\newfloat{algorithm}{tbp}{loa}
\floatname{algorithm}{Algorithm}

\newcommand{\keywords}[1]{%
  \par\medskip\noindent\textbf{Keywords:} #1\par\medskip
}
\newenvironment{nomenclature}{%
  \section*{Nomenclature}
  \begin{description}[leftmargin=2.3cm,labelwidth=2cm,labelsep=0.3cm,font=\normalfont,nosep]
}{\end{description}}
\newcommand{\entry}[2]{\item[#1] #2}

\newsavebox{\arxivtablebox}
\newenvironment{arxivtablefit}{%
  \begin{lrbox}{\arxivtablebox}
}{%
  \end{lrbox}%
  \ifdim\wd\arxivtablebox>\linewidth
    \resizebox{\linewidth}{!}{\usebox{\arxivtablebox}}%
  \else
    \usebox{\arxivtablebox}%
  \fi
}
\newenvironment{arxivlongtable}{%
  \fontsize{7.5}{9}\selectfont
  \setlength{\tabcolsep}{0.75pt}%
  \let\arxivunderscore\_
  \renewcommand{\_}{\arxivunderscore\allowbreak}%
}{}
\hypersetup{pdfauthor={Noah J. Bagazinski and Faez Ahmed},
  pdftitle={GenISys: Integrated Ship Systems Design Using Subsystem-Conditional Tabular Diffusion},
  pdfkeywords={Generative design, ship design, diffusion models, multidisciplinary design}}
\title{GenISys: Integrated Ship Systems Design Using Subsystem-Conditional Tabular Diffusion}
\author{%
  Noah J.~Bagazinski\thanks{Corresponding Author.}\\
  \small Department of Mechanical Engineering\\
  \small Massachusetts Institute of Technology\\
  \small Cambridge, MA 02139\\
  \small \texttt{noahbagz@mit.edu}
  \and
  Faez~Ahmed\\
  \small Department of Mechanical Engineering\\
  \small Massachusetts Institute of Technology\\
  \small Cambridge, MA 02139\\
  \small \texttt{faez@mit.edu}
}
\date{}
\begin{document}
\maketitle
\begin{abstract}
Ship design requires coordinating coupled subsystems to satisfy engineering constraints and balance competing performance and cost objectives. Generative artificial intelligence has supported component design, but integrated ship generation must also account for interactions among hull geometry, structure, propulsion, and operation. This paper presents GenISys, a framework for generating integrated systems, demonstrated through early-stage bulk-carrier design. Its dataset contains 217,999 proposed designs, including 16,247 that satisfy the formulated engineering constraints. The proposed simultaneous conditional denoising diffusion probabilistic model (SCDDPM) generates subsystems together, allowing each subsystem model to respond to changes elsewhere during generation. Experiments compare this approach with monolithic whole-ship generation and sequential subsystem generation across five independent training runs per architecture. In a study targeting lower cost per tonne of cargo transported, designs were generated for a required voyage range of 5,000 nautical miles. SCDDPM produced designs satisfying the formulated engineering constraints and meeting a common low-cost target in 55.70\% of generation requests, compared with 51.10\% for the monolithic model and 47.54\% for the sequential model. These correspond to 9.0\% and 17.2\% more qualifying designs, respectively, while maintaining similar estimated transport costs. These results show how generative artificial intelligence can coordinate subsystems to generate feasible, high-performing integrated systems.
\end{abstract}

\begin{figure*}[t]
\centering\includegraphics[width=6.0in]{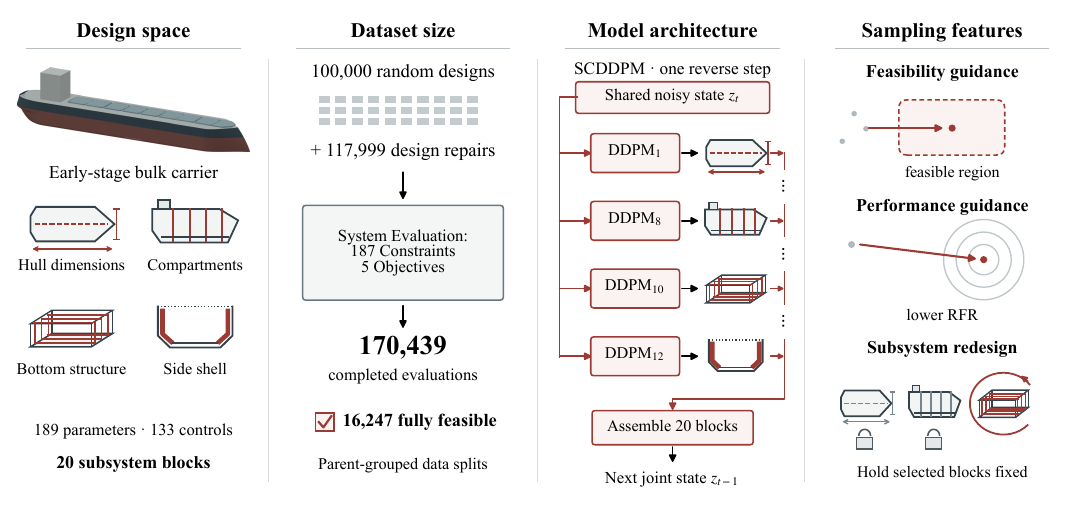}
\caption[Overview of the GenISys project]{Overview of the GenISys project: integrated design framework, dataset construction, model architecture, and sampling capabilities. Random and one-pass repair proposals share one evaluator for rule-based engineering constraints and performance. Each subsystem model updates its part of the same intermediate ship design, and the updates are combined at each generation step. Guidance directs generation toward constraint satisfaction or improved performance. During redesign, selected subsystems can change while the remaining independent design parameters stay fixed. Every proposed design is evaluated. Illustrations are schematic; Figure~\ref{fig_2} details the information flow.}
\label{fig_1}
\end{figure*}

\section{Introduction}
Ship design brings together many engineering decisions to create a vessel that can carry out its intended mission. For a bulk carrier, hull form, structure, machinery, fuel capacity, and cargo arrangement must work together to transport cargo over a specified voyage range. These decisions are closely coupled. Increasing hull size can provide more displacement and cargo space, but can also increase structural mass, resistance, and powering requirements. The resulting balance determines how much cargo the ship can carry and the cost of transporting it. Ship design therefore requires coordinated assessment of the complete vessel as its constituent subsystems are developed \cite{papanikolaou2010holistic}.

Ship-design processes manage these interactions through repeated evaluation and revision. Evans' design spiral describes a progression through principal dimensions, hull form, arrangement, powering, weight, stability, and cost, with earlier decisions revisited as the design develops \cite{evans1959basic}. Set-based design supports this process by retaining alternatives while their tradeoffs are examined, rather than committing immediately to one configuration \cite{singer2009set}. Page similarly advocates continuous concept development to explore alternatives and support later design decisions \cite{page2022case}. Computational optimization can help coordinate these choices, but evaluating coupled engineering responses requires computational effort for each candidate \cite{hart2010IMDO}. This motivates methods that help designers explore promising alternatives while retaining the ability to revise subsystem decisions.

Artificial intelligence (AI) offers ways to learn design representations and propose alternatives for engineering assessment. Generative methods such as PaDGAN and TopoDiff demonstrate how design generation can account for performance objectives \cite{chen2021padgan,maze2023topodiff}. In ship design, Khan et al.'s ShipHullGAN generates hull forms across multiple ship types \cite{khan2023shiphullgan}, while their shape-supervised dimension reduction method constructs compact design spaces using geometric information associated with shape and physical behavior \cite{khan2022shapesupervised}. ShipGen and C-ShipGen further demonstrate diffusion-based hull generation with engineering constraints and performance guidance \cite{bagazinski2023shipgen,bagazinski2024cshipgen}. Together, these studies establish opportunities to support design exploration through learned representations and guided generation.

Extending these capabilities to integrated ship design requires coordinating decisions across subsystems. A generated hull must accommodate structure, machinery, fuel, and cargo, and the assembled design must satisfy their combined engineering requirements. Its economic performance also depends on these interactions: a reduction in resistance may offer little transport benefit if the accompanying design changes substantially reduce cargo capacity. The opportunity addressed here is to generate coupled subsystem designs, guide them toward engineering and economic objectives, and redesign selected subsystems while preserving other design decisions. This requires both a shared representation of the ship and an evaluator that connects subsystem choices to whole-vessel outcomes.

This paper introduces GenISys, a framework for generating integrated systems, demonstrated through preliminary bulk-carrier design. It combines a parametric ship representation, engineering and economic evaluation, and a synthetic dataset for learning and comparing generation methods. Figure~\ref{fig_1} presents the framework. Within GenISys, the proposed simultaneous conditional denoising diffusion probabilistic model (SCDDPM) generates subsystem designs together, allowing each subsystem model to respond to decisions elsewhere throughout generation. Subsystem-level feasibility guidance directs changes to an individual subsystem's design parameters using the engineering constraints assigned to that subsystem. The design studies compare SCDDPM with two alternatives. The whole-ship denoising diffusion probabilistic model (DDPM) generates the independent ship-design parameters using a single model. The hierarchical conditional denoising diffusion probabilistic model (HCDDPM) generates subsystems one at a time in a prescribed sequence. The studies examine engineering feasibility, guidance toward lower transport cost, and selective subsystem redesign. These comparisons assess complete generation pipelines, whose model sizes, training effort, and guidance configurations differ.

The contributions are:
\begin{enumerate}
\interlinepenalty=10000
\item An integrated bulk-carrier design framework and synthetic dataset covering 189 design parameters, including 133 generator-controlled parameters grouped into 20 subsystems, 187 rule-based engineering constraints, and five performance objectives. The dataset includes 16,247 designs that satisfy all formulated engineering constraints.
\item Proposed SCDDPM architecture that enables simultaneous subsystem generation, subsystem-level feasibility guidance, and redesign of selected subsystems while preserving the remaining independent design parameters, without requiring a prescribed subsystem generation sequence.
\item For bulk carriers with a required voyage range of 5,000 nautical miles, SCDDPM produced more designs that satisfied the formulated engineering constraints and met a common low-cost target. For the same number of design requests, it produced 9.0\% more such designs than DDPM and 17.2\% more than HCDDPM, while maintaining similar estimated costs per tonne of cargo transported.
\end{enumerate}

\section{Related Work}
This section describes prior research that informed the work done in this paper. The subsections detail generative artificial intelligence for engineering design, ship design with artificial intelligence, and systems design with generative artificial intelligence. Comparisons of related work to this work are provided throughout this section. 

\subsection{Generative Artificial Intelligence for Engineering Design}
Generative engineering design seeks feasibility, performance, and useful variation beyond statistical similarity \cite{regenwetter2023beyond}. Performance-augmented diverse generative adversarial networks (PaDGAN) promote high-quality diverse designs, while diffusion models guide generation using learned objectives or constraints \cite{chen2021padgan,maze2023topodiff,arechiga2023drag}. Diffusion models can guide generated designs toward specified performance goals. Existing approaches use desired objective values, examples of iterative optimization, or preference and reward models to influence generation. These approaches support design searches using existing datasets, engineering constraints, and multiple competing objectives \cite{krishnamoorthy2023ddom,giannone2023trajectory,annadani2025preference,keramati2026reward}.

Regenwetter et al. use negative data to reduce constraint-violating outputs, including hulls subject to hydrodynamic constraints \cite{regenwetter2024negative}. This motivates retaining infeasible examples and their labels. GenISys supplies such records for coupled ship systems alongside its fully feasible training subset.

\subsection{Ship Design With Artificial Intelligence}
Evans' design spiral describes iterative coordination among ship-design disciplines \cite{evans1959basic}. Computer-aided design (CAD) and holistic design connect geometry, disciplinary analyses, and decision criteria \cite{nowacki2010five,papanikolaou2010holistic,papanikolaou2022holistic}. Design Building Blocks and arrangement methods address spatial synthesis \cite{andrews2006simulation,ShipArrangements}, including generative subdivision proposals \cite{deleon2025subdivision}. Multidisciplinary and robust optimization coordinate coupled decisions \cite{peri2001design,brown2003multiple,hart2010IMDO,diez2010robust}, with information exchange formalized through multidisciplinary design architectures and tools \cite{martins2013mdo,gray2019openmdao}.

AI methods address resistance, structure, arrangement, and hull synthesis \cite{abramowski2013application,ao2023artificial,ao2024integrated,htein2025ai,kim2025review}. ShipHullGAN, ShipGen, and C-ShipGen provide parametric hull-generation precedents \cite{khan2023shiphullgan,bagazinski2023shipgen,bagazinski2024cshipgen}. Structural constraint handling has also benefited from adaptive repair: Samanipour and Jelovica identify variables responsible for violated constraints; Cai and Jelovica automate variable--constraint mapping; and Jelovica and Cai extend adaptive repair within structural optimization \cite{samanipour2020adaptive,cai2021adaptive,jelovica2024improved}. Later neural mapping supports autonomous repair, including a chemical-tanker case \cite{cai2023neuralrepair}. MiDShip supplies a representative three-hold structural dataset and an equation-informed repair approach that GenISys extends to the coupled ship-system evaluator \cite{bagazinski2026midship}.

These hull, structural, and arrangement capabilities motivate a common evaluated representation connecting subsystem decisions to cargo allocation, loading, powering, and economics.

\subsection{Systems Design With Generative Artificial Intelligence}
System-level generation has precedents beyond diffusion. DeLPhy generates varied system designs to explore trade-offs between performance goals, with demonstrations on a Formula SAE racing vehicle and propellers. It learns a compact representation of example designs and a model that predicts their performance. A search method called partial simulated annealing then explores this representation to improve selected objectives while keeping others near specified targets \cite{cobb2022diverse}. Chen and Fuge condition separate part generators on their parents \cite{chen2019hierarchical}. In this paper, the hierarchical conditional denoising diffusion probabilistic model (HCDDPM) adapts this ordered conditioning to diffusion. Zhou et al. encode assembly graphs to complete bicycle design parameters \cite{zhou2024bridging}, establishing engineering partial-design completion as a prior capability.

Other methods coordinate generators or constraint models. Diffusion-based continuous constraint solvers combine constraint-specific energies \cite{yang2023diffusionccsp}; compositional inverse design coordinates generative representations \cite{wu2024compositional}, and generative models address mechanical component interfaces \cite{etesam2025gearformer}. Ghiglino et al. generate conceptual aircraft with separate topology and masked-parameter diffusion models and simulation-based evaluation \cite{ghiglino2026planes}. Whole-system generation and conditional completion therefore already extend beyond ship hulls.

CoDi uses one diffusion model for continuous values and another for categorical choices. At each generation step, each model uses the other's noisy intermediate values as context. Additional training compares matched and mismatched combinations to help preserve relationships between the two variable groups \cite{lee2023codi}. SCDDPM instead groups parameters by engineering subsystem and applies Gaussian diffusion and ordinary denoising losses, adding interfaces for subsystem-level feasibility guidance and redesign. Other approaches compose scores or energies \cite{liu2022composable,du2023reducereuse}, learn the influence of pretrained conditional denoisers \cite{huang2023collaborative}, or reconcile regional denoising paths \cite{bartal2023multidiffusion}. These coordinate overlapping predictions; SCDDPM assembles disjoint subsystem outputs.

GLUE coordinates frozen pretrained subsystem generators through a system-level latent representation, with data-driven and data-free training variants \cite{aebersold2025glue}. GLUE emphasizes generator reuse; GenISys learns coupled design distributions from repair-derived data. SCDDPM learns relationships between subsystems from ship designs evaluated as complete systems. These learned relationships support subsystem generation and subsystem-level feasibility guidance. Generating compatible subsystem combinations depends on how well those relationships are represented in the training data.

Guidance and locking also have precedents. Sung et al. combine feasibility and cooling-performance guidance for battery arrangements \cite{sung2024cooling}. RePaint provides a precedent for replacing known image regions during reverse diffusion \cite{lugmayr2022repaint}. In GenISys, locked design coordinates are replaced at each reverse step using a forward-noised reference trajectory. Their prescribed physical values are restored exactly before evaluation. The sampler uses coordinate replacement without RePaint's resampling steps.

GenISys evaluates subsystem parameterization, information exchange, and guidance interfaces within a coupled ship representation linking geometry, structure, loading, and economics.

\section{Design Representation and Evaluation}
This section describes the design variables, subsystem partition, engineering evaluations, and constraint-based feasibility labels.

GenISys combines the parametric hull representations and hull-generation work from Ship-D, ShipGen, and C-ShipGen \cite{bagazinski2023shipD,bagazinski2023shipgen,bagazinski2024cshipgen} with the representative three-hold structural model and equation-informed augmentation workflow from MiDShip \cite{bagazinski2026midship}. It adds whole-vessel subdivision, component mass budgets, powering and fuel estimates, cargo allocation, departure loading, and annual transport economics. Their integration couples displacement and component masses to payload, hold volumes and locations to allocation and trim, loading to structural demand, and annual cost and cargo throughput to required freight rate (RFR), the modeled cost per tonne of cargo transported. The evaluator defines a coupled concept-design benchmark for comparing the three diffusion architectures. Reported feasibility and performance apply within its published engineering formulations and stated project assumptions, detailed in Appendix~\ref{app_C}.

\subsection{Mixed Design Variables and Subsystem Partition}
\label{sec:design-parameters}
The generator selects 133 independent ship-design parameters, including hull proportions, structural dimensions, and the number of cargo holds. These comprise 84 continuous parameters, 33 whole-number parameters, and 16 two-option controls. The two-option controls include 13 choices between T-shaped and L-shaped stiffener sections and three hull configuration switches. The complete ship representation contains 189 parameters. The other 56 values are supplied through fixed settings, reconstruction rules, and the cargo-loading solver; this group also includes unused fields and alternate names for the same quantity. The generated parameters are grouped into 20 nonoverlapping subsystems (Table~\ref{tab_A1} in Appendix~\ref{app_A}). Appendix~\ref{sec_D1} identifies every parameter's role and reconstruction; Appendix~\ref{app_D} provides the complete parameter and constraint tables.

The hull variables define length overall (LOA) and nondimensional bow, stern, beam, depth, bilge, keel, and bulb geometry. Operating variables specify draft fraction, Froude number, and voyage range. Subdivision variables specify end-region fractions and the number of cargo holds. Structural variables describe plate thicknesses, member counts and sections, brackets, hoppers, and representative hatch-opening dimensions. The vessel family is restricted to steel bulk carriers with an inner-side shell and without distinct cargo-region longitudinal bulkheads.

The intervals used for random sampling specify where initial candidate parameter values were drawn. They do not determine whether a design is feasible. Feasibility is determined by evaluating the complete design against the formulated constraints, and repair can produce feasible designs outside the initial sampling intervals. Appendix~\ref{sec_D1} reports the observed parameter ranges in the full dataset and its fully feasible subset.

Equation~\eqref{eq_1} couples several dimensions before evaluation. Here $s_{BH}$ is cargo-region bulkhead spacing, $N_H$ is the number of holds, and $f_q$ are the fractions allocated to noncargo end regions:
\begin{equation}
\resizebox{0.90\linewidth}{!}{$\displaystyle
 L_{\mathrm{OA}}=\frac{s_{BH}N_H}{1-\sum_{q\in\mathcal E}f_q},\qquad
 T=f_TD,\qquad U=Fn\sqrt{gL_{WL}}.
 \label{eq_1}
$}
\end{equation}
Random dataset construction starts from bulkhead spacing and hold count and calculates ship length. During sampling, the generators instead predict ship length and hull proportions. Other dimensions, including beam, depth, and the length of the representative three-hold structural model, are calculated from the predicted hull and subdivision parameters, as detailed in Appendix~\ref{sec_D1}. The predicted ship length is retained rather than recalculated from structural spacing. Cargo fractions follow the departure loading solution. Equation~\eqref{eq_1} expresses geometric consistency, not an instruction to overwrite learned LOA.

\subsection{Multidisciplinary Performance Evaluation}
The evaluator constructs the hull and hydrostatic grid, determines draft and service speed, predicts resistance and installed power, estimates voyage fuel, and allocates cargo and fuel spaces. Structural, outfit, machinery, fuel, and miscellaneous deadweight estimates then determine the available cargo mass. Cargo is distributed among the holds while respecting hold-volume limits and the total cargo mass. The allocation seeks to minimize the difference between forward and aft drafts. The departure trim check requires this difference to be no greater than 1\% of the equilibrium draft: $|T_a-T_f|\leq0.01T_{eq}$, where $T_a$ and $T_f$ are the aft and forward drafts, and $T_{eq}$ is the equilibrium draft calculated by the loading solver. The loading solution supplies centers of gravity, hydrostatics, and longitudinal weight distributions. Appendix~\ref{sec_C3} details the component-mass budget, cargo-allocation outputs, and loading conditions.

Composite trapezoidal integration gives displacement, moments, and still-water loads. Equation~\eqref{eq_2} defines shear and bending moment from the net upward distributed load $q(x)=b(x)-w(x)$:
\begin{equation}
 V(x)=\int_0^x q(s)\,ds,\qquad
 M(x)=\int_0^x V(s)\,ds.
 \label{eq_2}
\end{equation}
Here $x$ is longitudinal position in metres from the first integration station; buoyancy per unit length $b$, downward weight per unit length $w$, and net upward load per unit length $q$ are in N/m. Shear $V$ and bending moment $M$ are in N and N\,m, respectively.
Numerical consistency checks, including a three-design grid-refinement test, are reported in Appendix~\ref{sec_C7}.

For full-cargo departure, still-water loads are combined with rule-based wave loads for hogging and sagging---upward and downward bending of the hull. The three-hold structural model provides a representative cross-section through the largest modeled hatch opening. Gross deck and bottom section moduli are calculated from the modeled plate and member dimensions, with the hatch opening deducted. These quantities describe the section's resistance to longitudinal bending. The same opening dimensions are copied to each modeled hold to estimate preliminary cover mass and center of gravity. This does not define a designed per-hold hatch layout or engineered hatch covers; the structural representation also does not provide full-length stress fields or finite-element analysis.

Resistance and powering use the empirical Holtrop--Mennen method and its later statistical revision \cite{holtropMennen1982,holtrop1984}. The implementation checks for singular or mathematically inadmissible inputs to these equations. Machinery demand follows from installed power; voyage-plus-reserve fuel is deliberately budgeted at full installed power as a conservative onboard allowance. Appendix~\ref{sec_C2} specifies the powering and fuel equations and scenario inputs. The preliminary economic formulation belongs to the conceptual-ship-design benchmark discussed by Hart and Vlahopoulos; its printed cost coefficients and annual-transport equations are traceable to Yang's earlier ship-design model \cite{hart2010IMDO,yang2000minimax}. GenISys applies these coefficients to its own evaluated structural mass, power, voyage, and cargo responses. Economic outputs include acquisition cost and annual voyage, capital, and running costs. Equation~\eqref{eq_3} relates annual cargo, annual total cost, and RFR, where $m_C$ is cargo per voyage and $N_V$ is voyages per year:
\begin{equation}
\begin{aligned}
 Q_{yr}&=m_CN_V,\\
 C_{yr}&=C_{\mathrm{voy,yr}}+C_{\mathrm{cap,yr}}+C_{\mathrm{run,yr}},\\
 \mathrm{RFR}&=\frac{C_{yr}}{Q_{yr}}.
\end{aligned}
 \label{eq_3}
\end{equation}
Required freight rate (RFR) represents the modeled cost per tonne of cargo transported. It is calculated by dividing annual total cost by annual cargo transported and is expressed in pounds sterling per tonne. These empirical cost estimates support comparisons under the same assumptions; they are not market quotations. The exploratory free-range study allows speed, range, dimensions, and cargo capacity to vary. The central performance sweep and confirmation fix voyage range while allowing the remaining editable controls, including speed and dimensions, to vary; cargo capacity follows from their evaluation.

\subsection{Constraint Labels and Feasibility}
Completed designs in the feasibility-guidance study and all 29,970 completed retraining-confirmation evaluations use 187 formulated constraints. The full conditional catalogue contains 197 named checks: 162 assigned to local subsystem labels and 35 to system scope. Not every catalogue entry applies to every configuration. Most completed source-dataset evaluations also use 187 checks; four source designs use 182. The rule-based engineering constraints cover hull validity, empirical-method domains, geometry and subdivision, capacities, loading equilibrium, and preliminary structural requirements informed by classification and regulatory sources, including the American Bureau of Shipping (ABS) Marine Vessel Rules (MVR) and the International Association of Classification Societies (IACS) Common Structural Rules (CSR) \cite{ABS_MVR,IACS_CSR_2025}. Appendix~\ref{sec_D3} lists the complete catalogue, recorded sources, and local or system assignment.

A design is fully feasible only if its evaluation completes and the formulated engineering constraints are satisfied. In Equation~\eqref{eq_4}, $a_k=1$ identifies a check included in the feasibility assessment, $P_k(\mathbf x)$ equals one when that check passes and zero when it fails, and $\chi_{\mathrm{eval}}$ equals one when evaluation completes and zero otherwise. Full feasibility is
\begin{equation}
 F(\mathbf x)=\chi_{\mathrm{eval}}
 \prod_{k:a_k=1}P_k(\mathbf x).
 \label{eq_4}
\end{equation}
A local label restricts the product to that subsystem's checks. Appendix~\ref{sec_D3} defines residual signs, strict comparisons, numerical allowances, and the conditions under which each constraint is evaluated. Incomplete evaluations have unknown labels and remain unsuccessful requests. Throughout this paper, ``fully feasible'' means satisfying the formulated engineering constraints in Equation~\eqref{eq_4}. The reported evaluations do not include tank free-surface corrections needed for corrected initial stability, and the flooding calculation is a coarse diagnostic rather than a complete damage-stability assessment. These outputs are therefore reported separately and excluded from the 187-check acceptance criterion. Complete vessel approval is outside the evaluator's scope.

\section{Dataset Construction}
This section explains random dataset construction, equation-informed repair, and the resulting populations and training splits.

\subsection{Random Population and Feasible-Data Augmentation}
Dataset construction began with 5,000 starting sets of structural parameters, referred to as structural parents. Each was combined with four hull variants and five operating conditions, giving 20 ship designs per structural parent and 100,000 designs in total. Of these, 81,675 completed evaluation, and none satisfied the formulated constraints. Completed random designs had a mean of 32.29 violations, a median of 33, and a range of 8--56. The 18,325 incomplete evaluations comprise 5,873 physical-infeasibility rejections, 2,531 empirical-method-domain rejections, and 9,921 numerical or implementation failures.

To obtain feasible training examples, completed sources with fewer than 33 violations were selected for three stochastic repair alternatives each. The saved selection contains 39,333 sources from 3,273 structural parents. One additional source meets the numerical selection rule but is absent from the recorded selection without an explanation; reported counts use the realized population. Repair generated 117,999 alternatives, of which 88,764 completed evaluation and 16,247 were fully feasible, giving an overall feasibility rate of 13.77\% across all submitted repairs. The 29,235 incomplete repairs comprise 15,610 physical rejections, 9 method-domain rejections, and 13,616 numerical or implementation failures.

\subsection{One-Pass Stochastic Parametric Repair}
\label{sec:repair-method}
The authors developed an equation-informed repair procedure for the coupled ship-system evaluator, extending the approach introduced in MiDShip \cite{bagazinski2026midship}. The procedure uses the equations behind failed constraints to determine which design parameters to change. Adaptive structural-repair methods provide precedents for constraint-to-variable localization \cite{samanipour2020adaptive,cai2021adaptive,jelovica2024improved,cai2023neuralrepair}. Hull and dimension changes precede subdivision, scantlings, hull-girder capacity, and cargo allocation. The procedure generates three repair alternatives for each selected design. Where several permitted parameter changes can address a failed check, one is chosen randomly. The margin by which a proposed change is intended to satisfy the check is also varied randomly.

Each repair alternative received one repair pass followed by a complete ship evaluation, applying a consistent procedure across all candidate designs. A single pass can reduce constraint violations without satisfying every coupled constraint. Only outcomes that pass all counted checks are classified as fully feasible; other outcomes are retained as unsuccessful repair attempts. The remaining violations illustrate the difficulty of constructing fully feasible training data, even with equation-informed repair. Performance objectives were not optimized during repair. Appendix~\ref{app_B} explains the repair rules and search procedures.

\subsection{Dataset Outcomes and Training Splits}
Table~\ref{tab_1} shows submitted, completed, and feasible populations. Each completed repair is compared with its original design. If several repair alternatives complete for the same original design, that original appears once for each completed alternative. In this matched comparison, the mean violation count falls from 24.95 before repair to 1.66 after repair, a 93.33\% reduction. The separate mean of 32.29 violations reported above describes the entire completed random population. Every completed alternative improved on its source, but the selected sources and shared parents preclude interpreting this as an independent random-sample success rate. The 16,247 feasible repairs descend from 6,884 source designs in 1,113 structural parents.

\begin{table*}[t]
\caption[Dataset construction and feasibility]{Dataset construction outcomes. Overall feasibility uses all submitted designs. Counts describe data coverage; bold marks the higher feasibility rate of the two component populations. The combined row is an aggregate.}
\label{tab_1}
\centering
\begin{arxivtablefit}
\begin{tabular}{lrrrr}
\toprule
Population & Submitted & Completed & Fully feasible & Overall feasibility (\%) $\uparrow$\\
\midrule
Random & 100,000 & 81,675 & 0 & 0.00\\
Repaired & 117,999 & 88,764 & 16,247 & \textbf{13.77}\\
Combined & 217,999 & 170,439 & 16,247 & 7.45\\
\bottomrule
\end{tabular}
\end{arxivtablefit}
\end{table*}

The combined dataset retains all 217,999 proposed designs, including 47,560 incomplete evaluations. Model fitting uses only completed designs with known labels. Shared structural parents and their descendants remain in the same training, validation, or test group, preventing operating variants and repair relatives from crossing splits. The completed-design split contains 119,143 training, 25,700 validation, and 25,596 test designs. Feasible-only generators use the 11,195 fully feasible training designs; full-data generators use all 119,143 known-label training designs. These populations differ in both label composition and sample size. Most repaired training designs that remain infeasible fail the hopper-structure checks (Appendix~\ref{app_A}). The hopper subsystem includes upper and lower hopper plating, brackets, and stiffeners. This concentration of failures motivates a redesign study that changes only the hopper parameters while keeping the other independent design parameters fixed.

An additional 50,000 uniformly proposed hulls supply positive and negative examples for hull feasibility classifiers, with 35,000/7,500/7,500 assigned to training/validation/test. They do not train the whole-ship generators or ship-global classifiers and do not carry whole-ship feasibility labels. Because every completed ship design passes the hull feasibility checks, the ship dataset contains no examples of failing hulls. The additional hull dataset provides both passing and failing examples so the classifiers can learn to distinguish them.

\section{Subsystem-Conditional Diffusion}
This section defines preprocessing, the hierarchical and simultaneous diffusion architectures, and feasibility conditioning and feasibility guidance.

\subsection{Transformation and Denoising Networks}
Before training, parameters with different units and ranges are mapped to a common numerical scale for diffusion. Each value's position within its training-data distribution determines its position on a standard-normal scale. Extreme transformed values are bounded to keep them finite. Each population uses scikit-learn's \texttt{QuantileTransformer} with \texttt{output\_distribution='normal'}, fitted only to its training designs. Equation~\eqref{eq_5} maps physical control $j$ to its transformed coordinate:
\begin{equation}
 z_j=\mathcal Q_{j,train}(x_j),
 \label{eq_5}
\end{equation}
where $\mathcal Q_{j,train}$ is the fitted featurewise operator. It interpolates empirical quantiles and maps cumulative probabilities through the standard-normal inverse cumulative distribution function (CDF). The inverse transformation returns model outputs to physical parameter values by interpolating the fitted quantiles within their fitted marginal ranges. The transform is shared across architectures within each training population and differs from the physical-space scaling used for distance metrics. This provides continuous transformed coordinates for tabular diffusion \cite{kotelnikov2023tabddpm}.

After inverse transformation, whole-number controls are rounded to the nearest whole number and two-option controls are set to 1 for values of at least 0.5 and to 0 otherwise. For stiffener-shape controls, 1 selects a T-shaped section and 0 selects an L-shaped section. Deterministic reconstruction supplies dependent coordinates. Structural model construction applies the automatic parameter adjustments inherited from MiDShip, including snapping hopper-bracket endpoints to structural-member positions and enforcing compatible dimensions. MiDShip documents class-specific parameter cleanup and bracket alignment in its Sec.~3.1 \cite{bagazinski2026midship}; these inherited operations explain differences between submitted and evaluated structural parameters. Feasibility labels and RFR describe the resulting fully evaluated design; the unevaluated proposal has no assigned feasibility label. These adjustments are distinct from stochastic design repair or diffusion-based subsystem redesign. No explicit latent clipping is applied during diffusion; the fitted inverse transform and geometry adjustments still restrict the reconstructed physical designs.

The forward process follows Ho et al.'s noise-prediction DDPM formulation and time convention \cite{ho2020ddpm}. Equation~\eqref{eq_6} gives the noisy state from the clean transformed design $\mathbf z_0$, with $t=1,\ldots,T$ and $T=1000$:
\begin{equation}
 \mathbf z_t=\sqrt{\bar a_t}\mathbf z_0+
 \sqrt{1-\bar a_t}\boldsymbol\epsilon,
 \qquad \boldsymbol\epsilon\sim\mathcal N(\mathbf 0,\mathbf I),
 \label{eq_6}
\end{equation}
where $a_t=1-\beta_t$, $\bar a_t=\prod_{s=1}^{t}a_s$, and $\bar a_0=1$. The implementation uses zero-based schedule indices $k=0,\ldots,T-1$, related by $t=k+1$. Let $\mathcal I_\ell$ contain the $d_\ell$ controls owned by block $\ell$. Equation~\eqref{eq_7} defines the coordinate-averaged loss minimized independently by each subsystem denoiser:
\begin{equation}
 \mathcal L_\ell=\mathbb E\!\left[
 \frac{1}{d_\ell}\sum_{j\in\mathcal I_\ell}
 \left(\epsilon_j-\epsilon_{\theta_\ell,j}
 (\mathbf z_t^{(\ell)},t,\mathbf c_t^{(\ell)})\right)^2\right].
 \label{eq_7}
\end{equation}
The expectation is over training rows, uniformly sampled diffusion times, and Gaussian corruption; the implementation averages over both batch rows and owned coordinates. Block optimizers are separate, with no additional cross-block loss weighting. For the monolithic model, one block owns all 133 controls. The context $\mathbf c_t^{(\ell)}$ contains optional pass/fail labels as feasibility-conditioning inputs and the architecture-specific predecessor or complementary coordinates defined below.

Equations~\eqref{eq_8} and~\eqref{eq_9} give the tested reverse mean and fixed posterior variance for each owned subvector:
\begin{align}
 \boldsymbol\mu_{\theta_\ell,t}
 &=\frac{1}{\sqrt{a_t}}\left[
 \mathbf z_t^{(\ell)}-
 \frac{\beta_t\boldsymbol\epsilon_{\theta_\ell}}
 {\sqrt{1-\bar a_t}}\right],
 \label{eq_8}\\
 \sigma_t^2=\widetilde\beta_t
 &=\beta_t\frac{1-\bar a_{t-1}}{1-\bar a_t}.
 \label{eq_9}
\end{align}
The noise prediction in Equation~\eqref{eq_8} has the inputs in Equation~\eqref{eq_7}. Unguided sampling uses $\mathbf z_{t-1}^{(\ell)}=\boldsymbol\mu_{\theta_\ell,t}+\sigma_t\boldsymbol\eta_t^{(\ell)}$, with independent standard-normal innovations. At $t=1$, $\widetilde\beta_1=0$ and the update returns the mean. No predicted-clean-state clipping is applied. All diffusion coordinates and their noise quantities are dimensionless.
The schedule contains 1,000 linearly spaced variances from 0.001 to 0.2. Networks use 256-dimensional input/time embeddings, sigmoid linear unit (SiLU) activations, batch normalization, dropout, and residual connections. The monolithic DDPM has four hidden layers of 1,024 units; each HCDDPM and SCDDPM subsystem denoiser has four hidden layers of 256 units. The decomposed models contain 20 denoisers apiece. Network capacities are not matched across architectures.

Each denoiser receives 10,000 optimizer updates with one newly drawn batch of 1,024 designs per update, using AdamW with initial learning rate of $2.5\times10^{-4}$. These updates are not complete dataset passes. After training, the models were sampled to generate candidate ship designs. The feasibility-guidance ablation, feasibility-strength sweep, and two-arm fixed-range study stop at $\mathbf z_1$, omitting $\mathbf z_1\to\mathbf z_0$ ($t=1$, code index $k=0$). The fixed-range selection, fresh-fit confirmation, and independent parent-cohort studies include this transition; Appendix~\ref{sec_D4} records the protocols.

\subsection{Hierarchical Conditional Denoising Diffusion Probabilistic Model}

HCDDPM adapts Chen and Fuge's hierarchical interdependent-part generation to tabular diffusion \cite{chen2019hierarchical}. Its disciplinary progression resembles a pass through Evans' design spiral \cite{evans1959basic}; the tested sampler does not revisit earlier blocks. Let $\mathbf x^{(\ell)}$ denote subsystem $\ell$'s physical subvector and $y^{(\ell)}$ its local feasibility label. Equation~\eqref{eq_10} defines the ordered HCDDPM model for the fixed partition $L=20$:
\begin{equation}
 p_\theta(\mathbf{x}\mid\mathbf{y})
 :=\prod_{\ell=1}^{L}
 p_{\theta_\ell}\!\left(
 \mathbf{x}^{(\ell)}\mid\mathbf{x}^{(<\ell)},
 \mathbf{y}^{(\ell)}\right).
 \label{eq_10}
\end{equation}
Here $\mathbf{y}^{(\ell)}$ is an optional subsystem pass/fail conditioning label and is
omitted for feasible-only training.
Each factor is a separate DDPM. The frozen order proceeds from six hull-form blocks through powering and compartment arrangement, followed by 12 structural blocks; Table~\ref{tab_A1} gives the complete subsystem partition. During training, downstream denoisers receive observed predecessor blocks from the same row. During sampling, each block completes its full reverse chain before the next denoiser receives the generated predecessor state. Early generation errors therefore become downstream conditions and cannot be revised.

The designer-selected sequence determines available context. Range lies in block 7 and is unavailable to the first six hull blocks, even when prescribed by a sampling lock. Choosing an effective order can require prior system knowledge or sequence analysis. The comparison examines sequential and simultaneous subsystem generation using models with similar parameter counts: 10.89 million for HCDDPM and 11.30 million for SCDDPM (Appendix~\ref{app_I}). Differences in training and guidance also limit attribution of the results to generation order alone. Other hierarchical configurations can supply global mission inputs or revisit decisions.

Figure~\ref{fig_2}(a) shows four representative blocks in that fixed sequence. Each depicted DDPM completes its own reverse chain before its output becomes a condition for later blocks.

\subsection{Simultaneous Conditional Denoising Diffusion Probabilistic Model}

\begin{figure*}[!t]
\centering\includegraphics[width=6.0in]{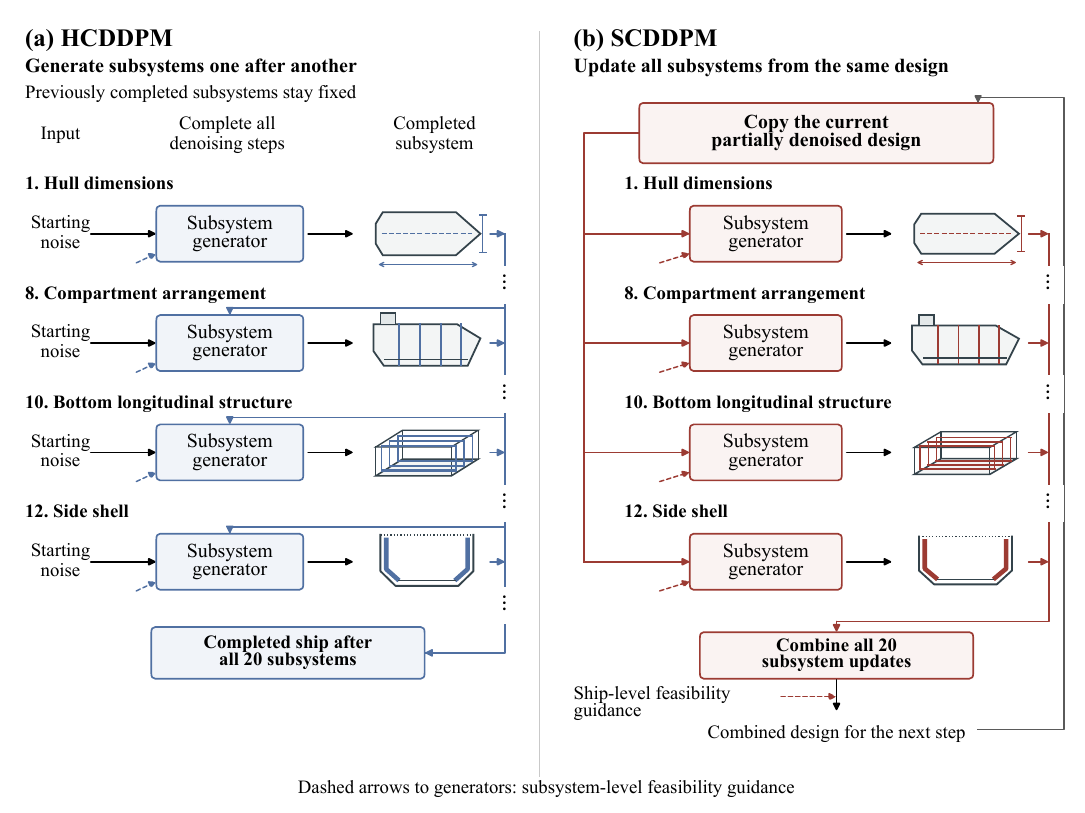}
\caption[Sequential and simultaneous subsystem generation]{Sequential and simultaneous subsystem generation. (a) HCDDPM generates each subsystem to completion before beginning the next. Previously completed subsystems remain fixed and provide context for later ones. (b) SCDDPM updates all subsystems using the same partially denoised design at each step, then combines their proposed updates before the next step. Dashed arrows show optional feasibility guidance. Subsystem-level feasibility guidance changes only the associated subsystem's design parameters. Ship-level feasibility guidance acts on the combined SCDDPM design after the subsystem updates have been assembled. Four example subsystems are illustrated: hull dimensions, compartment arrangement, longitudinal bottom structure, and side shell. Intermediate subsystems are omitted.}
\label{fig_2}
\end{figure*}

For block $\ell$ and diffusion time $t\in\{1,\ldots,T\}$, let $\mathbf{z}_t^{(-\ell)}$ denote the diffusion-time states
of all coordinates outside $\mathcal{I}_\ell$, and let $\mathbf{c}^{(\ell)}$
denote the optional subsystem pass/fail conditioning label. Equation~\eqref{eq_11} specifies the inputs to the 20 SCDDPM conditional denoisers:
\begin{equation}
 \boldsymbol{\epsilon}_{\theta_\ell}
 \left(\mathbf{z}^{(\ell)}_t,t,\mathbf{z}_t^{(-\ell)},
 \mathbf{c}^{(\ell)}\right),
 \qquad \ell=1,\ldots,L.
 \label{eq_11}
\end{equation}
During training, the complementary blocks are corrupted at the same diffusion time as the generated block. At reverse step $t$, the current joint state is copied before any update, and every block proposal is computed from that same start-of-step state:
\begin{equation}
 \mathbf{z}_{t-1}^{(\ell)}=
 R_{\theta_\ell}\!\left(
 \mathbf{z}_{t}^{(\ell)},\mathbf{z}_{t}^{(-\ell)},
 \mathbf{c}^{(\ell)},t,\boldsymbol{\eta}_t^{(\ell)}
 \right).
 \label{eq_12}
\end{equation}
Here $R_{\theta_\ell}$ is the block's reverse update, $\boldsymbol{\eta}_t^{(\ell)}$ is its reverse-process innovation, and the
condition is omitted for feasible-only training. Equation~\eqref{eq_12}
is synchronous or Jacobi-like even if the 20 denoisers are evaluated serially.
``Simultaneous'' therefore denotes reciprocal conditioning at every reverse
time, not necessarily parallel hardware execution. Unlike HCDDPM, a later
block can influence an earlier block at the next reverse step. Both monolithic
DDPM and SCDDPM condition on the full noisy system state. SCDDPM organizes
joint denoising through subsystem-owned output models, with every block
updated from the same start-of-step state before assembly. These block
interfaces support subsystem-level feasibility guidance and locked-subsystem generation.
SCDDPM uses approximately $20T$ denoiser evaluations per generated batch, compared
with $T$ for the monolithic model; HCDDPM likewise uses approximately $20T$,
but arranges them as 20 complete sequential chains. Before guidance and locking, concatenated block predictions and innovations give a standard joint DDPM update. Figure~\ref{fig_2}(b) shows the shared-state update and assembly of block proposals.

\subsection{Feasibility Conditioning and Feasibility Guidance}
Feasibility conditioning supplies a pass/fail label as an additional input to the generator. Generators trained only on fully feasible designs receive no such input. When training uses both feasible and infeasible designs, DDPM receives a whole-ship pass/fail label, while HCDDPM and SCDDPM receive subsystem pass/fail labels. During generation, these inputs are set to ``pass'' to request feasible designs. Engineering evaluation then determines whether the generated designs actually pass. Unknown-label designs are excluded. Consequently, feasible-only versus full-data comparisons change the population, sample size, fitted transform, and conditioning inputs together. Appendix~\ref{app_G} provides details of alternative training on locally feasible subsystem data.

For fixed-range generation, voyage range is prescribed at sampling time. The generators are trained across voyage ranges; a requested range is imposed by locking its control coordinate during reverse diffusion. The locked coordinate follows the corresponding forward-noised reference trajectory, and its exact physical value is restored before evaluation. Feasibility and performance guidance act only on editable coordinates; the requested range itself is fixed rather than approached through a guidance gradient. A different range within the supported design domain, such as 2,000 nmi, can therefore be requested without retraining the generator. The five-training confirmation tests this sampling procedure at 5,000 nmi; it does not establish equivalent performance at other ranges.

System guidance denotes the ship-global classifier; combined guidance applies local subsystem classifiers followed by this system classifier. Classifiers are trained on known-label training designs in the corresponding generator coordinates. Each population has a ship-global classifier, a hull-global classifier, and separate contextual local classifier collections for HCDDPM and SCDDPM. Local gradients act only on the owned block. The hull principal-dimension and hull-midbody classifiers have no failing training examples, so neither supplies subsystem-level feasibility guidance. Their engineering checks remain part of the feasibility assessment. Classifiers use a 128-dimensional input embedding, three 128-unit hidden layers, and a sigmoid pass output. Informative classifiers receive 150,000 updates with batches of 1,024. These feasibility classifiers are trained on clean transformed designs without timestep inputs and are queried on reverse-process states during sampling. Separate performance regression models are timestep-conditioned (Appendix~\ref{app_H}).

Let $p_j(\mathbf z)$ be a classifier's pass probability. Equation~\eqref{eq_13} gives the gated log-probability direction following GenPFN~\cite{bagazinski5498646genpfn}:
\begin{equation}
 \mathbf g_j(\mathbf z)=\mathbb 1[p_j(\mathbf z)<0.7]
 \nabla_{\mathbf z}\log\max(p_j(\mathbf z),10^{-6}).
 \label{eq_13}
\end{equation}
Equation~\eqref{eq_14} gives the implemented noise-blend update for one active guidance stage:
\begin{equation}
 \mathbf z_{t-1}=\boldsymbol\mu_\theta(\mathbf z_t,t,\mathbf c)
 +\sigma_t\big[(1-\alpha)\boldsymbol\eta+\alpha\mathbf g(\mathbf z_t)\big],
 \label{eq_14}
\end{equation}
where $\sigma_t$ is the posterior standard deviation and $\boldsymbol\eta$ is Gaussian noise. Setting a satisfied classifier's gradient to zero does not restore the removed noise fraction. Thus, increasing $\alpha$ changes both the guidance displacement and stochastic exploration. The update differs from the standard timestep-conditioned classifier formulation \cite{dhariwal2021guidance}.

Equation~\eqref{eq_15} caps the actual classifier displacement $\mathbf d_t=\alpha\sigma_t\mathbf g_t$ to limit instability:
\begin{equation}
 \mathbf d_t^{cap}=\mathbf d_t\min\left(1,
 \frac{\sqrt{\beta_t}\sqrt{d_b}}{\|\mathbf d_t\|_2}\right),
 \label{eq_15}
\end{equation}
with zero displacement left unchanged. Here $d_b$ is the number of affected coordinates, equal to 133 for a full-vector move. Monolithic guidance caps the summed ship-global and hull-global moves. HCDDPM guides only the active block. SCDDPM supports subsystem-level feasibility guidance, ship-global feasibility guidance after block assembly, or the two stages in sequence. Combined SCDDPM guidance first updates individual subsystems, then applies ship-global feasibility guidance to the assembled design. Both stages use the configured feasibility-guidance strength. Each subsystem update, the ship-global update, and their combined displacement are capped. Adding the ship-global stage adds another update; the strength is applied per stage rather than divided between the stages. Appendix~\ref{sec_D5} gives the exact fixed-range sampler order; Appendices~\ref{sec_D4} and~\ref{sec_D1} specify study variants and physical reconstruction.

\section{Feasibility Guidance: Noise and Gradient Ablation}
This section evaluates how reverse-process noise and feasibility guidance affect feasibility and mean nearest-neighbor distance among generated designs.

The results compare the complete generation pipelines used in this study. Because the models differ in size, training effort, and guidance settings, the observed differences cannot be attributed to simultaneous subsystem conditioning alone.

The studies address separate questions: the ablation separates noise reduction from guidance; performance experiments select guidance strengths and confirm outcomes across fresh fits; redesign tests changes to selected subsystems. Appendix~\ref{sec_D4} summarizes their purposes and comparison methods; Appendix~\ref{sec_C7} reports engineering numerical verification.

\subsection{Experimental Design and Outcome Measures}
The primary feasibility-guidance experiment reuses one feasible-only fit per architecture and separates noise reduction from feasibility guidance. Each architecture generates four arms: ordinary reverse noise without guidance (A), noise multiplier 0.3 without guidance (B), ordinary noise with guidance (C), and noise multiplier 0.3 with guidance (D). Feasibility guidance uses the capped classifier displacement at strength $\alpha=0.7$ in Equation~\eqref{eq_15}. Setting A provides the baseline with standard noise and no feasibility guidance. Setting B tests noise reduction alone, and setting C tests feasibility guidance with standard noise. Setting D tests whether feasibility guidance provides further improvement when noise is already reduced. The sampler runs $t=T,\ldots,2$, omitting $\mathbf z_1\to\mathbf z_0$. DDPM uses ship-global plus hull guidance, HCDDPM guides the active block locally, and SCDDPM applies ship-global guidance after block assembly. These contexts are not architecture-matched classifier targets.

All 133 controls remain free. Ten independently seeded batches of 100 proposals per arm give 12,000 requests. Within each architecture and batch, all four arms share initial states and corresponding reverse-process innovations. Each request receives the normal full ship-system evaluation and retained outputs. Equation~\eqref{eq_16} defines feasible and completed yields, where $N$, $N_C$, and $N_F$ are requested, completed, and fully feasible counts:
\begin{equation}
 r_F=N_F/N,\qquad r_C=N_C/N.
 \label{eq_16}
\end{equation}
Incomplete evaluations remain unsuccessful requests; their unknown constraint labels are not assigned artificial violation counts. Planned paired $t$ tests compare the ten batch feasibility differences for D--B, C--A, and B--A in each architecture, with Holm adjustment over nine tests. Intervals are unadjusted 95\% paired $t$ intervals and describe sampling variation conditional on the saved fits. For HCDDPM, enabling guidance also changes how model outputs are converted into physical design parameters. Its measured guidance gains therefore cannot be attributed to guidance alone.

\subsection{Feasible Yield and Sampling Effects}
\begin{figure*}[!t]
\centering\includegraphics[width=6.0in]{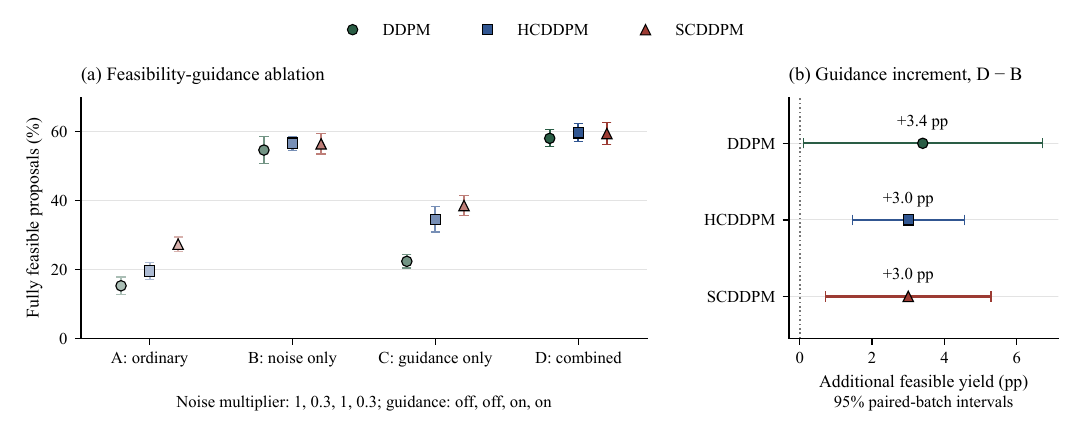}
\caption[Separating noise reduction and feasibility guidance]{Effects of noise reduction and feasibility guidance on feasible design yield in a four-arm ablation with 1,000 requested proposals per architecture and arm: A, ordinary noise without guidance; B, reduced noise without guidance; C, ordinary noise with guidance; D, reduced noise with guidance. All guidance uses strength 0.7, and reduced-noise arms use multiplier 0.3. The ablation sampler ends at $\mathbf z_1$, omitting the final $\mathbf z_1\rightarrow\mathbf z_0$ transition. (a) Mean feasibility and 95\% $t$ intervals across ten 100-request batches. (b) Paired D--B increments and unadjusted 95\% paired $t$ intervals. Feasibility includes all requests. Holm adjustment covers the nine planned contrasts; intervals quantify sampling variation within fixed fits. Architecture-specific guidance contexts and the HCDDPM parameter-conversion difference are described in the text.}
\label{fig_3}
\end{figure*}

Figure~\ref{fig_3} shows the effects of noise reduction and feasibility guidance on feasible yield. Of 12,000 requests, 11,771 completed evaluation and 5,031 satisfied the formulated engineering constraints. Unguided feasible yields were 15.3\%, 19.6\%, and 27.4\% for DDPM, HCDDPM, and SCDDPM, respectively (Table~\ref{tab_F1}). SCDDPM exceeded DDPM by 12.1 percentage points (79.1\% relatively) and HCDDPM by 7.8 points (39.8\% relatively). These descriptive architecture comparisons condition on one saved fit per architecture.

Noise reduction alone raised feasibility by 39.4, 37.1, and 29.1 percentage points for DDPM, HCDDPM, and SCDDPM. Guidance with ordinary noise added 7.1, 15.0, and 11.2 points. Beyond reduced noise, guidance added a further 3.4 points for DDPM (95\% interval 0.09--6.71), 3.0 for HCDDPM (1.45--4.55), and 3.0 for SCDDPM (0.71--5.29). All nine planned arm contrasts meet the adjusted significance criterion ($p_{\mathrm{Holm}}\leq0.0451$; Appendix~\ref{app_E}). HCDDPM's guidance comparisons include the parameter-conversion change described above; DDPM and SCDDPM isolate guidance within their respective fixed-fit protocols. The blended feasible yields converge to 58.1\%, 59.7\%, and 59.5\%. Most of the blended feasibility gain is therefore attainable by reducing noise, with smaller incremental gains in the guidance-enabled arms. Noise and guidance interact, so their effects should not be interpreted as an additive decomposition.

\subsection{Nearest-Neighbor Distances Among Fully Feasible Designs}
Diversity analysis retains only completed, fully feasible generations. For each condition, 200 subsets of 100 designs are selected from the already generated, fully feasible designs. No design appears twice within one subset, but different subsets can overlap. The standard deviation (SD) across these subsets measures sensitivity to which designs are included; it does not measure variation between independently trained models. Physical scaling is fitted only to feasible training data, with no clipping. All distances use the 133 independent parameters selected by the generators, using their evaluated, cleaned values. Appendix~\ref{app_F} reports directional distances to both feasible training and source-parent-separated test references.

Distances compare ship parameters after scaling them using the feasible training data, so they have no physical units. Mean nearest-neighbor distance measures the average distance from each generated design to its nearest other generated design. Larger values mean neighboring designs are more separated. The bidirectional Chamfer distance measures distances between generated and reference designs in both directions; smaller values indicate closer agreement.

In arm D, SCDDPM's mean nearest-neighbor distance is 2.455, versus 2.041 for HCDDPM, a 20.3\% increase in mean nearest-neighbor distance (Table~\ref{tab_F1}). HCDDPM has a larger mean nearest-neighbor distance without guidance (3.516 versus 3.324). Reduced noise accounts for most of the concentration; guidance with ordinary noise changes mean nearest-neighbor distance little. SCDDPM's larger mean nearest-neighbor distance under guidance does not mean closer agreement with the reference designs: its training-reference bidirectional Chamfer distance is 6.085, versus 6.052 for HCDDPM and 5.656 for DDPM. No exact duplicate 133-control vectors occur within any of the 72 feasible-generation cells; absence of duplicates does not establish novelty or rule out memorization.

\subsection{Design-Population Visualization}
\begin{figure*}[!t]
\centering\includegraphics[width=6.0in]{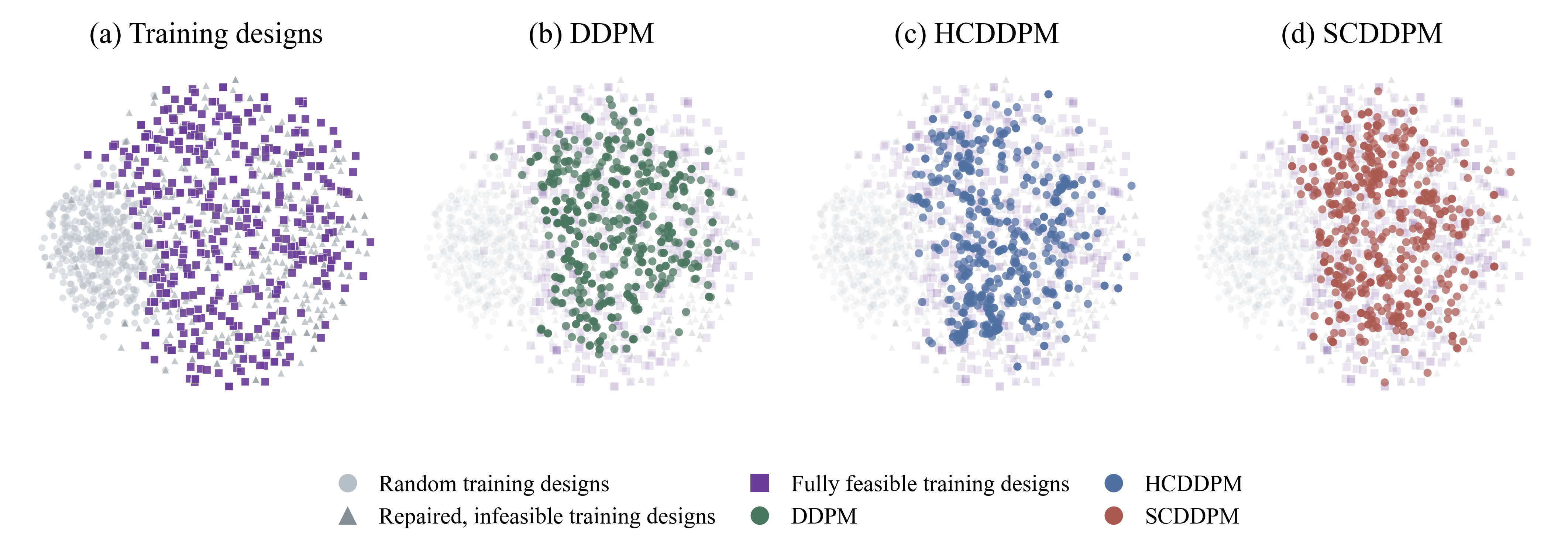}
\caption[Matched-count design-population embedding]{Training populations and fully feasible generated designs in one jointly fitted t-SNE projection. (a) Disjoint completed-random, repaired-infeasible, and fully feasible training groups, with 400 designs each; purple squares identify fully feasible training examples. Panels (b)--(d) each overlay 512 fully feasible designs from DDPM, HCDDPM, and SCDDPM, respectively, on the same training background. Appendix~\ref{app_F} specifies the sampling conditions. The projection is qualitative: area, density, and distance do not measure coverage, feasible yield, or architectural superiority.}
\label{fig_4}
\end{figure*}

Figure~\ref{fig_4} shows four views of one jointly fitted t-distributed stochastic neighbor embedding (t-SNE) \cite{vandermaaten2008tsne}. The four panels share the same projection and display limits, showing how the training populations and fully feasible designs from the three models relate.

The displayed feasible designs from all three models lie in similar regions of this projection, overlapping repaired and fully feasible training examples. The figure illustrates how generated designs relate to the training data; numerical distance measures provide the quantitative comparisons between models.

\section{Representative Integrated Ship Evaluation}
This section traces subsystem interactions through one evaluated ship, connecting geometry and mass budgets to cargo allocation, loading, and preliminary structural adequacy.

\begin{figure*}[!t]
\centering\includegraphics[width=6.0in]{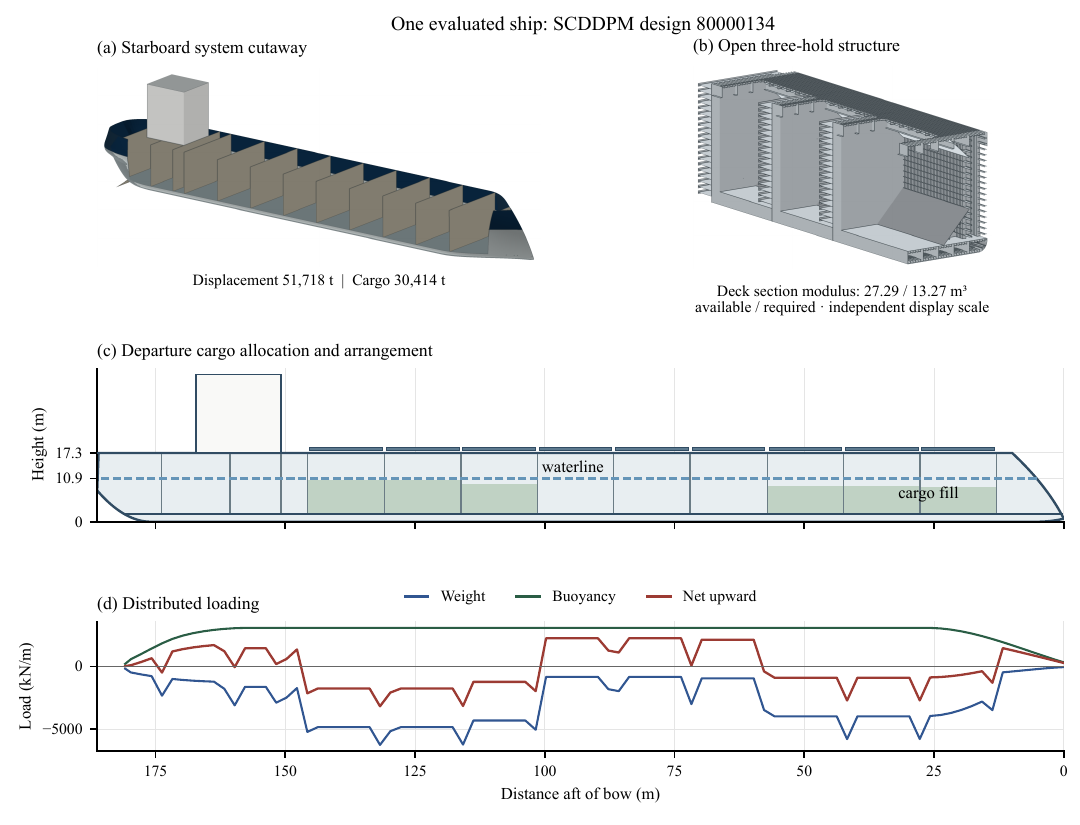}
\caption[Linked ship-system evaluation example]{Integrated evaluation of SCDDPM design 80000134 from the independent performance-validation cohort in Appendix~\ref{app_H}. (a) Starboard cutaway and (b) representative structure, displayed at different scales for clarity. Stern left, bow right. The entire forwardmost structural bulkhead family is hidden only in (b) but retained in evaluation. (c) Departure cargo arrangement and (d) saved loading share a longitudinal scale. Deckhouse and hatch-cover profiles are illustrative. Loads are 2-metre bin averages, with buoyancy positive and weight negative. Section-modulus callouts compare available and required gross deck values. Figure~\ref{fig_C1} shows unbinned shear and bending.}
\label{fig_5}
\end{figure*}

Figure~\ref{fig_5} shows the linked hull subdivision, representative structure, departure cargo arrangement, and saved distributed loading for one fully feasible SCDDPM validation design. Hull and draft determine available displacement; machinery, fuel, and structural weights reduce the cargo budget; hold volume and longitudinal placement determine whether that budget can be allocated; and the resulting loading supplies bending demand for the structural constraints. These relationships make the integrated model useful beyond generating separate hull and structural geometries. At a fixed hull, draft, and speed, the evaluator computes resistance before cargo allocation: reduced structural weight can release displacement for cargo without changing resistance. A resistance benefit from lighter structure would require a different, fixed-cargo comparison.

For this departure condition, 51,717.9\,t displacement balances 20,677.5\,t modeled lightship, 30,413.8\,t cargo, 274.2\,t fuel, and 352.4\,t miscellaneous deadweight. The saved loading is used to evaluate the total-load hull-girder strength constraint, which compares an available gross deck section modulus of 27.29\,m$^3$ with a required 13.27\,m$^3$. The three-hold half-model supplies representative mass and section properties, following MiDShip \cite{bagazinski2026midship}. Mirroring that half-model for display, hiding the forward bulkhead, and extending an inner-bottom display surface do not alter evaluated mass or section properties. The illustration explains the coupling without implying full-length scantling detail or a verified hatch-cover arrangement.

\section{Performance Guidance}
This section evaluates fixed-range performance guidance, repeatability across fresh generator fits, and locked-subsystem redesign.

\subsection{Guidance Selection and Outcome Definition}
The objective of RFR performance guidance is to reduce RFR. For comparative evaluation, a dataset-derived RFR reference value provides a common basis for comparing generators. Qualifying yield is the fraction of requests producing fully feasible designs with RFR at or below this reference value. Equation~\eqref{eq_17} defines it:
\begin{equation}
 Y(c)=\frac{1}{N}\sum_{i=1}^{N}\mathbb 1[F(\mathbf x_i)=1,\ \mathrm{RFR}(\mathbf x_i)\leq c].
 \label{eq_17}
\end{equation}
At 5,000 nmi, the RFR reference value is $c=24.5577$\,\pounds/t, calculated as the 25th percentile of the saved evaluated RFR values of the 603 fully feasible training designs within 4,750--5,250 nmi. The primary comparison reference and its numerical values were fixed before the guidance-strength selection sweep. Further RFR reductions remain desirable below the reference value. Incomplete or infeasible proposals contribute zero qualifying success. Feasible yield and mean RFR conditional on feasibility remain separate outcomes.

A separate selection sweep reuses one saved feasible-only generator fit per architecture to generate designs with prescribed voyage ranges of 1,000, 5,000, and 8,000 nmi. At each range, it tests RFR strengths $\gamma\in\{0,0.01,0.03,0.10\}$ with feasibility-guidance strength $\alpha=0.7$. Its 36,000 requests yielded 35,949 completed evaluations and 20,560 fully feasible designs. All 27 within-architecture nonzero-strength contrasts improve the qualifying yield after Holm adjustment ($p_{\mathrm{Holm}}\leq0.00152$). The complete cells, paired-batch tests, and range-specific outcomes remain in Appendix~\ref{app_J}; these saved-fit outcomes are not pooled with confirmation.

The frozen choice maximizes observed $Y(c)$ in each architecture's 5,000-nmi selection sweep: $\gamma=0.10$ for DDPM and SCDDPM, and $0.03$ for HCDDPM. HCDDPM's selected setting produced only one more qualifying proposal than $\gamma=0.10$ in the selection data (49.9\% versus 49.8\%). The inherited feasibility guidance uses system-plus-hull constraints for DDPM, local active-block constraints for HCDDPM, and local plus system constraints for SCDDPM. Performance guidance uses an RFR regression model to estimate modeled cost per tonne of cargo transported from the design parameters. For DDPM and SCDDPM, this regression model receives the current parameters for the entire ship. For HCDDPM, it receives the completed earlier subsystems and the subsystem currently being generated (Appendix~\ref{app_H}). Performance gradients point toward lower predicted RFR, are normalized over editable coordinates, and share a displacement cap with feasibility guidance.

\subsection{Cost Reduction-Guided Sampling Design Study}
This study uses guidance to reduce required freight rate when generating ship designs with a prescribed voyage range of 5,000 nautical miles. The confirmation compares DDPM, HCDDPM, and SCDDPM with guidance settings fixed by the preceding selection sweep. Model names in the confirmation results refer to these settings. Appendix~\ref{app_I} reports capacities, training work, and costs.

The confirmation requires generated bulk-carrier designs to have a prescribed voyage range of 5,000 nmi. Other editable controls, including speed and dimensions, remain free; cargo capacity follows from their system evaluation. Fifteen generators are trained afresh: five complete fits per architecture, all using the same 11,195 fully feasible training designs across all voyage ranges. Initialization, minibatch, and noising randomness change between fits; the source-parent split, fitted normalizers, classifiers, RFR regression models, evaluator, and economic assumptions remain fixed. Each denoiser receives 10,000 optimizer updates.

Each fit requests 1,000 proposals with feasibility guidance alone ($\gamma=0$) and 1,000 with its selected RFR strength, always with $\alpha=0.7$. Both arms use the same reduced-noise feasibility-guidance protocol; an unguided arm is not repeated. Within each training run, the two guidance settings are compared using ten paired batches of 100 generation requests. Corresponding requests share their starting noise and the random noise used during generation, including noise for locked parameters. This pairing reduces differences caused by random inputs when assessing the effect of added performance guidance. Separate random-number streams keep these inputs aligned. The reverse process runs $t=T,\ldots,1$, including $\mathbf z_1\to\mathbf z_0$, with guidance active at $t=800,\ldots,2$ (code indices $k=799,\ldots,1$), as in the selection sweep.

\begin{figure*}[t]
\centering\includegraphics[width=6.0in]{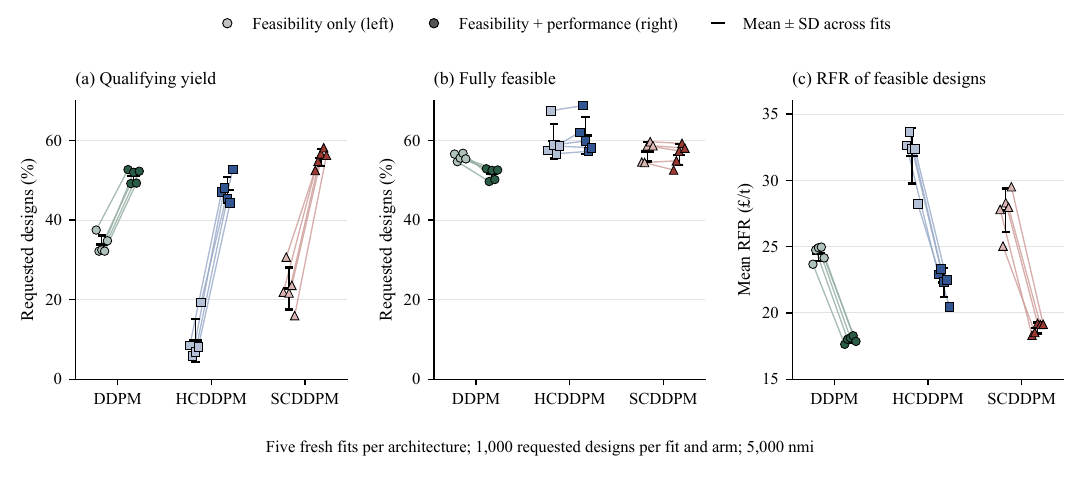}
\caption[Independent generator-retraining confirmation]{Feasibility and freight-rate outcomes for designs with a prescribed 5,000-nmi voyage range. Five independent fits per architecture each request 1,000 designs with feasibility guidance alone and 1,000 with the selected additional RFR performance guidance. Points and connecting lines show paired fit outcomes; bars show means $\pm$ between-fit sample SD. A qualifying design satisfies the formulated engineering constraints and has RFR at most the fixed RFR reference value of 24.5577\,\pounds/t. Yields include all requests; RFR uses fully feasible designs.}
\label{fig_6}
\end{figure*}

The independent unit is the complete generator fit, not an individual ship or batch. Three paired $t$ tests compare the five within-fit differences in $Y(c)$; two independent-sample Welch tests compare SCDDPM against DDPM and HCDDPM. The five planned statistical tests are adjusted together using Holm's method to account for multiple comparisons. The reported 95\% confidence intervals are not adjusted for multiple comparisons and reflect variation across independent training runs. Figure~\ref{fig_6} shows paired fit-level outcomes, and Table~\ref{tab_2} shows between-fit means and SDs; all 30 fit/arm outcomes and five contrasts appear in Appendix~\ref{app_I}.

\begin{table*}[t]
\caption[Independent five-fit confirmation at 5,000 nmi]{Outcomes for designs sampled with the required voyage range fixed at 5,000 nmi, using feasibility-guidance strength 0.7 and RFR strengths $\gamma$ frozen before confirmation. Each of five fresh training runs requests 1,000 designs per arm. For each architecture and guidance setting, RFR is first averaged across fully feasible designs within each training run. The five run means are then averaged; $\pm$ gives their sample standard deviation. RFR reduction compares these averages for feasibility-only and feasibility-plus-performance guidance. Yields use the feasibility-plus-performance arm and include all generation requests, including incomplete and infeasible results. Yield entries are mean $\pm$ sample SD across the five training runs; a qualifying design satisfies the formulated engineering constraints and has RFR at most 24.5577\,\pounds/t. Bold marks the highest yields and RFR reduction and the lowest mean RFR values across architectures; it does not indicate statistical significance.}
\label{tab_2}
\centering\small
\setlength{\tabcolsep}{4pt}
\begin{arxivtablefit}
\begin{tabular}{lrccccr}
\toprule
Architecture & $\gamma$ & \shortstack{Fully feasible\\yield (\%)} & \shortstack{Qualifying\\yield (\%)} & \shortstack{Feasibility-only\\RFR (\pounds/t)} & \shortstack{Feasibility + performance\\RFR (\pounds/t)} & \shortstack{RFR reduction\\(\%)}\\
\midrule
DDPM & 0.10 & $51.58\pm1.51$ & $51.10\pm1.71$ & $\mathbf{24.49\pm0.56}$ & $\mathbf{17.98\pm0.25}$ & 26.6\\
HCDDPM & 0.03 & $\mathbf{61.26\pm4.59}$ & $47.54\pm3.29$ & $31.85\pm2.10$ & $22.30\pm1.10$ & 30.0\\
SCDDPM & 0.10 & $56.46\pm2.69$ & $\mathbf{55.70\pm2.16}$ & $27.73\pm1.65$ & $18.86\pm0.41$ & \textbf{32.0}\\
\bottomrule
\end{tabular}
\end{arxivtablefit}
\end{table*}

Of 30,000 requests, 29,970 completed evaluation and 17,110 were fully feasible. Performance guidance improved $Y(c)$ in all five fits of every architecture. Mean paired gains are 17.26 percentage points for DDPM (95\% interval 15.36--19.16), 37.80 for HCDDPM (33.80--41.80), and 32.92 for SCDDPM (25.45--40.39); all remain significant after adjustment ($p_{\mathrm{Holm}}\leq0.000769$).

SCDDPM achieves the highest mean qualifying yield: 55.70\%, compared with 51.10\% for DDPM and 47.54\% for HCDDPM. Its gains are 4.60 percentage points over DDPM (95\% interval 1.73--7.47; $p_{\mathrm{Holm}}=0.00630$) and 8.16 over HCDDPM (3.99--12.33; $p_{\mathrm{Holm}}=0.00492$), equivalent to relative increases of 9.0\% and 17.2\%. The descriptive feasible-yield means are 61.26\% for HCDDPM, 56.46\% for SCDDPM, and 51.58\% for DDPM. Pooled across fits, 99.1\% of feasible performance-guided DDPM outputs and 98.7\% of SCDDPM outputs have RFR at or below the reference value. Their qualifying-yield difference therefore largely reflects higher feasible yield after RFR performance guidance. DDPM has the lowest mean feasible RFR (17.98\,\pounds/t, versus 18.86 for SCDDPM and 22.30 for HCDDPM). At the higher reference value defined by the training median, HCDDPM leads qualifying yield (60.10\%, versus 56.46\% and 51.58\%; Appendix~\ref{app_I}).

A secondary equal-sized feasible-set analysis averages results from 200 subsets of 100 already generated feasible designs within each fit before summarizing across fits. Under selected performance guidance, SCDDPM's 133-control nearest-neighbor distance is 20.3\% greater than HCDDPM's. This descriptive difference in mean nearest-neighbor distance does not establish closer agreement with held-out reference designs as measured by the bidirectional Chamfer distance (Appendix~\ref{app_I}).

Adding RFR performance guidance lowers the across-fit mean feasible RFR by 26.6\% for DDPM, 30.0\% for HCDDPM, and 32.0\% for SCDDPM (Table~\ref{tab_2}). It also improves the cost--cargo tradeoff among feasible designs across all three architectures at 1,000, 5,000, and 8,000 nmi in the separate two-arm study (Appendix~\ref{app_J}). Guided sets show greater Pareto coverage when annual cost is minimized and annual cargo transported is maximized; all nine net-coverage differences remain significant after Holm adjustment. Equation~\eqref{eq_3} links these outcomes to RFR. These feasible-population comparisons complement qualifying yield per request. Appendix~\ref{app_K} compares low-RFR dimensions and published-vessel particulars.

\subsection{Locked-Subsystem Redesign}
\label{sec:hopper-redesign}
Equation-informed repair constructs training data; diffusion-based redesign changes selected controls of an existing design. In a Hopper-only cohort of 100 starting designs, guidance raises fully feasible child yield from 10.6\% to 17.7\% for DDPM, 17.4\% to 25.3\% for HCDDPM, and 15.4\% to 23.9\% for SCDDPM; all three paired starting-design gains survive Holm adjustment. The performance-guided redesign study sought lower resistance by changing the bow and stern bulbs, and lower total steel weight by changing bulkhead or longitudinal bottom structure. In the independent starting-design confirmation, adding performance guidance produced a statistically supported increase in fully feasible designs that improved on their starting designs only for HCDDPM bulkhead redesign. Both feasibility guidance alone and feasibility-plus-performance guidance produced fully feasible designs that improved on their starting designs. These secondary studies demonstrate feasibility restoration and task-dependent objective improvement. Full protocols, starting-design coverage, and lock audits remain in Appendices~\ref{app_L} and~\ref{app_M}.

\section{Discussion}
The dataset illustrates how strongly subsystem decisions restrict the feasible design space of an integrated system. Randomly combining hull, structural, and operating parameters produced no designs satisfying the formulated constraints, while equation-informed repair supplied feasible examples for training (Table~\ref{tab_1}). For generative AI, the learning problem therefore extends beyond reproducing plausible values for individual components. A generated ship must combine those values consistently: structural and machinery weights affect available cargo mass, cargo arrangement affects loading, and loading determines structural demand. The complete-system evaluation in Figure~\ref{fig_5} shows why engineering feasibility must be assessed after these decisions have been brought together.

SCDDPM provides a way to organize this coordination around subsystem models. Each subsystem update uses information about the rest of the intermediate design, allowing decisions to remain revisable as generation proceeds. This is useful when changes in one subsystem alter the requirements placed on another. The hierarchical comparison also shows why the information available to each subsystem matters: its hull models cannot respond to the prescribed voyage range because that requirement enters later in the sequence. DDPM likewise allows joint revision of its design parameters; SCDDPM gives this process explicit subsystem interfaces for feasibility guidance and selective redesign. The practical question is therefore how subsystem organization and global information can support compatible design decisions.

The low-RFR guidance study illustrates the importance of considering feasibility and performance together. At the common cost target, SCDDPM returned more qualifying designs per request, largely because it retained more feasible outputs under performance guidance than DDPM. DDPM had a slightly lower mean estimated cost, but this difference was considered negligible because the approximate cost model cannot establish a corresponding real-world cost difference. HCDDPM generated more feasible designs overall and led qualifying yield when the cost target was relaxed. Choosing a generator consequently depends on the required engineering and performance outcomes, as well as computational effort: the higher yield per request from SCDDPM did not give it the lowest measured time per qualifying design (Appendix~\ref{app_I}).

The engineering properties of the generated ships help explain what lower transport cost means in this model. Performance guidance was associated with greater mean draft and deadweight across the architectures (Table~\ref{tab_I7}), linking the economic outcome to changes in the ship's cargo-carrying capability. These are comparisons between generated populations, rather than controlled changes to an otherwise fixed vessel. The selected generated examples also have higher block coefficients and lower deadweight-to-displacement ratios than the published comparators (Appendix~\ref{app_K}). These differences motivate closer examination of the physical representation and training population when judging the practical quality of generated designs. Satisfying the formulated constraints and achieving low modeled cost are useful outcomes, but comparison with real ships requires more detailed engineering evidence.

The noise-and-guidance ablation further shows that sample quality depends on how the learned distribution is explored. Reducing reverse-process noise produced most of the increase in feasible designs, with feasibility guidance providing a smaller additional gain. Greater concentration can help locate feasible designs while limiting the range of alternatives available for exploration. SCDDPM's larger mean nearest-neighbor distance than HCDDPM under the selected guidance did not establish better agreement with held-out designs as measured by bidirectional Chamfer distance. Feasible yield, performance, and variation therefore describe different aspects of a useful generated design population.

The redesign studies extend this interpretation to revising an existing system. Restricting changes to selected subsystem controls allows alternatives to be explored while retaining other independent design decisions. Feasibility guidance improved the recovery of feasible designs in the Hopper study. Added performance guidance produced a statistically supported improvement in only one of nine confirmation comparisons. The demonstrated ability to explore targeted modifications therefore motivates further study of how guidance models represent the engineering consequences of a proposed change.

\subsection{Limitations}
The 133 design controls and 187 rule-based engineering constraints capture important subsystem interactions, but do not define a complete bulk-carrier design. Engineering detail beyond the scope of this study could change the predicted loads, cargo capacity, and cost, or introduce additional design requirements. Satisfying the formulated constraints therefore establishes feasibility within the modeled design problem, while the performance and completeness of the vessel still require detailed engineering assessment.

The models learn from a synthetic population whose feasible examples were produced through the repair procedure. That procedure shapes the design choices and relationships represented in the training data, so the learned distribution may not capture the range of solutions found in other design settings. The results consequently establish performance within the studied bulk-carrier design space; generalization to other representations and engineering systems remains to be tested.

The study tests one method for providing global design information to SCDDPM's subsystem models, and compares the architectures as complete generation pipelines. Because other model and guidance choices also differ between these pipelines, their effects cannot be separated from the conditioning method. The comparison shows what the tested configurations can achieve, without establishing how much of the improvement comes from simultaneous conditioning alone or whether another conditioning strategy would coordinate subsystems more effectively.

\section{Conclusion}
GenISys demonstrates generative design of integrated bulk-carrier concepts by connecting subsystem parameters to rule-based engineering constraints and performance evaluation. Its dataset contains 217,999 proposed designs, including 16,247 that satisfy the formulated constraints. SCDDPM organizes generation around subsystem models that use a shared intermediate design, supporting subsystem-level feasibility guidance and selective redesign.

For bulk carriers with a required voyage range of 5,000 nautical miles, SCDDPM produced 9.0\% more designs that satisfied the formulated constraints and met a common low-cost target than DDPM, and 17.2\% more than HCDDPM, while maintaining similar estimated costs per tonne of cargo transported. These findings support the use of generative AI to coordinate coupled subsystem decisions and explore feasible integrated systems before detailed engineering design.

\subsection{Future Work}
Future work should increase the detail of the ship representation so that generated designs can be benchmarked more accurately against real bulk carriers. More complete hull, structural, arrangement, and machinery descriptions would support comparisons of mass distribution, cargo capacity, stability, powering, and operating performance. Benchmarking should use vessel particulars and measured or validated performance under comparable loading and operating conditions, together with improved cost estimates.

The engineering formulation should also be extended with additional design rules and physics-based machine-learning models trained on numerical simulations. Hydrodynamic and structural simulations could provide training data for models that represent design responses beyond the present empirical approximations. Their predictions would need to be assessed against independent simulations and engineering calculations before being used for feasibility or performance guidance.

Further studies should investigate how alternative conditioning methods communicate global requirements to subsystem models. Training and tuning the feasibility and performance guidance models for the intermediate designs encountered during generation could improve the quality of generated sample sets. These improvements should be assessed through engineering feasibility, performance, and variation among the resulting designs.

Applying GenISys and SCDDPM to other integrated engineering systems would test whether the approach generalizes beyond bulk carriers. Systems with different subsystem interactions, design representations, and data requirements would help establish where simultaneous subsystem generation is useful and what changes are needed for a new application.

\section*{Datasheet for the GenISys dataset}
The following subsections provide the datasheet for the GenISys dataset~\cite{gebru2021datasheets}. Counts refer to submitted design records unless a completed-evaluation subset is named explicitly.

\subsection*{Motivation}
\textbf{Why was this dataset created?}

GenISys was created to study generation, constraint evaluation, and modification of coupled early-stage bulk-carrier designs. It supplies a shared evaluated design space for comparing whole-system and subsystem-conditional generative models.

\textbf{Who created this dataset?}

The dataset was created by the MIT DeCoDE Lab.

\subsection*{Composition}
\textbf{What do the instances that comprise the dataset represent?}

Each instance is a synthetic parametric bulk-carrier proposal, rather than a surveyed or as-built vessel.

\textbf{How many instances are there in total?}

The dataset contains 217,999 submitted proposals: 100,000 random designs and 117,999 one-pass design-repair alternatives. Evaluations completed for 170,439 proposals, including 16,247 that satisfied the formulated engineering constraints. A further 50,000 hull-only proposals support hull-classifier training and are separate from the whole-ship dataset.

\textbf{What does each instance consist of?}

Each record contains a parametric proposal and its evaluation status. The evaluator uses 189 parameters reconstructed from 133 independent controls in 20 subsystems. Completed records include evaluations of the formulated constraints and engineering and economic responses. The catalogue contains 197 named checks; not every check applies to every design.

\textbf{Is there any missing data?}

The 47,560 incomplete whole-ship evaluations lack a complete set of constraint and performance results. Any recorded partial checks can support the diagnostics described in Appendix~\ref{app_E}; missing results remain unknown. Incomplete evaluations remain unsuccessful generation requests. Completed records carry the evaluated design vector and associated constraint evaluations and performance responses.

\subsection*{Data Collection Process}
\textbf{How was the data collected?}

Dataset construction began with 5,000 starting sets of structural parameters, referred to as structural parents. Each was combined with four hull variants and five operating conditions, giving 20 ship designs per structural parent and 100,000 designs in total. Completed random designs with fewer than 33 counted violations were eligible for three stochastic repair alternatives each. The selected repair sources share structural parents with the random population, so the two populations are not independent. The authors developed an equation-informed repair script using the constraint evaluation code. Every proposal was passed to the ship-system evaluator once, and the evaluation result was retained.

\textbf{Who was involved in the data collection process?}

The dataset was generated by researchers at MIT using automated parametric-design and evaluation pipelines.

\textbf{How was it decided which data to collect and which to exclude?}

All submitted proposals remain in the dataset, including incomplete evaluations. Only completed designs with known labels were used for model fitting.

\textbf{How were the data divided for model fitting?}

Completed designs were split by structural-parent group into 119,143 training, 25,700 validation, and 25,596 test records. Feasible-only generators trained on the 11,195 fully feasible training designs; full-data generators trained on all 119,143 completed training designs. The repair-source selection and parent relationships limit interpretation of these splits as independent draws from a fleet population.

\subsection*{Preprocessing, Cleaning, and Labeling}
\textbf{Was any preprocessing, cleaning, or labeling involved?}

The evaluator reconstructed dependent coordinates and applied the automated structural-geometry cleanup inherited from MiDShip. It recorded completion status, evaluations of the formulated constraints, and engineering responses. For model fitting, a featurewise quantile transform was fitted on the training split only. Discrete controls were reconstructed after sampling before exact evaluation. Full feasibility means satisfying the formulated engineering constraints; the optional rudimentary flooding calculation does not contribute to that label.

\textbf{How was this done and by whom?}

The processing and labeling were performed automatically using Python scripts developed by the authors.

\subsection*{Uses}
\textbf{For which engineering-design tasks is this dataset suitable?}

The completed records support early-stage ship-system generation, feasibility classification, performance prediction, and subsystem redesign under the implemented evaluator. These are research tasks for rule-based design evaluation; satisfying the implemented engineering constraints does not establish suitability for ship construction.

\textbf{Has this dataset been used for any tasks already?}

The present study compares three diffusion architectures through feasibility guidance, fixed-range freight-rate guidance, and locked-subsystem redesign. The confirmation experiment requires generated designs to have a prescribed voyage range of 5,000 nmi and uses five independent training runs per architecture. Generated evaluation cohorts are experiment outputs and are not added to the 217,999-proposal dataset count.

\textbf{What performance metrics are relevant for assessing tasks using this dataset?}

The five principal performance objectives are ship acquisition cost, annual total cost, required freight rate, structural steel weight, and resistance. Generative-model assessments also report feasible yield, qualifying yield, and design-space distance measures.

\subsection*{Distribution}
\textbf{How is this dataset distributed?}

The dataset repository is hosted on Hugging Face (\url{https://huggingface.co/datasets/DeCoDELab/GenISys}), and the accompanying code repository is hosted on GitHub (\url{https://github.com/noahbagz/GenISys}).

\textbf{Are there any restrictions or licenses on its use?}

The dataset and accompanying code are intended exclusively for non-commercial research and educational use. The formal license terms will be specified in the respective release repositories.

\textbf{How long will the dataset remain available?}

The dataset is intended to remain available indefinitely through the release repositories.

\subsection*{Maintenance}
\textbf{Who is responsible for the maintenance of the dataset?}

The dataset is maintained by the authors through Hugging Face: \url{https://huggingface.co/datasets/DeCoDELab/GenISys}.

\textbf{How can individuals submit corrections or updates?}

Individuals can submit corrections and updates through issues or pull requests in the GitHub repository.

\textbf{Is there a versioning system in place?}

The codebase is versioned through GitHub: \url{https://github.com/noahbagz/GenISys}.

\section*{Acknowledgements}
The authors acknowledge the MIT Office of Research Computing and Data for providing high-performance computing resources that contributed to this work.
The authors used OpenAI GPT-5.6, GPT-6, and GPT-6.1 through Codex to help write code used in the experiments and edit the manuscript. All research ideas and findings are those of the authors. The authors reviewed and edited the content and take full responsibility for the publication.

\section*{Conflict of Interest}
The authors declare no conflicts of interest.

\begin{nomenclature}
\entry{$C_{yr}$}{annual total cost (\pounds/year)}
\entry{$D_C,D_{NN}$}{bidirectional Chamfer and mean nearest-neighbor distances}
\entry{$F$}{full feasibility under the implemented engineering constraints}
\entry{$M$}{still-water bending moment (N\,m)}
\entry{$Q_{yr}$}{annual cargo (t/year)}
\entry{$V$}{still-water shear force (N)}
\entry{$Y(c)$}{requested fraction fully feasible with RFR at most $c$}
\entry{$a_t,\bar a_t$}{diffusion signal factor and its cumulative product}
\entry{$b,w,q$}{buoyancy, downward weight, and net upward load per unit length (N/m)}
\entry{$d_\ell,\mathcal I_\ell$}{owned coordinate count and index set for block $\ell$}
\entry{$\mathbf x,\mathbf z$}{physical and transformed design vectors}
\entry{$\alpha$}{feasibility-guidance strength}
\entry{$\beta_t,\widetilde\beta_t$}{forward and reverse diffusion variances}
\entry{$\gamma$}{RFR-guidance strength}
\entry{$\sigma_t$}{reverse-process standard deviation}
\entry{$\chi_{\mathrm{eval}}$}{evaluation-completion indicator}
\entry{DDPM}{denoising diffusion probabilistic model}
\entry{HCDDPM}{hierarchical conditional DDPM}
\entry{SCDDPM}{simultaneous conditional DDPM}
\entry{RFR}{required freight rate}
\end{nomenclature}

\appendix
\counterwithin{figure}{section}
\counterwithin{table}{section}
\counterwithin{algorithm}{section}
\renewcommand{\thefigure}{\Alph{section}\arabic{figure}}
\renewcommand{\thetable}{\Alph{section}\arabic{table}}
\renewcommand{\thealgorithm}{\Alph{section}\arabic{algorithm}}
\onecolumn
\raggedbottom
\section{Dataset and Subsystem Detail}
\label{app_A}
This appendix details the subsystem partition, constraint-violation distribution, and concentrated Hopper-structure failures in the dataset.

\begin{figure}[H]
\centering\includegraphics[width=6in]{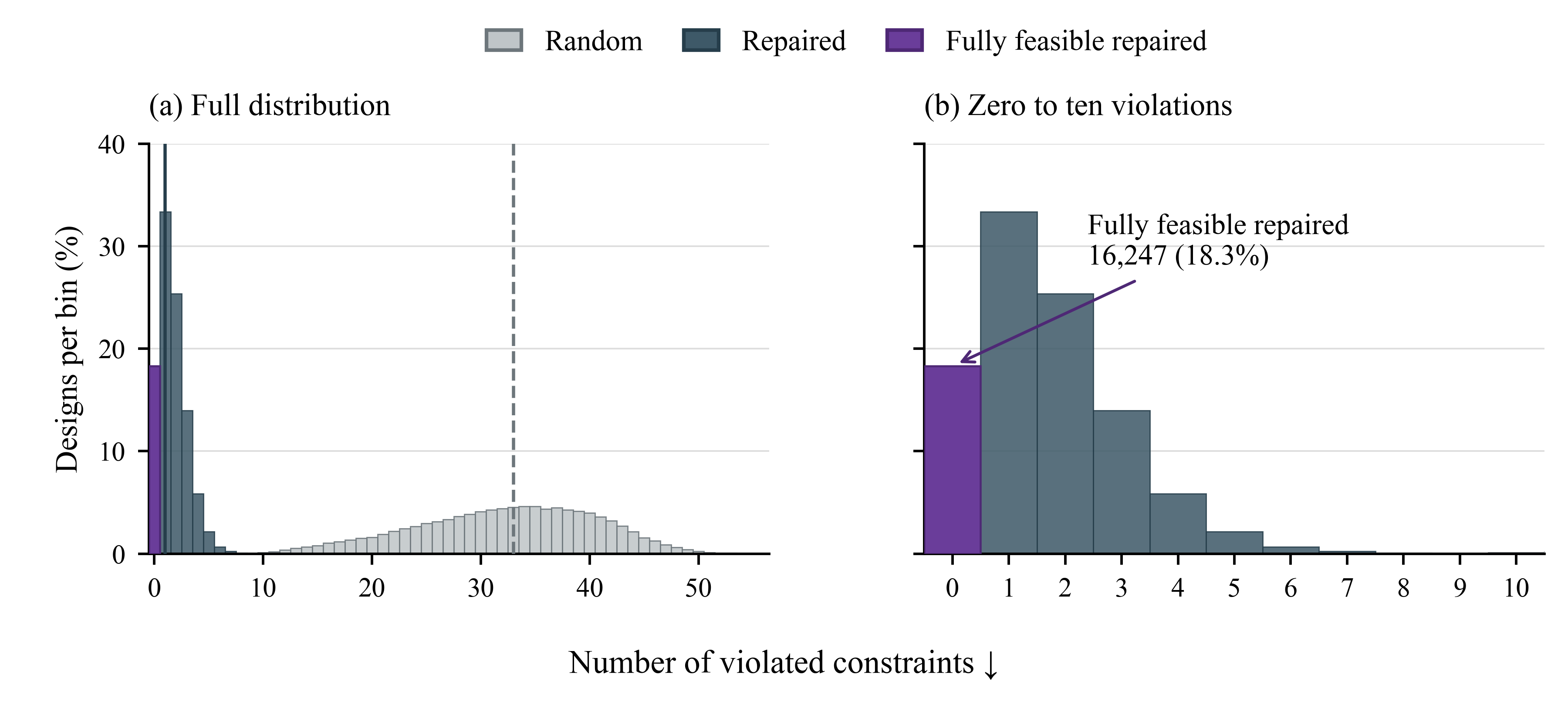}
\caption[Constraint-violation distributions and near-zero detail]{Constraint violations among completed random and repaired designs. (a) Full violation-count distributions; vertical lines mark the population medians. (b) Detail from zero through ten violations; zero indicates full feasibility under the counted checks. Both panels use one-violation bins and percentages of the corresponding completed population, with no renormalization of the close-up. The repaired population was produced from selected random designs with fewer initial violations.}
\label{fig_A1}
\end{figure}

Table~\ref{tab_A1} shows the subsystem partition; Figure~\ref{fig_A1} shows the dataset's constraint violations.

\begin{table}[H]
\caption[Subsystem blocks and local checks]{Distribution of 133 learned design parameters across 20 subsystems. The Checks column lists 162 subsystem checks. An additional 35 checks apply to the complete ship and are not included in this subtotal.}
\label{tab_A1}
\centering\small
\begin{arxivtablefit}
\begin{tabular}{lrr}
\toprule
Subsystem & Controls & Checks\\
\midrule
Hull principal dimensions & 7 & 7\\
Hull midbody & 2 & 8\\
Hull bow & 9 & 14\\
Hull stern & 10 & 12\\
Hull bulbous bow & 6 & 6\\
Hull stern bulb & 8 & 7\\
Powering & 3 & 6\\
Compartment arrangement & 6 & 17\\
Bottom shell and double bottom & 3 & 15\\
Longitudinal bottom structure & 7 & 12\\
Transverse bottom structure & 5 & 8\\
Side shell & 2 & 5\\
Longitudinal side structure & 7 & 2\\
Transverse side structure & 9 & 6\\
Top deck plating & 1 & 5\\
Longitudinal deck structure & 8 & 7\\
Transverse deck structure & 8 & 8\\
Bulkhead structure & 11 & 5\\
Hopper structure & 18 & 10\\
Hatch structure & 3 & 2\\
\midrule
Total & 133 & 162\\
\bottomrule
\end{tabular}
\end{arxivtablefit}
\end{table}

\subsection{Hopper-Structure Failure Concentration}
Of the 61,975 completed repaired training designs, 43,048 (69.5\%) fail the hopper-structure checks. These represent 84.8\% of the 50,780 repaired training designs that remain infeasible. Among 26,955 designs that pass the system-level checks but fail exactly one subsystem, hopper structure is the failing subsystem in 24,009 cases (89.1\%). System-level checks are separate from subsystem checks; full feasibility requires both to pass. The next most frequently failed local label is bulkhead structure, with 5,347 failures (8.6\% of completed repaired training designs); a design may fail more than one subsystem. The hopper block contains 18 independent controls for brackets, plating, and stiffeners, with ten local checks on bracket geometry and thickness and hopper plating (Appendix~\ref{sec_D2}). These counts identify a concentrated source of residual infeasibility in the repaired training population. They do not establish why a particular design failed or whether changing only hopper controls would make the whole design feasible; that question is examined in the locked-subsystem redesign experiment of Sec.~\ref{sec:hopper-redesign}.

\section{Equation-Informed Dataset Repair}
\label{app_B}
This appendix describes the equation-informed repair procedure, including constraint localization, execution order, randomized margins, and bounded searches.

The dataset-augmentation method follows constraint-to-variable localization from adaptive structural-repair research \cite{samanipour2020adaptive,cai2021adaptive,jelovica2024improved,cai2023neuralrepair}. The authors developed the equation-informed repair procedure, extending the approach introduced in MiDShip \cite{bagazinski2026midship}. The repair procedure maps each repairable engineering check to the design parameters in its governing equation or structural definition. When a check fails, the procedure uses this mapping to select parameter changes intended to address the failure. Diagnostic, fixed-context, inactive, and unmapped rows are not treated as repair targets.

Repair is ordered so that upstream changes propagate into downstream calculations. Hull-form and principal-dimension corrections are applied first, followed by subdivision and primary spacings, plate and local-member scantlings, hull-girder sections, tank or system capacity, and departure cargo allocation. Equation~\eqref{eq_B1} gives the schematic passing-side target used by direct lower-bound rules:
\begin{equation}
 y_{new}\geq(1+m)y_{req},
 \qquad m\sim\mathcal{U}(0.25,0.50),
 \label{eq_B1}
\end{equation}
Direct upper-bound rules target a value no greater than $y_{max}/(1+m)$, placing it below the permitted maximum $y_{max}$. For maximum-spacing constraints, the repair increases the member count to achieve this smaller spacing and rounds the count upward to a whole number. Discrete member counts are rounded to an admissible passing value. When more than one approved physical action can address the same failure, the saved repair seed selects the action; independently drawn margins place the proposal beyond, rather than exactly on, the constraint boundary. This stochasticity generates alternative designs.

For each repair alternative, \emph{one pass} means that the complete repaired systems evaluation is not fed back to formulate a second proposal. Constructing each alternative's proposal can still require bounded equation-level searches: coupled hull-domain controls are searched over a finite coordinate set, bracket and member-count choices are discrete, and section repair uses a 61-point depth search followed by bounded flange-width and thickness trials. These internal calculations use the governing local equations rather than repeated calls to the complete ship-system evaluator. Repair-domain limits are recorded separately from the random-sampling bounds. A repair may leave the original sampling interval when required to reach a physically directed passing-side target, so random and repaired populations must not be interpreted as draws from the same proposal distribution.

Each selected source design has three repair alternatives. Each alternative consists of one repair proposal followed by one complete ship-system evaluation. The evaluation result is not used for a second repair pass, so the same single-pass procedure is applied consistently across candidate designs. After the coordinated edits, all derived hull--structure couplings are restored, the structural-element geometry is regenerated, cargo payload and departure allocation are recomputed, and the complete systems evaluator is executed. Consequently, a remaining violation or evaluation failure is preserved as data rather than triggering hidden iterative optimization. Performance objectives are not optimized during repair; any change in cost, power, payload, or other performance is a consequence of the feasibility-directed parameter changes.

\section{Performance Metrics and Calculations}
\label{app_C}

This appendix inventories the engineering responses calculated by the GenISys
system evaluator, including outputs that are retained only for diagnosis and
are not used as generator inputs, classifier labels, or regression-model
targets. The outputs include scalar performance measures, condition-specific
stability and loading responses, structural properties, and longitudinal load
distributions. A value absent
from a partially completed evaluation is not imputed. Design coordinates,
scenario inputs, solver counters, and pass/fail flags are identified below
because they explain a reported response, but they are not themselves
performance measures. The separate constraint catalogue in
Appendix~\ref{sec_D3} records the rule-by-rule criteria.

\subsection{Hull geometry and hydrostatics}

The parametric hull generator gives length, beam, depth, and the submerged
sections used by the hydrostatic integrator. The scalar outputs comprise
waterline length, wetted surface area, displaced volume, longitudinal and
vertical centers of buoyancy, block coefficient, midship coefficient, and
prismatic coefficient. At the evaluated waterline, the integrator also reports
waterplane area, longitudinal center of flotation, and transverse and
longitudinal second moments of the waterplane. Forward and aft perpendiculars
and their separation are reported separately from overall length; the
structural-rule length proxy is not silently replaced by length between
perpendiculars (LBP). Draft, Froude number, speed, Reynolds number, and voyage range
are retained as operating context. These quantities arise from the generated
hull and the project's numerical integration, rather than from Holtrop's
resistance regression. Displacement mass is seawater density times displaced
volume; its difference from modeled loading is retained as a closure metric. The recorded block coefficient is $C_{B,\mathrm{OA}}=\nabla/(L_{\mathrm{OA}}BT)$, using full design beam and operating draft. The same definition of block coefficient is used for the published-vessel estimates in Appendix~\ref{app_K}; coefficients based on LBP or waterline length have a different denominator.

\subsection{Resistance, installed power, and fuel}
\label{sec_C2}

The calm-water resistance implementation follows Holtrop and Mennen's
empirical component method and the later re-analysis
\cite{holtropMennen1982,holtrop1984}. Equation~\eqref{eq_C1} gives the implemented decomposition:
\begin{equation}
 R_T=R_F(1+k_1)+R_{\mathrm{app}}+R_W+R_B+R_A,
 \qquad
 R_F=\tfrac12\rho U^2 S\frac{0.075}{(\log_{10}Re-2)^2},
 \label{eq_C1}
\end{equation}
where $S$ is wetted area, $R_{\mathrm{app}}$ is the assigned appendage term,
$R_W$ is predicted wave resistance, $R_B$ is the conditional bow-bulb term,
and $R_A$ is the model--ship correlation allowance. The form factor $1+k_1$
and wave terms are evaluated from the generated hydrostatic coefficients and
form parameters. The recorded entrance half-angle and bulb transverse area
are intermediate geometry inputs. Appendage area and resistance factor are
\emph{GenISys scenario inputs}, not automatically resolved appendage geometry.
The present implementation assumes no submerged-transom correction in this
sum. Domain checks reject singular or inadmissible equation inputs but do not
turn the regression into a high-fidelity towing-tank or computational fluid dynamics (CFD) prediction.

Equation~\eqref{eq_C2} gives installed main-engine power, voyage duration, and carried fuel:
\begin{equation}
\resizebox{0.90\linewidth}{!}{$\displaystyle
 P_{\mathrm{installed}}=
 \frac{R_T U}{1000 f_{\mathrm{MCR}}\eta_{\mathrm{prop}}}
 \quad[\mathrm{kW}],\qquad
 t_{\mathrm{sea}}=\frac{R_{\mathrm{voyage}}}{24U_{\mathrm{kn}}},
 \qquad
 m_{\mathrm{fuel}}=
 \left(24P_{\mathrm{installed}}s_{\mathrm{main}}
       +1000q_{\mathrm{aux}}\right)(t_{\mathrm{sea}}+t_{\mathrm{reserve}}).
 \label{eq_C2}
$}
\end{equation}
The maximum continuous rating (MCR) fraction, aggregate propulsive efficiency, specific fuel consumption,
auxiliary daily fuel, and reserve duration are explicit scenario assumptions
in Table~\ref{tab_C1}; no propeller--engine matching is solved.
Fuel inventory is intentionally budgeted at full installed main-engine power
throughout the voyage duration plus reserve days. This provides a conservative
onboard-fuel allowance for tank capacity, departure loading, and available
cargo capacity. The MCR fraction sizes installed power; no service-load
reduction is applied to this fuel allowance.
The computed minimum engine-room length uses a linear installed-power
correlation plus a 10\% and 2\,m clearance allowance. The evaluator also
records a preliminary deadweight proxy and a bulk-carrier \emph{required}
energy efficiency design index (EEDI) reference, phase, and reduction factor. Those are regulatory reference
values, not an achieved EEDI or a propulsion-emissions simulation.

\begin{table}[H]
\caption[Powering and fuel scenario]{Powering and fuel assumptions used in the evaluations. These are preliminary model assumptions, rather than selected machinery or calibrated vessel particulars.}
\label{tab_C1}
\centering\small
\begin{arxivtablefit}
\begin{tabular}{llrl}
\toprule
Quantity & Symbol & Value & Unit\\
\midrule
Service-power/MCR fraction & $f_{\mathrm{MCR}}$ & 0.85 & dimensionless\\
Aggregate propulsive efficiency & $\eta_{\mathrm{prop}}$ & 0.637 & dimensionless\\
Main-engine specific fuel consumption & $s_{\mathrm{main}}$ & 0.19 & kg/kWh\\
Auxiliary fuel & $q_{\mathrm{aux}}$ & 0.2 & t/day\\
Fuel reserve & $t_{\mathrm{reserve}}$ & 5 & day\\
Appendage wetted area & $S_{\mathrm{app}}$ & 1,000 & m$^2$\\
Appendage resistance multiplier & $1+k_2$ & 2.0 & dimensionless\\
\bottomrule
\end{tabular}
\end{arxivtablefit}
\end{table}

\subsection{Structural mass, loading, and strength}
\label{sec_C3}

The structural element table and gross plate thicknesses give modeled steel
weight. Outfit, machinery, hatch-cover, optional deckhouse, lightship growth
margin, miscellaneous deadweight, voyage fuel, ballast, and cargo are then
accounted for in the displacement budget. The output records all of these
component masses, deadweight, available cargo mass, modeled lightship mass,
full-cargo displacement, and full-cargo longitudinal, vertical, and transverse
centers of gravity. Hatch-cover count, weight, and projected lateral area are
also recorded. The white deckhouse used only for paper visualization has zero
engineering mass unless a separate explicit deckhouse mass scenario was
provided; the displayed examples use no such scenario.

The frozen confirmation allocator keeps total cargo mass $m_C$ equal to the available cargo mass calculated from displacement, modeled lightship, voyage fuel, and miscellaneous deadweight. For hold $i$, let $V_i$ be its generated volume, $c_i=\rho_CV_i/m_C$ its capacity fraction, and $f_i$ its cargo-mass fraction. Admissible allocations satisfy $0\leq f_i\leq c_i$ and $\sum_i f_i=1$. Here $\rho_C=2400$\,kg/m$^3$; seawater and steel densities are 1025 and 7850\,kg/m$^3$. Nonpositive payload or insufficient combined capacity is retained as an unsuccessful design outcome.

Two limiting cargo allocations are constructed by filling holds from the forward end or from the aft end, using their geometric midpoints to determine the order. Combining these allocations in different proportions preserves total cargo mass and hold-capacity limits. Each trial's cargo distribution and trim are then evaluated using the physical model. A required cargo longitudinal center of gravity (LCG) is estimated from displacement times hull longitudinal center of buoyancy (LCB) minus the noncargo and fuel moments, divided by $m_C$; projection onto the attainable interval supplies a direct trial allocation. The active hybrid solver first evaluates 33 equally spaced points on the extreme-allocation segment, recalculating cargo fill, component distributions, and displacement--moment equilibrium at each point. Adjacent valid trials with opposite trim signs receive up to 48 bisection iterations. Invalid equilibrium trials are excluded from candidate selection. If no valid segment trial remains, the evaluator records an allocation-equilibrium rejection.

If the best segment trial still exceeds the departure trim limit, sequential least-squares programming (SLSQP) varies all hold fractions under the same bounds and sum constraint. Its objective is squared trim divided by the squared trim limit, plus $10^{-6}$ times the variance of relative hold filling $f_i/c_i$. The four starting candidates are the best segment trial, direct LCG allocation, capacity-proportional allocation, and segment midpoint, with duplicates removed. Each solve permits 80 iterations and uses \texttt{ftol}=$10^{-10}$. After enforcing individual hold bounds, a small imbalance in the cargo fractions is corrected by adding cargo to holds with spare capacity or removing cargo from loaded holds. Correction is attempted only when the fractions differ from a total of one by at most $10^{-5}$. The corrected fractions must sum to one within $10^{-9}$ and satisfy hold-capacity bounds within $10^{-9}$. These tolerances accommodate small numerical errors in the solver's cargo fractions. Final selection uses the smallest absolute normalized trim among physically evaluated candidates, breaking ties by relative-fill variance. Optimizer success alone is not an acceptance criterion. The selected allocation is recomputed and receives the normal full-system constraint evaluation.

The cover weight and center of gravity (CG) in this budget come from representative hatch-opening
dimensions and a provisional cover areal-mass assumption, not engineered
per-hold covers. The cargo allocator reports total available hold capacity, total and minimum
hold margins, required and achieved cargo center, and the extrema reachable
from the hold geometry. Saved per-hold records provide start and end stations,
capacity, loaded volume, and fill height. The equilibrium solver reports mean,
forward, and aft drafts, signed and absolute trim, trim tolerance and margin,
waterplane properties, and force and moment residuals. Its iteration counts,
interpolation fraction, search brackets, and success flags describe numerical
behavior rather than independent ship performance.

\begin{figure}[t]
\centering\includegraphics[width=\textwidth]{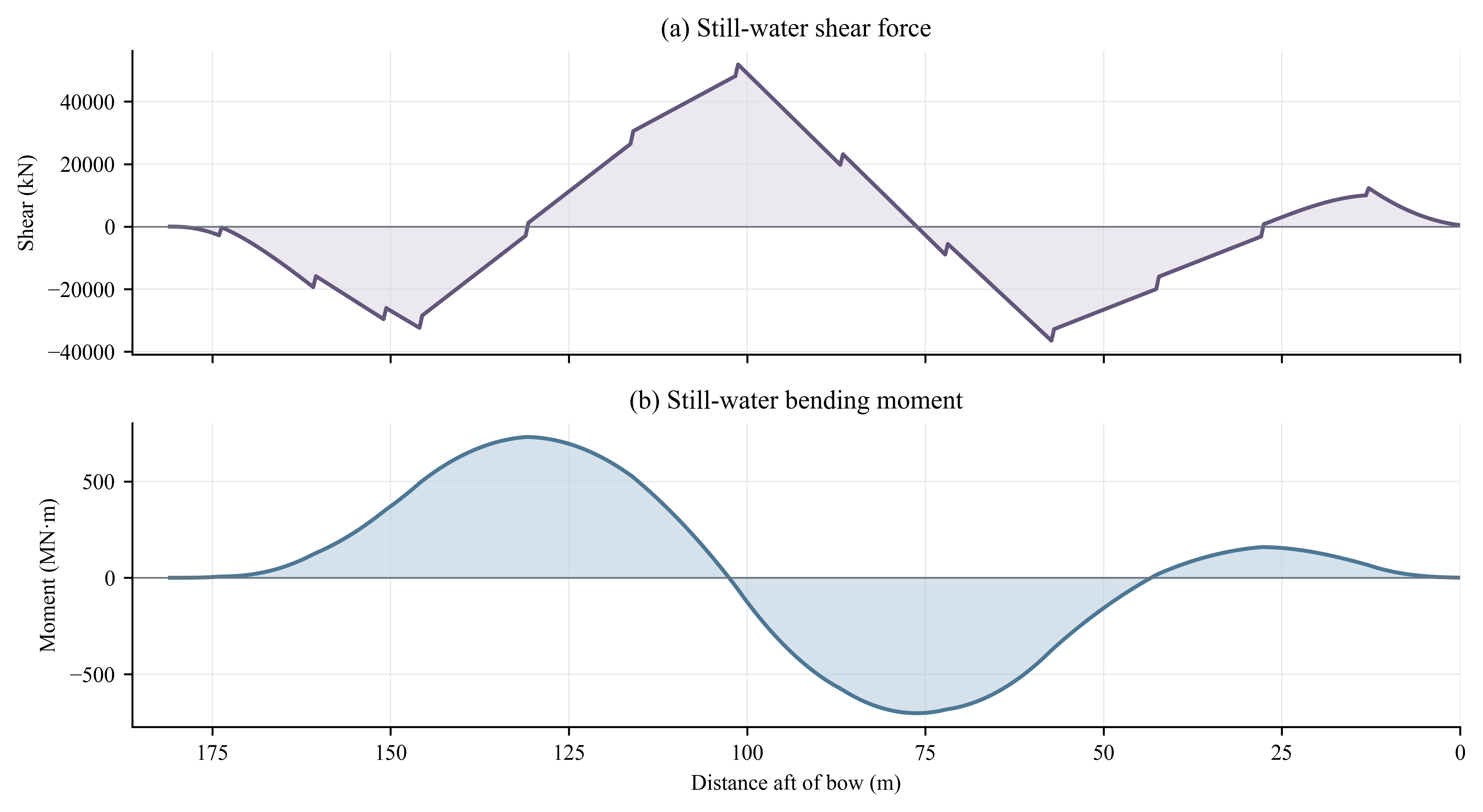}
\caption[Still-water shear and bending response]{Saved, unbinned full-cargo departure response for feasible SCDDPM validation design 80000134: (a) still-water shear force and (b) still-water bending moment. The bow is at the right. The curves are one evaluated example, not population averages or full-length stress results; Figure~\ref{fig_5} shows the corresponding arrangement and distributed loading.}
\label{fig_C1}
\end{figure}

For each modeled loading condition, the net upward load per unit length is
$q=b-w$, where $b$ is buoyancy per unit length and $w$ is downward weight
per unit length. The integration
in Equation~\eqref{eq_2} yields still-water shear and bending arrays. The
compiled record gives positive and negative shear and bending extrema and
their governing condition identifiers, as well as structural load-case count.
The hull-girder strength assessment uses full-cargo departure loads.
An arrival condition is also evaluated using the same cargo and 10\% of
departure fuel, without adding compensating ballast. Arrival calculations
are included in the numerical-consistency checks, but arrival loads do not
govern the hull-girder strength assessment.
The three-hold structural model supplies gross deck and bottom section
properties and member-level rule checks; wave sagging and hogging branches
are combined with still-water response at the same station where the
formulated structural constraints require them \cite{ABS_MVR,IACS_CSR_2025}.
These rule-based constraints assess preliminary elastic strength and scantlings; they do not cover fatigue, buckling,
ultimate strength, or full-ship finite-element predictions. Figure~\ref{fig_C1} shows the saved shear and bending response for the integrated example.

MiDShip documents a related but distinct
three-hold structural dataset. It defines three explicit geometry-derived
study metrics: volume-normalized steel mass, effective longitudinal
cross-sectional area, and a yield-based maximum elastic bending-moment proxy
using 250\,MPa steel yield strength and an author-selected safety factor of two
\cite{bagazinski2026midship}. GenISys reuses the parametric structural
representation and evaluates integrated vessel loading and gross scantling
responses; MiDShip's maximum-moment proxy is not presented as the GenISys
still-water bending \emph{demand}. The MiDShip manuscript also identifies
12,753 released three-hold designs spanning random, stochastic gradient Langevin dynamics (SGLD)-generated, and
repaired subsets, with 25 implemented ABS-derived structural constraints
\cite{bagazinski2026midship}.

\subsection{Upright stability, subdivision, and numerical evidence}

For each named intact condition, the hydrostatic solver retains equilibrium
draft, freeboard-related geometry, displacement and buoyancy centers,
uncorrected transverse and longitudinal metacentric heights, and linear trim.
Corrected initial transverse metacentric height (GM) is emitted only when explicit tank-group free-surface
corrections are available. These corrections are not supplied in the reported evaluations. Consequently, the 187-check configuration does not enforce a minimum initial transverse GM; retained uncorrected heights are diagnostic outputs. Finite-angle righting-arm curves, weather
criteria, and downflooding limits are outside the current evaluated scope.
A coarse one- or two-compartment flooding search can additionally report
sample count, monotonic-boundary observation, and minimum compartment margins.
Because this coarse search is not a complete damage-stability assessment,
its outputs are optional diagnostics and do not contribute to the counted
feasibility checks.

The scalar performance record preserves mass--displacement difference;
normalized volume and first-moment residuals; distributed mass and first
moment residuals; net vertical force and its normalized fraction; net
vertical first moment; and aft shear and bending closure residuals. These
quantities audit the grid integration and equilibrium solution. They should
not be optimized as standalone physical objectives.

\subsection{Acquisition, voyage, and annual economics}

Hart and Vlahopoulos place a conceptual ship-design benchmark within an
integrated multidisciplinary optimization architecture \cite{hart2010IMDO}.
The acquisition, voyage, capital, running, and annual-transport coefficients
below are printed in Yang's earlier ship-design benchmark
\cite{yang2000minimax}. GenISys uses these configurable economic assumptions while replacing the
benchmark's simple mass and power correlations with evaluated GenISys
structural and powering results. With steel,
outfit, cargo, and deadweight (DWT) masses in tonnes and installed power in kW,
Equation~\eqref{eq_C3} defines acquisition cost. Equation~\eqref{eq_C4} relates port time to annual voyages, and Equation~\eqref{eq_C5} gives trip fuel and port costs. Equations~\eqref{eq_C6} and~\eqref{eq_C7} define annual cost components and cargo throughput; Equation~\eqref{eq_C8} gives annual total cost and RFR:
\begin{align}
 C_{\mathrm{ship}}&=1.3\left(
  2000W_{\mathrm{steel}}^{0.85}
  +3500W_{\mathrm{outfit}}
  +2400P_{\mathrm{installed}}^{0.8}\right),\label{eq_C3}\\
 t_{\mathrm{port}}&=2\left(\frac{m_C}{8000}+0.5\right),
 &N_V&=\frac{350}{t_{\mathrm{sea}}+t_{\mathrm{port}}},\label{eq_C4}\\
 C_{\mathrm{fuel,trip}}&=1.05\frac{m_{\mathrm{fuel}}}{1000}(100),
 &C_{\mathrm{port,trip}}&=6.3W_{\mathrm{DWT}}^{0.8},\label{eq_C5}\\
 C_{\mathrm{voy,yr}}&=N_V(C_{\mathrm{fuel,trip}}+C_{\mathrm{port,trip}}),
 &C_{\mathrm{cap,yr}}&=0.2C_{\mathrm{ship}},\label{eq_C6}\\
 C_{\mathrm{run,yr}}&=40000W_{\mathrm{DWT}}^{0.3},
 &Q_{\mathrm{yr}}&=N_Vm_C,\label{eq_C7}\\
 C_{\mathrm{yr}}&=C_{\mathrm{voy,yr}}+C_{\mathrm{cap,yr}}
  +C_{\mathrm{run,yr}},
 &\mathrm{RFR}&=C_{\mathrm{yr}}/Q_{\mathrm{yr}}.
 \label{eq_C8}
\end{align}

Costs are in pounds sterling and RFR in pounds per transported tonne.
Each trip's fuel cost is calculated from the modeled voyage fuel plus reserve
fuel. This differs from Yang's benchmark, which prices only fuel consumed
during days at sea. The compiled economic detail also records steel, outfit,
deadweight and cargo tonnes; port and voyage-cycle days; trips per year;
separate trip fuel and port costs; annual voyage, capital, running, and
total costs; ship acquisition cost; annual cargo; and RFR. Fuel price,
annual days, handling rate, and cost coefficients can be overridden by
an explicit scenario, with the input basis saved beside the result.
No inflation date, financing contract, shipyard quote, or market calibration
is implied by these inherited benchmark coefficients.

\subsection{Complete output-family index}

Table~\ref{tab_C2} maps the full computed response
families to their calculation methods. Evaluation status, numerical solver
diagnostics, and pass/fail outcomes support interpretation but are not
additional physical performance measures. Constraint residuals remain
fully itemized in Appendix~\ref{sec_D3}.

\begingroup
\footnotesize
\setlength{\tabcolsep}{4pt}
\begin{arxivlongtable}
\begin{longtable}{p{0.24\textwidth}p{0.43\textwidth}p{0.25\textwidth}}
\caption[Calculated response families]{Calculated GenISys response families, including diagnostics not used
as training targets.}\label{tab_C2}\\
\hline
Family & Outputs retained & Implementation and scientific basis\\
\hline
\endfirsthead
\hline
Family & Outputs retained & Implementation and scientific basis\\
\hline
\endhead
Hull hydrostatics & Waterline length, wetted area, displacement, buoyancy
centers, block/midship/prismatic coefficients, waterplane area and moments,
flotation center, perpendiculars and LBP &
Hull generator and hydrostatic integrator.\\
Resistance & Reynolds number, friction/form, appendage, wave, bulb, allowance
and total resistance; entrance angle and bulb area intermediates &
Hull design module; Holtrop--Mennen \cite{holtropMennen1982,holtrop1984}.\\
Power and voyage & Effective/installed power, minimum machinery-space
length, days at sea, main and auxiliary fuel demand, reserve fuel &
Hull design module; Equation~\eqref{eq_C2} and explicit project
scenario.\\
Environmental reference & Preliminary deadweight proxy, reference and
required bulk-carrier EEDI proxies, phase and reduction factor &
Powering constraints; regulatory reference only.\\
Mass budget & Steel, outfit, machinery, hatch, optional deckhouse, margin,
miscellaneous deadweight, fuel, ballast, cargo, lightship, total
displacement, and displacement margin &
Weights and system-design modules; element areas and
component accounting.\\
Cargo arrangement & Hold capacities, loaded volume, fill heights, capacity
margins, cargo longitudinal center of gravity (LCG) bounds/target/achievement, full-cargo displacement and CG &
Cargo-hold and system-design modules; compartment and
allocation records.\\
Equilibrium & Mean/forward/aft draft, signed/absolute trim, one-percent
trim margin, waterplane response, force and moment residuals &
Hydrostatic, upright-stability, and system-design solvers.\\
Longitudinal response & Distributed weight/buoyancy/net load, shear and
bending arrays, positive/negative extrema and governing states,
wave-combined structural load envelope &
Weights and structure modules; Equation~\eqref{eq_2}
and cited scantling rules.\\
Structural properties & Gross member masses and section properties,
deck/bottom section moduli, stiffener and plate responses &
Structure module; three-hold geometry and the cited scantling rules
\cite{ABS_MVR,IACS_CSR_2025}.\\
Intact and damaged diagnostics & Condition displacement, GM and trim;
optional floodable-length and compartment-margin diagnostics &
Stability and hydrostatic solvers;
conditional and non-certifying outputs.\\
Economics & Acquisition and trip costs, annual voyage/capital/running/total
cost, annual cargo, required freight rate &
System-design and economic-input modules;
Equation~\eqref{eq_C8}; benchmark basis
\cite{yang2000minimax,hart2010IMDO}.\\
Numerical closure & Mass, distributed-load and first-moment residuals,
force and moment closure, aft shear and bending residuals &
Weights and system-design modules; integration and
solver diagnostics.\\
\hline
\end{longtable}
\end{arxivlongtable}
\endgroup

\subsection{Evaluator Numerical Verification}
\label{sec_C7}
A retrospective verification audit covers all 30,000 confirmation requests at 5,000 nmi: 29,970 completed evaluations and 17,110 fully feasible candidates. Both full-cargo departure and arrival supplied closure diagnostics for every completed candidate, giving 59,940 condition records. Numerical residuals measure how closely the calculated mass, forces, and moments satisfy their balance equations. Each residual is compared with a specified numerical tolerance. All recorded residuals in this audit were finite and within their tolerances. These tolerances assess calculation consistency and are separate from physical design limits. Unavailable results are not counted as passes. The audit retains the recorded feasibility labels and saves per-condition medians and upper quantiles alongside the maxima in Table~\ref{tab_C3}.
\begin{table}[H]
\caption[Confirmation numerical-closure residuals]{Largest absolute normalized named-condition residuals in the confirmation audit. Maxima are taken across departure and arrival, with 29,970 completed and 17,110 feasible records per condition. Every listed residual is finite and within its numerical tolerance. Mass is normalized by condition mass $m$, force/shear by $mg$, and first moment/bending by $mgL$, where $L$ is the integration length and $g$ is gravitational acceleration. These checks assess calculation consistency within the preliminary concept-design model.}
\label{tab_C3}
\centering\small
\begin{arxivtablefit}
\begin{tabular}{lrrr}\toprule
Quantity & Completed max. & Feasible max. & \shortstack{Numerical\\tolerance}\\\midrule
Distributed mass & $6.28\times10^{-16}$ & $6.07\times10^{-16}$ & $1.00\times10^{-8}$\\
Vertical force & $6.15\times10^{-12}$ & $5.61\times10^{-12}$ & $1.00\times10^{-6}$\\
First moment & $1.21\times10^{-11}$ & $1.12\times10^{-11}$ & $1.00\times10^{-6}$\\
Endpoint shear & $6.15\times10^{-12}$ & $5.61\times10^{-12}$ & $1.00\times10^{-6}$\\
Endpoint bending & $1.82\times10^{-11}$ & $1.67\times10^{-11}$ & $1.00\times10^{-6}$\\
\bottomrule\end{tabular}
\end{arxivtablefit}
\end{table}
Among the 17,110 fully feasible candidates, two fuel-capacity margins, three nominal hull-girder shear-stress margins, and one deck section-modulus margin lie within 1\% of their respective thresholds. Counts need not denote distinct designs. These selected-margin diagnostics do not impose new feasibility tolerances or establish regulatory safety margins.
Scalar recomputation checks the recorded displacement--density, installed-power, conservative fuel, and RFR identities. An analytic rectangular-prism benchmark checks simultaneous displacement and longitudinal-moment equilibrium for level, trimmed, and shallow conditions. At 2,001 stations, the largest endpoint-draft error against closed-form values is $1.00\times10^{-7}$ m. The production beam integrator is checked against uniform and linear closed-form loads. On three deterministically selected feasible hulls (median displacement, largest length-to-beam ratio, and smallest draft-to-depth ratio), increasing waterplane resolution from 101 by 1,000 to 401 by 4,000 gives maximum observed changes of 0.0386\% in displacement volume and 0.214\% in wetted area. Independent integration reproduces their saved shear and bending curves; maximum bending discrepancy against exact integration of the same piecewise-linear load is 0.0081\% of peak bending. That comparison checks integration of a fixed saved field; the full reevaluation below separately rebuilds the component loads at finer resolution.
These checks assess numerical consistency and grid sensitivity within the preliminary design calculations; they do not validate predicted vessel performance against measurements. Some closure identities are enforced by mass normalization and equilibrium solving, and three cases do not establish population-wide convergence. The resulting vessels remain concept designs whose full safety and performance require further engineering assessment before construction.

For a nonzero recorded threshold $h_k$ and signed evaluation $e_k$, the diagnostic margin is $-e_k/|h_k|$, positive on the passing side. Each design's cargo-capacity margin is the minimum over its active holds; inactive zero-threshold holds are excluded. Table~\ref{tab_C4} reports the selected physical and structural margins. A normalized margin of 0\% denotes the constraint threshold; positive values indicate the passing side. For a minimum section-modulus requirement, a 100\% margin means the available section modulus is twice the required value, and larger margins are possible. For trim, a margin near 100\% means the forward--aft draft difference is near zero relative to the allowed limit.

\begin{table}[H]
\caption[Selected feasible-design margins]{Selected normalized margins among all 17,110 fully feasible confirmation designs. Every selected margin is finite. Counts within 1\% or 5\% of a threshold are descriptive diagnostics, not new acceptance tolerances.}
\label{tab_C4}
\centering\small
\begin{arxivtablefit}
\begin{tabular}{lrrrr}
\toprule
Criterion & Minimum (\%) & Median (\%) & Within 1\% & Within 5\%\\
\midrule
Minimum active-hold capacity & 13.647 & 39.32 & 0 & 0\\
Fuel-tank capacity & 0.184 & 59.12 & 2 & 7\\
Departure trim limit & 100.000 & 100.00 & 0 & 0\\
Nominal side-shell shear stress & 0.408 & 41.62 & 3 & 23\\
Bottom total-load section modulus & 53.126 & 210.84 & 0 & 0\\
Deck total-load section modulus & 0.947 & 100.29 & 1 & 9\\
\bottomrule
\end{tabular}
\end{arxivtablefit}
\end{table}

\subsubsection{Full Reevaluation at Finer Grids}
The same three concepts were reevaluated once at each of three grid resolutions, retaining their saved design controls and structural-element definitions. Designs 81014494, 81026406, and 81010055 represent median displacement, the largest length-to-beam ratio, and the smallest draft-to-depth ratio, respectively, in the feasible confirmation population. Their run-qualified identities were fixed in the preceding audit before refinement. Each design was evaluated at the three numerical resolutions listed in Table~\ref{tab_C_grid_settings}. Each run repeats the normal cargo allocation, component-mass distribution, buoyancy equilibrium, longitudinal loads, structural constraints, and performance evaluation. Only numerical resolution changes.

\begin{table}[H]
\caption[Numerical grid settings]{Numerical resolutions used in the three-design grid-refinement study. The two calculations use different grid dimensions, identified separately below.}
\label{tab_C_grid_settings}
\centering\footnotesize
\setlength{\tabcolsep}{4pt}
\begin{arxivtablefit}
\begin{tabular}{p{0.22\textwidth}p{0.26\textwidth}ccc}
\toprule
Calculation & Grid dimensions & Baseline & Intermediate & Finest\\
\midrule
Hydrostatics and component loads & Longitudinal stations $\times$ waterlines & $1{,}001\times101$ & $2{,}001\times201$ & $4{,}001\times401$\\
Hull volume and resistance & Waterlines $\times$ nominal points per waterline & $101\times1{,}000$ & $201\times2{,}000$ & $401\times4{,}000$\\
\bottomrule
\end{tabular}
\end{arxivtablefit}
\end{table}

Nominal points per waterline specifies the hull-geometry sampling resolution; it is distinct from the hydrostatic station count.

All nine evaluations completed and satisfied the formulated engineering constraints; no check changed pass/fail status. The baseline responses agreed with the saved values to a maximum relative difference of $8.22\times10^{-15}$. From baseline to the finest grid, maximum absolute changes across the three cases were 0.142\% in cargo mass, 0.229\% in resistance, 0.641\% in peak departure shear, 0.178\% in peak departure bending, and 0.088\% in RFR (Table~\ref{tab_C5}). Considering either finer grid, the largest bending change from baseline was 0.344\%; the largest change from the intermediate to the finest grid was 0.319\%. Changes were not uniformly monotone, so these observations describe sensitivity in three cases without establishing an asymptotic convergence order or a population-wide error bound. They support the numerical consistency of these concept evaluations and do not establish full vessel safety or performance for construction.

\begin{table}[H]
\caption[Full-evaluator grid refinement on three concepts]{Full engineering reevaluation of three previously selected concept designs. Each row rebuilds hydrostatics, cargo allocation, and component loads. $n_x$ and $n_z$ are the hydrostatic station and waterline counts. Peak shear and bending use full-cargo departure; full feasibility requires satisfying the formulated engineering constraints.}
\label{tab_C5}
\centering\small
\begin{arxivtablefit}
\begin{tabular}{lrrrrrr}
\toprule
Design & $n_x\times n_z$ & Cargo (t) & $\max|V|$ (MN) & $\max|M|$ (MN\,m) & RFR (\pounds/t) & Feasible\\
\midrule
81014494 & $1001\times101$ & 13482.83 & 39.299 & 708.700 & 25.337 & Yes\\
81014494 & $2001\times201$ & 13485.43 & 39.499 & 706.265 & 25.341 & Yes\\
81014494 & $4001\times401$ & 13484.24 & 39.551 & 708.284 & 25.351 & Yes\\
\addlinespace
81026406 & $1001\times101$ & 12994.52 & 35.423 & 666.301 & 25.203 & Yes\\
81026406 & $2001\times201$ & 13000.25 & 35.426 & 665.229 & 25.200 & Yes\\
81026406 & $4001\times401$ & 13001.72 & 35.483 & 667.348 & 25.202 & Yes\\
\addlinespace
81010055 & $1001\times101$ & 3313.24 & 18.023 & 448.505 & 79.860 & Yes\\
81010055 & $2001\times201$ & 3317.82 & 18.071 & 447.932 & 79.781 & Yes\\
81010055 & $4001\times401$ & 3317.94 & 18.098 & 447.706 & 79.790 & Yes\\
\addlinespace
\bottomrule
\end{tabular}
\end{arxivtablefit}
\end{table}

\section{Design Parameters, Constraints, and Study Protocols}
\label{app_D}
This appendix provides coordinate reconstruction, the evaluator-parameter schema, subsystem-control partition, engineering-constraint catalogue, and experiment protocols. Observed ranges are not design bounds; recorded rule bases do not constitute independent regulatory verification.

\subsection{Complete Design-Parameter Schema}
\label{sec_D1}
Table~\ref{tab_D1} lists all 189 parameters and their observed minimum and maximum values in the full dataset of 217,999 designs and the 16,247 fully feasible designs, including all training, validation, and test splits. Both sets of ranges use the final parameter values recorded in the consolidated dataset. The full dataset includes incomplete evaluations; recorded parameter values in those rows do not establish feasibility. The table summarizes observed values, not feasibility limits, and combinations of values within these ranges are not necessarily feasible. Initial sampling intervals describe how candidates were drawn; they are separate from the ranges observed after repair and evaluation. The suffix \texttt{torc} records a two-option flange choice: a T-shaped stiffener section (1) or an L-shaped stiffener section (0); \texttt{h}, \texttt{t}, and \texttt{w} denote member height, thickness, and flange width.

Counts and coded choices have no physical units; Table~\ref{tab_D1} marks their units with a dash. A vessel-class indicator is 1 for the selected class and 0 for the others.

The 189 coordinates comprise 45 hull-form, 3 operating, 21 subdivision-and-cargo, and 120 structural parameters: 120 continuous parameters, 43 whole-number parameters, 23 two-option fields, and 3 fixed vessel-class coordinates. Of the 23 two-option fields, 16 are learned controls: 13 T/L stiffener-shape choices and three hull configuration switches (\texttt{bit\_EP\_S}, \texttt{bit\_BB}, and \texttt{bit\_SB}). The remaining two-option fields are fixed, inactive, or copied from another field. Initial sampling bounds describe how dataset candidates were constructed. Generation status describes how each parameter is supplied when a trained model is sampled. These are different classifications. Thirty-one parameters have identical lower and upper initial sampling bounds. For example, \texttt{Beta} has both bounds set to zero and is supplied as a preset during generation; ``project-fixed'' is a separate status used for seven parameters. The status column identifies 133 learned controls, 7 project-fixed values, 2 decoder presets, 23 inactive fields, 2 aliases, 7 derived values, and 15 solver states.

The following parameter reconstruction is used in the confirmation study. Whole-number outputs are rounded to the nearest whole number; a value exactly halfway between two whole numbers is rounded to the even one. Two-option controls are set to 1 for values of at least 0.5 and to 0 otherwise. For stiffener-shape controls, these values select T-shaped and L-shaped sections, respectively. Continuous physical outputs are retained without repair-domain clipping. The inverse quantile transform still limits their marginal range. Unrepresentable controls are retained as unsuccessful requests, with unavailable derived values marked missing.

Project-fixed values are \texttt{cat tanker}=0, \texttt{cat container}=0, \texttt{cat bulkcarrier}=1, \texttt{longitudinal bulkhead bit}=0, \texttt{num longitudinal bulkheads}=0, \texttt{Hatch Openings Bit}=1, and \texttt{l\_overhang}=0.03. The two presets are \texttt{Beta}=0 (constant in the training population) and \texttt{web frame w}=0 (a separate flange inactive for the inner-side-shell configuration). Nonzero flange widths recorded in the dataset do not make that field a learned control. All fields marked inactive are set to zero.

The aliases are \texttt{Bc}=\texttt{Bd} and \texttt{bit\_EP\_T}=\texttt{bit\_EP\_S}. Derived deck girder and intermediate-stiffener counts copy their corresponding bottom counts. Equation~\eqref{eq_D1} gives the remaining dimensional reconstruction, where $f_o=0.03$ is the structural overhang fraction at each end and the end-region fractions $f_q$ are those in Equation~\eqref{eq_1}:
\begin{equation}
\begin{aligned}
 B&=L_{\mathrm{OA}}B_d,\qquad D=L_{\mathrm{OA}}D_d,\\
 T&=f_TD,\qquad R_b=\tfrac12BR_c,\\
 L_{3h}&=\frac{3L_{\mathrm{OA}}(1-\sum_{q\in\mathcal E}f_q)}{N_H(1-2f_o)}.
\end{aligned}
\label{eq_D1}
\end{equation}
Here \texttt{Bd}, \texttt{Dd}, and \texttt{Rc} supply $B_d$, $D_d$, and $R_c$; $L_{3h}$ includes the two structural overhangs. The frozen decoder disables the optional bilge-fit adjustment. Cargo fractions are initialized to zero and supplied by the evaluator for active holds. Subsequent MiDShip structural cleanup can alter constructed geometry; raw model controls, decoded vectors, and evaluated geometry are therefore distinct. Feasibility and performance refer to the evaluated state.
\begingroup
\footnotesize
\setlength{\tabcolsep}{1.5pt}
\begin{arxivlongtable}
\begin{longtable}{r p{0.12\textwidth} p{0.19\textwidth} p{0.095\textwidth} l l rrrr}
\caption[Parameter schema and observed dataset ranges]{Complete 189-parameter schema and observed ranges in all 217,999 designs and the 16,247 fully feasible designs, using the final consolidated parameter data across all dataset splits. Continuous values are rounded to six significant digits. These ranges summarize observed values, not feasibility limits; combinations of values within the ranges are not necessarily feasible.}\label{tab_D1}\\
\hline
 & & & & & & \multicolumn{2}{c}{All designs (217,999)} & \multicolumn{2}{c}{Fully feasible (16,247)}\\
Index & Class & Parameter & Units & Type & Status & Minimum & Maximum & Minimum & Maximum\\
\hline
\endfirsthead
\multicolumn{10}{c}{Table \thetable\ continued}\\
\hline
 & & & & & & \multicolumn{2}{c}{All designs (217,999)} & \multicolumn{2}{c}{Fully feasible (16,247)}\\
Index & Class & Parameter & Units & Type & Status & Minimum & Maximum & Minimum & Maximum\\
\hline
\endhead
\hline
\multicolumn{10}{r}{Continued on next page}\\
\endfoot
\hline
\endlastfoot
0 & Hull & LOA & m & float & Learned & 100.459 & 349.944 & 100.459 & 314.583\\
1 & Hull & Lb & -- & float & Learned & 0.0750004 & 0.199993 & 0.0750566 & 0.199912\\
2 & Hull & Ls & -- & float & Learned & 0.0500068 & 0.199997 & 0.0500184 & 0.199994\\
3 & Hull & Bd & -- & float & Learned & 0.100436 & 0.18 & 0.101449 & 0.179967\\
4 & Hull & Dd & -- & float & Learned & 0.0750002 & 0.12796 & 0.0750047 & 0.127794\\
5 & Hull & Bs & -- & float & Learned & 4.47851e-05 & 1 & 0.000242636 & 1\\
6 & Hull & WL & -- & float & Learned & 0.200679 & 0.635049 & 0.203932 & 0.630204\\
7 & Hull & Bc & -- & float & Alias & 0.100436 & 0.18 & 0.101449 & 0.179967\\
8 & Hull & Beta & -- & float & Preset & 0 & 0 & 0 & 0\\
9 & Hull & Rc & -- & float & Learned & 0.1 & 0.134704 & 0.1 & 0.133326\\
10 & Hull & Rk & -- & float & Learned & 9.29026e-05 & 0.999992 & 9.29026e-05 & 0.999652\\
11 & Hull & Bow\_A & -- & float & Learned & -0.499974 & 0.499964 & -0.499161 & 0.499689\\
12 & Hull & Bow\_B & -- & float & Learned & -0.499852 & 0.499963 & -0.498849 & 0.499924\\
13 & Hull & Bk & -- & float & Learned & 5.53125e-06 & 0.099998 & 6.85184e-06 & 0.0999979\\
14 & Hull & Kappa\_Bow & -- & float & Learned & 1.02203e-05 & 0.249994 & 1.02203e-05 & 0.249884\\
15 & Hull & Delta\_Bow\_A & -- & float & Learned & -0.499977 & 0.499886 & -0.499939 & 0.49979\\
16 & Hull & Delta\_Bow\_B & -- & float & Learned & -0.999912 & 0.999994 & -0.999912 & 0.999785\\
17 & Hull & Drift\_A & -- & float & Learned & -0.99976 & 0.999915 & -0.99976 & 0.998505\\
18 & Hull & Drift\_B & -- & float & Learned & -0.999907 & 0.99949 & -0.998302 & 0.999407\\
19 & Hull & Drift\_C & degrees & float & Learned & 0.000514722 & 44.9987 & 0.0426837 & 44.9897\\
20 & Hull & bit\_EP\_S & -- & bool & Learned & 0 & 1 & 0 & 1\\
21 & Hull & bit\_EP\_T & -- & bool & Alias & 0 & 1 & 0 & 1\\
22 & Hull & Trans\_A & -- & float & Learned & -0.249985 & 0.249944 & -0.249907 & 0.249789\\
23 & Hull & SK & -- & float & Learned & 0.100063 & 0.66997 & 0.10026 & 0.66983\\
24 & Hull & Kappa\_Stern & -- & float & Learned & 0.500081 & 0.999952 & 0.500129 & 0.999338\\
25 & Hull & Delta\_Stern\_A & -- & float & Learned & -0.499944 & 0.499993 & -0.498648 & 0.499544\\
26 & Hull & Delta\_Stern\_B & -- & float & Learned & -0.999949 & 0.999955 & -0.999189 & 0.998877\\
27 & Hull & Beta\_Trans & degrees & float & Learned & 0.000206483 & 44.9927 & 0.00908157 & 44.9464\\
28 & Hull & Bc\_Trans & -- & float & Learned & 0.0500032 & 0.17926 & 0.0500446 & 0.17926\\
29 & Hull & Rc\_Trans & -- & float & Learned & 2.49068e-05 & 0.499998 & 3.50882e-05 & 0.49986\\
30 & Hull & Rk\_Trans & -- & float & Learned & -0.99863 & 0.999956 & -0.977055 & 0.999956\\
31 & Hull & bit\_BB & -- & bool & Learned & 0 & 1 & 0 & 1\\
32 & Hull & bit\_SB & -- & bool & Learned & 0 & 1 & 0 & 1\\
33 & Hull & Lbb & -- & float & Learned & 9.38095e-06 & 0.0999917 & 1.11284e-05 & 0.0999917\\
34 & Hull & Hbb & -- & float & Learned & 6.50415e-06 & 0.999989 & 6.50415e-06 & 0.999429\\
35 & Hull & Bbb & -- & float & Learned & 0.100027 & 0.5 & 0.100096 & 0.499882\\
36 & Hull & Lbbm & -- & float & Learned & -0.999906 & 0.999803 & -0.997078 & 0.999355\\
37 & Hull & Rbb & -- & float & Learned & 0.200012 & 0.799929 & 0.200018 & 0.799874\\
38 & Hull & Kappa\_SB & -- & float & Learned & 0.00017174 & 0.999982 & 0.000669505 & 0.999278\\
39 & Hull & Lsb & -- & float & Learned & 2.673e-06 & 0.1 & 3.60889e-05 & 0.0999583\\
40 & Hull & HSBOA & -- & float & Learned & 0.65001 & 0.999946 & 0.650057 & 0.999885\\
41 & Hull & Hsb & -- & float & Learned & 9.29314e-05 & 0.999834 & 0.00055755 & 0.999211\\
42 & Hull & Bsb & -- & float & Learned & 0.100018 & 0.499997 & 0.100067 & 0.499845\\
43 & Hull & Lsbm & -- & float & Learned & 1.86692e-05 & 0.999915 & 5.7017e-05 & 0.999211\\
44 & Hull & Rsb & -- & float & Learned & 0.200037 & 0.799985 & 0.200261 & 0.799862\\
45 & Powering & Draft\_frac & -- & float & Learned & 0.250004 & 0.7 & 0.261949 & 0.7\\
46 & Powering & Fn\_ship & -- & float & Learned & 0.05 & 0.349999 & 0.0515972 & 0.284476\\
47 & Powering & Voyage\_Range & nautical miles & float & Learned & 100 & 9999.95 & 100 & 9998.89\\
48 & Bulkhead\_Spacing & frac\_CrashBlkhd & -- & float & Learned & 0.0500007 & 0.0799981 & 0.0500034 & 0.0799818\\
49 & Bulkhead\_Spacing & frac\_AftPeakBlkhd & -- & float & Learned & 0.05 & 0.0799994 & 0.0500163 & 0.0799886\\
50 & Bulkhead\_Spacing & frac\_lazarette\_L & -- & float & Learned & 0.0500024 & 0.0999979 & 0.0500024 & 0.0999545\\
51 & Bulkhead\_Spacing & frac\_engineRoom\_L & -- & float & Learned & 0.0300002 & 0.3 & 0.0301214 & 0.3\\
52 & Bulkhead\_Spacing & frac\_fuel\_tank\_L & -- & float & Learned & 0.0100046 & 0.0499988 & 0.0100121 & 0.0499642\\
53 & Bulkhead\_Spacing & Num\_Cargo\_Holds & -- & integer & Learned & 6 & 15 & 6 & 15\\
54 & Bulkhead\_Spacing & frac\_Cargo\_Hold\_0 & -- & float & Solver & 0 & 1 & 0 & 1\\
55 & Bulkhead\_Spacing & frac\_Cargo\_Hold\_1 & -- & float & Solver & 0 & 0.984909 & 0 & 0.788855\\
56 & Bulkhead\_Spacing & frac\_Cargo\_Hold\_2 & -- & float & Solver & 0 & 0.447895 & 0 & 0.412865\\
57 & Bulkhead\_Spacing & frac\_Cargo\_Hold\_3 & -- & float & Solver & 0 & 0.335058 & 0 & 0.245828\\
58 & Bulkhead\_Spacing & frac\_Cargo\_Hold\_4 & -- & float & Solver & 0 & 0.50308 & 0 & 0.4869\\
59 & Bulkhead\_Spacing & frac\_Cargo\_Hold\_5 & -- & float & Solver & 0 & 1 & 0 & 1\\
60 & Bulkhead\_Spacing & frac\_Cargo\_Hold\_6 & -- & float & Solver & 0 & 1 & 0 & 0.978223\\
61 & Bulkhead\_Spacing & frac\_Cargo\_Hold\_7 & -- & float & Solver & 0 & 1 & 0 & 1\\
62 & Bulkhead\_Spacing & frac\_Cargo\_Hold\_8 & -- & float & Solver & 0 & 1 & 0 & 0.973816\\
63 & Bulkhead\_Spacing & frac\_Cargo\_Hold\_9 & -- & float & Solver & 0 & 1 & 0 & 0.912001\\
64 & Bulkhead\_Spacing & frac\_Cargo\_Hold\_10 & -- & float & Solver & 0 & 1 & 0 & 1\\
65 & Bulkhead\_Spacing & frac\_Cargo\_Hold\_11 & -- & float & Solver & 0 & 1 & 0 & 1\\
66 & Bulkhead\_Spacing & frac\_Cargo\_Hold\_12 & -- & float & Solver & 0 & 1 & 0 & 1\\
67 & Bulkhead\_Spacing & frac\_Cargo\_Hold\_13 & -- & float & Solver & 0 & 1 & 0 & 1\\
68 & Bulkhead\_Spacing & frac\_Cargo\_Hold\_14 & -- & float & Solver & 0 & 1 & 0 & 0.978486\\
69 & Structure & L\_3h & m & float & Derived & 18.8899 & 69.9974 & 21.5321 & 69.6429\\
70 & Structure & B & m & float & Derived & 15.2686 & 59.6205 & 15.4114 & 50.4922\\
71 & Structure & T & m & float & Derived & 2.57693 & 26.2827 & 3.22131 & 18.902\\
72 & Structure & D & m & float & Derived & 10.048 & 39.6552 & 10.048 & 32.7658\\
73 & Structure & R\_b & m & float & Derived & 0.793211 & 3.45024 & 0.793211 & 2.97142\\
74 & Structure & l\_overhang & fraction of L\_3h & float & Fixed & 0.03 & 0.03 & 0.03 & 0.03\\
75 & Structure & Db & mm & float & Learned & 807.397 & 4186.72 & 1032.25 & 3575.7\\
76 & Structure & bottom shell thickness & mm & integer & Learned & 12 & 120 & 12 & 120\\
77 & Structure & side shell thickness & mm & integer & Learned & 9 & 120 & 12 & 120\\
78 & Structure & Top Deck Thickness & mm & integer & Learned & 10 & 100 & 22 & 100\\
79 & Structure & Inner Bottom Thickness & mm & integer & Learned & 9 & 80 & 27 & 80\\
80 & Structure & Trans Bulkhead Thickness & mm & integer & Learned & 10 & 78 & 10 & 61\\
81 & Structure & Inner Side Shell Thickness & mm & integer & Learned & 10 & 80 & 11 & 80\\
82 & Structure & Sheer strake thickness & mm & integer & Learned & 10 & 80 & 10 & 80\\
83 & Structure & Sheer strake height & mm & float & Learned & 1000.59 & 2499.97 & 1003.45 & 2496.64\\
84 & Structure & web frames per hold & -- & integer & Learned & 3 & 16 & 3 & 14\\
85 & Structure & num bottom girders & -- & integer & Learned & 3 & 16 & 3 & 14\\
86 & Structure & num intermediate bottom stiffeners & -- & integer & Learned & 3 & 12 & 3 & 12\\
87 & Structure & num top girders & -- & integer & Derived & 3 & 16 & 3 & 14\\
88 & Structure & num intermediate deck stiffeners & -- & integer & Derived & 3 & 12 & 3 & 12\\
89 & Structure & num side shell stiffeners & -- & integer & Learned & 15 & 54 & 15 & 52\\
90 & Structure & num vertical bulkhead stiffeners & -- & integer & Learned & 12 & 57 & 12 & 57\\
91 & Structure & num transverse bulkhead stiffeners & -- & integer & Learned & 12 & 52 & 15 & 52\\
92 & Structure & bottom girder thickness & mm & float & Learned & 6 & 22.5 & 14.6021 & 22.4992\\
93 & Structure & intermediate bottom stiffener h & mm & float & Learned & 120.01 & 6425.22 & 123.314 & 3242.18\\
94 & Structure & intermediate bottom stiffener t & mm & integer & Learned & 7 & 24 & 7 & 22\\
95 & Structure & intermediate bottom stiffener w & mm & float & Learned & 14.0312 & 521.314 & 14.0312 & 393\\
96 & Structure & intermediate bottom stiffener torc & -- & bool & Learned & 0 & 1 & 0 & 1\\
97 & Structure & trans deck beam h & mm & float & Learned & 300.058 & 2404.11 & 301.353 & 1558.09\\
98 & Structure & trans deck beam t & mm & integer & Learned & 10 & 25 & 15 & 25\\
99 & Structure & trans deck beam w & mm & float & Learned & 59.6589 & 472.562 & 101.112 & 437.076\\
100 & Structure & trans deck beam torc & -- & bool & Learned & 0 & 1 & 0 & 1\\
101 & Structure & deck girder h & mm & float & Learned & 500.492 & 2913.24 & 825.783 & 2809.37\\
102 & Structure & deck girder t & mm & integer & Learned & 10 & 25 & 14 & 25\\
103 & Structure & deck girder w & mm & float & Learned & 92.4512 & 663.298 & 92.4512 & 658.499\\
104 & Structure & deck girder torc & -- & bool & Learned & 0 & 1 & 0 & 1\\
105 & Structure & deck stiff h & mm & float & Learned & 120.148 & 10000 & 120.148 & 7381.21\\
106 & Structure & deck stiff t & mm & integer & Learned & 7 & 21 & 7 & 16\\
107 & Structure & deck stiff w & mm & float & Learned & 14.0562 & 440.817 & 14.077 & 334.025\\
108 & Structure & deck stiff torc & -- & bool & Learned & 0 & 1 & 0 & 1\\
109 & Structure & blkhd trans stiff h & mm & float & Learned & 120.013 & 10000 & 929.729 & 10000\\
110 & Structure & blkhd trans stiff t & mm & integer & Learned & 7 & 24 & 7 & 23\\
111 & Structure & blkhd trans stiff w & mm & float & Learned & 14.0035 & 2296.66 & 187.544 & 1263.37\\
112 & Structure & blkhd trans stiff torc & -- & bool & Learned & 0 & 1 & 0 & 1\\
113 & Structure & blkhd vert stiff h & mm & float & Learned & 120.072 & 10000 & 449.121 & 7397.81\\
114 & Structure & blkhd vert stiff t & mm & integer & Learned & 7 & 24 & 7 & 24\\
115 & Structure & blkhd vert stiff w & mm & float & Learned & 14.0184 & 1357.21 & 16.2785 & 501.883\\
116 & Structure & blkhd vert stiff torc & -- & bool & Learned & 0 & 1 & 0 & 1\\
117 & Structure & web frame h & mm & float & Learned & 1000.26 & 10000 & 1250.29 & 8151.64\\
118 & Structure & web frame t & mm & integer & Learned & 10 & 37 & 14 & 30\\
119 & Structure & web frame w & mm & float & Preset & 0 & 300 & 0 & 300\\
120 & Structure & web frame torc & -- & bool & Learned & 0 & 1 & 0 & 1\\
121 & Structure & floor t & mm & integer & Learned & 10 & 30 & 15 & 30\\
122 & Structure & side shell Long stiff h & mm & float & Learned & 120.069 & 1013.02 & 152.263 & 939.335\\
123 & Structure & side shell long stiff t & mm & integer & Learned & 7 & 19 & 7 & 12\\
124 & Structure & side shell long stiff w & mm & float & Learned & 14.0843 & 667.951 & 14.0843 & 335.04\\
125 & Structure & side shell long stiff torc & -- & bool & Learned & 0 & 1 & 0 & 1\\
126 & Structure & cat tanker & -- & 0/1 indicator & Fixed & 0 & 0 & 0 & 0\\
127 & Structure & cat container & -- & 0/1 indicator & Fixed & 0 & 0 & 0 & 0\\
128 & Structure & cat bulkcarrier & -- & 0/1 indicator & Fixed & 1 & 1 & 1 & 1\\
129 & Structure & floor web bracket L1 & mm & float & Learned & 3000.15 & 4587.39 & 3002.97 & 4587.39\\
130 & Structure & floor web bracket L2 & mm & float & Learned & 2831.62 & 5627.42 & 2840.91 & 4864.02\\
131 & Structure & floor web bracket t & mm & float & Learned & 10 & 37.4896 & 14 & 37.4884\\
132 & Structure & deck web bracket L1 & mm & float & Learned & -5138.89 & 4577.83 & 387.879 & 4253.62\\
133 & Structure & deck web bracket L2 & mm & float & Learned & 397.854 & 5445.44 & 422.067 & 4780.49\\
134 & Structure & deck web bracket t & mm & integer & Learned & 10 & 38 & 14 & 38\\
135 & Structure & cont step height & mm & float & Inactive & 0 & 0 & 0 & 0\\
136 & Structure & cont step width & mm & float & Inactive & 0 & 0 & 0 & 0\\
137 & Structure & cont step deck plate thickness & mm & float & Inactive & 0 & 0 & 0 & 0\\
138 & Structure & cont step side plate thickness & mm & float & Inactive & 0 & 0 & 0 & 0\\
139 & Structure & num cont step deck stiff & -- & integer & Inactive & 0 & 0 & 0 & 0\\
140 & Structure & cont step deck stiff h & mm & float & Inactive & 0 & 0 & 0 & 0\\
141 & Structure & cont step deck stiff t & mm & integer & Inactive & 0 & 0 & 0 & 0\\
142 & Structure & cont step deck stiff w & mm & float & Inactive & 0 & 0 & 0 & 0\\
143 & Structure & cont step deck stiff torc & -- & bool & Inactive & 0 & 0 & 0 & 0\\
144 & Structure & num cont step side stiff & -- & integer & Inactive & 0 & 0 & 0 & 0\\
145 & Structure & cont step side stiff h & mm & float & Inactive & 0 & 0 & 0 & 0\\
146 & Structure & cont step side stiff t & mm & integer & Inactive & 0 & 0 & 0 & 0\\
147 & Structure & cont step side stiff w & mm & float & Inactive & 0 & 0 & 0 & 0\\
148 & Structure & cont step side stiff torc & -- & bool & Inactive & 0 & 0 & 0 & 0\\
149 & Structure & bulk car bottom hopper t & mm & integer & Learned & 15 & 120 & 27 & 120\\
150 & Structure & bulk car top hopper t & mm & integer & Learned & 15 & 150 & 23 & 150\\
151 & Structure & num bulk car bottom hopper stiff & -- & integer & Learned & 3 & 6 & 3 & 6\\
152 & Structure & bulk car bottom hopper stiff h & mm & float & Learned & 120.016 & 319.981 & 120.058 & 319.926\\
153 & Structure & bulk car bottom hopper stiff t & mm & float & Learned & 7 & 12 & 7 & 12\\
154 & Structure & bulk car bottom hopper stiff w & mm & float & Learned & 14.0152 & 199.982 & 14.1553 & 199.945\\
155 & Structure & bulk car bottom hopper stiff torc & -- & bool & Learned & 0 & 1 & 0 & 1\\
156 & Structure & num bulk car top hopper stiff & -- & integer & Learned & 3 & 6 & 3 & 6\\
157 & Structure & bulk car top hopper stiff h & mm & float & Learned & 120.083 & 319.991 & 120.162 & 319.822\\
158 & Structure & bulk car top hopper stiff t & mm & integer & Learned & 7 & 12 & 7 & 12\\
159 & Structure & bulk car top hopper stiff w & mm & float & Learned & 14.0146 & 199.896 & 14.1842 & 199.447\\
160 & Structure & bulk car top hopper stiff torc & -- & bool & Learned & 0 & 1 & 0 & 1\\
161 & Structure & num intermediate trans frames & -- & integer & Learned & 2 & 9 & 2 & 9\\
162 & Structure & intermediate bottom frame h & mm & float & Learned & 120.046 & 1153.28 & 120.054 & 1153.28\\
163 & Structure & intermediate bottom frame t & mm & integer & Learned & 7 & 20 & 7 & 20\\
164 & Structure & intermediate bottom frame w & mm & float & Learned & 14.0455 & 432.388 & 14.0455 & 403.352\\
165 & Structure & intermediate bottom frame torc & -- & bool & Learned & 0 & 1 & 0 & 1\\
166 & Structure & intermediate side frame h & mm & float & Learned & 120.005 & 3643.61 & 120.222 & 3643.61\\
167 & Structure & intermediate side frame t & mm & integer & Learned & 7 & 24 & 7 & 23\\
168 & Structure & intermediate side frame w & mm & float & Learned & 14.0218 & 553.648 & 14.157 & 476.658\\
169 & Structure & intermediate side frame torc & -- & bool & Learned & 0 & 1 & 0 & 1\\
170 & Structure & intermediate deck frame h & mm & float & Learned & 120.008 & 400.928 & 120.035 & 377.149\\
171 & Structure & intermediate deck frame t & mm & integer & Learned & 7 & 12 & 7 & 12\\
172 & Structure & intermediate deck frame w & mm & float & Learned & 14.0694 & 233.661 & 14.2044 & 199.726\\
173 & Structure & intermediate deck frame torc & -- & bool & Learned & 0 & 1 & 0 & 1\\
174 & Structure & longitudinal bulkhead bit & -- & bool & Fixed & 0 & 0 & 0 & 0\\
175 & Structure & num longitudinal bulkheads & -- & integer & Fixed & 0 & 0 & 0 & 0\\
176 & Structure & long blkhd plate t & mm & integer & Inactive & 0 & 0 & 0 & 0\\
177 & Structure & long blkhd long stiff h & mm & float & Inactive & 0 & 0 & 0 & 0\\
178 & Structure & long blkhd long stiff t & mm & integer & Inactive & 0 & 0 & 0 & 0\\
179 & Structure & long blkhd long stiff w & mm & float & Inactive & 0 & 0 & 0 & 0\\
180 & Structure & long blkhd long stiff torc & -- & bool & Inactive & 0 & 0 & 0 & 0\\
181 & Structure & long blkhd vert stiff h & mm & float & Inactive & 0 & 0 & 0 & 0\\
182 & Structure & long blkhd vert stiff t & mm & integer & Inactive & 0 & 0 & 0 & 0\\
183 & Structure & long blkhd vert stiff w & mm & float & Inactive & 0 & 0 & 0 & 0\\
184 & Structure & long blkhd vert stiff torc & -- & bool & Inactive & 0 & 0 & 0 & 0\\
185 & Structure & Hatch Openings Bit & -- & bool & Fixed & 1 & 1 & 1 & 1\\
186 & Structure & Hatch Opening L & fraction of L\_3h & float & Learned & 0.166685 & 0.45 & 0.167401 & 0.410024\\
187 & Structure & Hatch Opening W & fraction of B & float & Learned & 0.250018 & 0.858638 & 0.370834 & 0.794444\\
188 & Structure & Hatch Opening R & mm & float & Learned & 25.0957 & 899.997 & 750.008 & 899.995\\
\end{longtable}
\end{arxivlongtable}
\endgroup

\subsection{Subsystem Controls and Local Checks}
\label{sec_D2}
Table~\ref{tab_D2} shows the assignment of 133 learned controls to the 20 blocks used by HCDDPM and SCDDPM. Each block receives its own local feasibility label from the check IDs listed below. The complete check names and recorded bases appear in Table~\ref{tab_D3}. System-scope checks act on the assembled design and are not owned by a single block.
\begingroup
\footnotesize
\setlength{\tabcolsep}{1.5pt}
\begin{arxivlongtable}
\begin{longtable}{r p{0.18\textwidth} p{0.49\textwidth} r p{0.13\textwidth}}
\caption[Complete subsystem controls and checks]{Frozen 20-subsystem model partition. Every learned control appears once. The last column identifies local checks in Table~\ref{tab_D3}; 35 further checks have system scope.}\label{tab_D2}\\
\hline
Order & Subsystem & Learned controls & Count & Local checks\\
\hline
\endfirsthead
\multicolumn{5}{c}{Table \thetable\ continued}\\
\hline
Order & Subsystem & Learned controls & Count & Local checks\\
\hline
\endhead
\hline
\multicolumn{5}{r}{Continued on next page}\\
\endfoot
\hline
\endlastfoot
1 & Hull Principal & LOA; Lb; Ls; Bd; Dd; Bs; WL & 7 & C078--C084 (7)\\
2 & Hull Midbody & Rc; Rk & 2 & C070--C077 (8)\\
3 & Hull Bow & Bow\_\allowbreak{}A; Bow\_\allowbreak{}B; Bk; Kappa\_\allowbreak{}Bow; Delta\_\allowbreak{}Bow\_\allowbreak{}A; Delta\_\allowbreak{}Bow\_\allowbreak{}B; Drift\_\allowbreak{}A; Drift\_\allowbreak{}B; Drift\_\allowbreak{}C & 9 & C050--C063 (14)\\
4 & Hull Stern & bit\_\allowbreak{}EP\_\allowbreak{}S; Trans\_\allowbreak{}A; SK; Kappa\_\allowbreak{}Stern; Delta\_\allowbreak{}Stern\_\allowbreak{}A; Delta\_\allowbreak{}Stern\_\allowbreak{}B; Beta\_\allowbreak{}Trans; Bc\_\allowbreak{}Trans; Rc\_\allowbreak{}Trans; Rk\_\allowbreak{}Trans & 10 & C085--C096 (12)\\
5 & Hull Bulbous Bow & bit\_\allowbreak{}BB; Lbb; Hbb; Bbb; Lbbm; Rbb & 6 & C064--C069 (6)\\
6 & Hull Stern Bulb & bit\_\allowbreak{}SB; Kappa\_\allowbreak{}SB; Lsb; HSBOA; Hsb; Bsb; Lsbm; Rsb & 8 & C097--C103 (7)\\
7 & Powering & Draft\_\allowbreak{}frac; Fn\_\allowbreak{}ship; Voyage\_\allowbreak{}Range & 3 & C104--C109 (6)\\
8 & Compartment Arrangement & frac\_\allowbreak{}CrashBlkhd; frac\_\allowbreak{}AftPeakBlkhd; frac\_\allowbreak{}lazarette\_\allowbreak{}L; frac\_\allowbreak{}engineRoom\_\allowbreak{}L; frac\_\allowbreak{}fuel\_\allowbreak{}tank\_\allowbreak{}L; Num\_\allowbreak{}Cargo\_\allowbreak{}Holds & 6 & C021--C037 (17)\\
9 & Bottom Shell and Double Bottom & Db; bottom shell thickness; Inner Bottom Thickness & 3 & C001--C015 (15)\\
10 & Structure Longitudinal Bottom & num bottom girders; num intermediate bottom stiffeners; bottom girder thickness; intermediate bottom stiffener h; intermediate bottom stiffener t; intermediate bottom stiffener w; intermediate bottom stiffener torc & 7 & C115--C126 (12)\\
11 & Structure Transverse Bottom & floor t; intermediate bottom frame h; intermediate bottom frame t; intermediate bottom frame w; intermediate bottom frame torc & 5 & C136--C143 (8)\\
12 & Side Shell & side shell thickness; Inner Side Shell Thickness & 2 & C110--C114 (5)\\
13 & Structure Longitudinal Side & Sheer strake thickness; Sheer strake height; num side shell stiffeners; side shell Long stiff h; side shell long stiff t; side shell long stiff w; side shell long stiff torc & 7 & C134--C135 (2)\\
14 & Structure Transverse Side & web frames per hold; web frame h; web frame t; web frame torc; num intermediate trans frames; intermediate side frame h; intermediate side frame t; intermediate side frame w; intermediate side frame torc & 9 & C152--C157 (6)\\
15 & Top Deck Plating & Top Deck Thickness & 1 & C158--C162 (5)\\
16 & Structure Longitudinal Deck & deck girder h; deck girder t; deck girder w; deck girder torc; deck stiff h; deck stiff t; deck stiff w; deck stiff torc & 8 & C127--C133 (7)\\
17 & Structure Transverse Deck & trans deck beam h; trans deck beam t; trans deck beam w; trans deck beam torc; intermediate deck frame h; intermediate deck frame t; intermediate deck frame w; intermediate deck frame torc & 8 & C144--C151 (8)\\
18 & Bulkhead Structure & Trans Bulkhead Thickness; num vertical bulkhead stiffeners; blkhd vert stiff h; blkhd vert stiff t; blkhd vert stiff w; blkhd vert stiff torc; num transverse bulkhead stiffeners; blkhd trans stiff h; blkhd trans stiff t; blkhd trans stiff w; blkhd trans stiff torc & 11 & C016--C020 (5)\\
19 & Hopper Structure & floor web bracket L1; floor web bracket L2; floor web bracket t; deck web bracket L1; deck web bracket L2; deck web bracket t; bulk car bottom hopper t; bulk car top hopper t; num bulk car bottom hopper stiff; bulk car bottom hopper stiff h; bulk car bottom hopper stiff t; bulk car bottom hopper stiff w; bulk car bottom hopper stiff torc; num bulk car top hopper stiff; bulk car top hopper stiff h; bulk car top hopper stiff t; bulk car top hopper stiff w; bulk car top hopper stiff torc & 18 & C040--C049 (10)\\
20 & Hatch Structure & Hatch Opening L; Hatch Opening W; Hatch Opening R & 3 & C038--C039 (2)\\
\end{longtable}
\end{arxivlongtable}
\endgroup

\subsection{Constraint Catalogue and Sources}
\label{sec_D3}
The catalogue contains 162 local and 35 system-scope checks. Its assignments match the frozen guidance map used in the experiments. Source abbreviations include the International Convention for the Safety of Life at Sea (SOLAS), the U.S. Code of Federal Regulations (CFR), the International Code on Intact Stability (IS Code), and gross tonnage (GT). SM denotes section modulus. The recorded basis identifies the implemented rule reference, physical relation, geometric check, empirical method domain, or project criterion; it is not a claim of independent rule verification. ABS marine-vessel rules and the IACS common structural rules are cited for their respective constraints \cite{ABS_MVR,IACS_CSR_2025}. The catalogue contains 197 conditional checks, and the evaluator selects checks according to the design configuration. The separate non-counted flooding diagnostic is excluded. IDs below are display references within this appendix. Geometric checks are described by their physical pass conditions. Chine half-breadth is half the hull width at the chine; the chine fillet is the rounded transition between the hull side and bottom.

The eight-column constraint record stores the check class, name, requirement, threshold, offered value, units, signed evaluation, and pass/fail label. For an offered quantity $q$ and requirement $h$, a minimum uses $g=h-q$ and a maximum uses $g=q-h$, in the recorded units. Closed comparisons use $g\leq0$ unless an explicit equality allowance applies. Strict checks preserve strictness: lightship mass must be below displacement and available cargo must be positive. Equation~\eqref{eq_D2} defines the corresponding pass predicates, where $\tau_k$ is the implemented numerical allowance and $b_k$ is the pass/fail result of a domain, geometry, or solver check:
\begin{equation}
P_k(\mathbf x)=
\begin{cases}
\mathbb 1[g_k(\mathbf x)\leq\tau_k], & \text{closed numerical comparison},\\
\mathbb 1[g_k(\mathbf x)<0], & \text{strict numerical comparison},\\
\mathbb 1[b_k(\mathbf x)=\mathrm{true}], & \text{pass/fail comparison}.
\end{cases}
\label{eq_D2}
\end{equation}
Thus a zero signed residual is not universally a pass. No common engineering tolerance is added to every check. For active hold, ballast, and fuel-tank volumes, the equality allowance is $64\epsilon_{64}\max(1,|h|)$ in m$^3$. Departure trim uses $g=|T_a-T_f|-0.01T_{eq}$ and allowance $64\epsilon_{64}\max(1,|T_a-T_f|,0.01|T_{eq}|)$ in metres, where $\epsilon_{64}=2^{-52}$ sets the numerical round-off scale. Distributed-mass closure requires an absolute mass-normalized residual at most $10^{-8}$; force, first-moment, endpoint-shear, and endpoint-bending closure use $10^{-6}$ with the normalizations in Table~\ref{tab_C3}. Their recorded signed evaluation subtracts the specified numerical tolerance from the absolute normalized residual. These numerical tolerances are distinct from physical design limits and optimizer stopping criteria. The phrase ``Engineering numerical-quality gate'' in Table~\ref{tab_D3} denotes these numerical tolerances.

Diagnostic rows without a pass/fail result are excluded. Explicit passes defined by the selected design configuration, including absent numbered holds, are retained as recorded; a missing result is not converted into a pass. When several recorded names refer to the same engineering check, that check is counted once. It passes only if every recorded version passes.
\begingroup
\footnotesize
\setlength{\tabcolsep}{1.5pt}
\begin{arxivlongtable}
\begin{longtable}{r p{0.16\textwidth} p{0.13\textwidth} p{0.24\textwidth} p{0.27\textwidth} p{0.10\textwidth}}
\caption[Complete engineering-constraint catalogue]{Complete 197-constraint engineering catalogue and frozen guidance assignment. The basis column identifies the requirement or method source used for each check; it does not imply independent rule verification or regulatory approval.}\label{tab_D3}\\
\hline
ID & Assigned to & Class & Check & Recorded basis or source & Units\\
\hline
\endfirsthead
\multicolumn{6}{c}{Table \thetable\ continued}\\
\hline
ID & Assigned to & Class & Check & Recorded basis or source & Units\\
\hline
\endhead
\hline
\multicolumn{6}{r}{Continued on next page}\\
\endfoot
\hline
\endlastfoot
C001 & Bottom Shell and Double Bottom & Structure & Bottom-shell absolute minimum thickness & ABS MVR 3-2-2/3.15 & mm\\
C002 & Bottom Shell and Double Bottom & Structure & Double-bottom height - absolute minimum & ABS MVR 3-2-4/1.2.ii & mm\\
C003 & Bottom Shell and Double Bottom & Structure & Double-bottom height - breadth twentieth & ABS MVR 3-2-4/1.2.ii & mm\\
C004 & Bottom Shell and Double Bottom & Structure & Double-bottom height - draft and breadth equation & ABS MVR 3-2-4/3.1.1(c) & mm\\
C005 & Bottom Shell and Double Bottom & Structure & Inner bottom plating - cargo pressure & ABS MVR 3-2-4/9.2 & mm\\
C006 & Bottom Shell and Double Bottom & Structure & Inner bottom plating - excavator wheel loading & ABS MVR 3-2-4/9.7; 3-2-3/5.17 & mm\\
C007 & Bottom Shell and Double Bottom & Structure & Inner bottom plating - general Rule thickness & ABS MVR 3-2-4/9.1 base equation; longitudinal framing & mm\\
C008 & Bottom Shell and Double Bottom & Structure & Inner bottom plating - grab loading & ABS MVR 3-2-4/9.5 & mm\\
C009 & Bottom Shell and Double Bottom & Structure & MVR bottom-shell pressure thickness & ABS MVR 3-2-2/3.15.1 & mm\\
C010 & Bottom Shell and Double Bottom & Structure & MVR inner-bottom optimized-cargo pressure thickness & ABS MVR 3-2-4/9.2 & mm\\
C011 & Bottom Shell and Double Bottom & Structure & MVR longitudinally framed minimum bottom-shell thickness & ABS MVR 3-2-2/3.15.2 & mm\\
C012 & Bottom Shell and Double Bottom & Structure & Min Inner Bottom Grab-Service Thickness & ABS 3-2-4/9.5 & mm\\
C013 & Bottom Shell and Double Bottom & Subdivision Arrangement & SOLAS double-bottom height & SOLAS II-1/9.2 & m\\
C014 & Bottom Shell and Double Bottom & Subdivision Arrangement & SOLAS double-bottom transverse protection to bilge & SOLAS II-1/9.1 & m half-breadth in mirrored hull model\\
C015 & Bottom Shell and Double Bottom & Subdivision Arrangement & SOLAS double-bottom without wells or omitted regions & SOLAS II-1/9.1; implemented geometry checks & Boolean\\
C016 & Bulkhead Structure & Structure & Horizontal bulkhead stiffener - watertight loading & ABS MVR 3-2-9/5.3; unsupported ends c=0.60 & cm\textasciicircum{}3\\
C017 & Bulkhead Structure & Structure & Vertical bulkhead stiffener - watertight loading & ABS MVR 3-2-9/5.3; unsupported ends c=0.60 & cm\textasciicircum{}3\\
C018 & Bulkhead Structure & Structure & Watertight bulkhead plating - absolute minimum & ABS MVR 3-2-9/5.1 absolute minimum & mm\\
C019 & Bulkhead Structure & Structure & Watertight bulkhead plating - pressure equation & ABS MVR 3-2-9/5.1 pressure equation & mm\\
C020 & Bulkhead Structure & Structure & Watertight bulkhead plating - spacing minimum & ABS MVR 3-2-9/5.1 spacing minimum & mm\\
C021 & Compartment Arrangement & Bulkhead\_Spacing & Min Engine Room Length & Hull-imposed Constraint & m\\
C022 & Compartment Arrangement & Cargo Hold Volume & Cargo Hold 0 Not Overfilled & Cargo Hold Volume >= Cargo Volume Filled & m\textasciicircum{}3\\
C023 & Compartment Arrangement & Cargo Hold Volume & Cargo Hold 1 Not Overfilled & Cargo Hold Volume >= Cargo Volume Filled & m\textasciicircum{}3\\
C024 & Compartment Arrangement & Cargo Hold Volume & Cargo Hold 10 Not Overfilled & Cargo Hold Volume >= Cargo Volume Filled & m\textasciicircum{}3\\
C025 & Compartment Arrangement & Cargo Hold Volume & Cargo Hold 11 Not Overfilled & Cargo Hold Volume >= Cargo Volume Filled & m\textasciicircum{}3\\
C026 & Compartment Arrangement & Cargo Hold Volume & Cargo Hold 12 Not Overfilled & Cargo Hold Volume >= Cargo Volume Filled & m\textasciicircum{}3\\
C027 & Compartment Arrangement & Cargo Hold Volume & Cargo Hold 13 Not Overfilled & Cargo Hold Volume >= Cargo Volume Filled & m\textasciicircum{}3\\
C028 & Compartment Arrangement & Cargo Hold Volume & Cargo Hold 14 Not Overfilled & Cargo Hold Volume >= Cargo Volume Filled & m\textasciicircum{}3\\
C029 & Compartment Arrangement & Cargo Hold Volume & Cargo Hold 2 Not Overfilled & Cargo Hold Volume >= Cargo Volume Filled & m\textasciicircum{}3\\
C030 & Compartment Arrangement & Cargo Hold Volume & Cargo Hold 3 Not Overfilled & Cargo Hold Volume >= Cargo Volume Filled & m\textasciicircum{}3\\
C031 & Compartment Arrangement & Cargo Hold Volume & Cargo Hold 4 Not Overfilled & Cargo Hold Volume >= Cargo Volume Filled & m\textasciicircum{}3\\
C032 & Compartment Arrangement & Cargo Hold Volume & Cargo Hold 5 Not Overfilled & Cargo Hold Volume >= Cargo Volume Filled & m\textasciicircum{}3\\
C033 & Compartment Arrangement & Cargo Hold Volume & Cargo Hold 6 Not Overfilled & Cargo Hold Volume >= Cargo Volume Filled & m\textasciicircum{}3\\
C034 & Compartment Arrangement & Cargo Hold Volume & Cargo Hold 7 Not Overfilled & Cargo Hold Volume >= Cargo Volume Filled & m\textasciicircum{}3\\
C035 & Compartment Arrangement & Cargo Hold Volume & Cargo Hold 8 Not Overfilled & Cargo Hold Volume >= Cargo Volume Filled & m\textasciicircum{}3\\
C036 & Compartment Arrangement & Cargo Hold Volume & Cargo Hold 9 Not Overfilled & Cargo Hold Volume >= Cargo Volume Filled & m\textasciicircum{}3\\
C037 & Compartment Arrangement & Structure & CSR uniform cargo-region bulkhead spacing & CSR-H Pt 1 Ch 2 Sec 2 [1.1.5] & m interval deviation\\
C038 & Hatch Structure & Structure & MVR strength-deck hatch-corner radius & ABS MVR 3-2-3/5.5.1 & mm\\
C039 & Hatch Structure & Structure & Min Hatch Area & self imposed & m\textasciicircum{}2\\
C040 & Hopper Structure & Structure & Lower hopper plating - grab loading & ABS MVR 3-2-4/9.5 & mm\\
C041 & Hopper Structure & Structure & Max Lower Hopper Bracket Angle & self imposed & degrees\\
C042 & Hopper Structure & Structure & Max Upper Hopper Bracket Angle & self imposed & degrees\\
C043 & Hopper Structure & Structure & Min Lower Hopper Bracket Angle & self imposed & degrees\\
C044 & Hopper Structure & Structure & Min Lower Hopper Bracket Thickness & self imposed & mm\\
C045 & Hopper Structure & Structure & Min Lower Hopper Grab-Service Thickness & ABS 3-2-4/9.5 & mm\\
C046 & Hopper Structure & Structure & Min Lower Hopper Plate Thickness & self imposed & mm\\
C047 & Hopper Structure & Structure & Min Upper Hopper Bracket Angle & self imposed & degrees\\
C048 & Hopper Structure & Structure & Min Upper Hopper Bracket Thickness & self imposed & mm\\
C049 & Hopper Structure & Structure & Min Upper Hopper Plate Thickness & self imposed & mm\\
C050 & Hull Bow & Hull & Bow Rake and Keel Rise Intersection fwd of Keel Rise start & Hull Geometry & meters\\
C051 & Hull Bow & Hull & Bow Rake and Keel Rise above baseline & Hull Geometry & meters\\
C052 & Hull Bow & Hull & Bow Rake and Keel Rise below deck & Hull Geometry & meters\\
C053 & Hull Bow & Hull & Bow Rake and Keel Rise in LOA & Hull Geometry & meters\\
C054 & Hull Bow & Hull & Bow taper length at Bow rake-Keel intersect is positive & Hull Geometry & meters\\
C055 & Hull Bow & Hull & Bow taper length at deck is positive & Hull Geometry & meters\\
C056 & Hull Bow & Hull & Bow taper length is positive in Bow Rake (1) & Hull Geometry & meters\\
C057 & Hull Bow & Hull & Bow taper length is positive in Bow Rake (2) & Hull Geometry & meters\\
C058 & Hull Bow & Hull & Drift Angle at Baseline < 90deg & Hull Geometry & degrees\\
C059 & Hull Bow & Hull & Drift Angle at Baseline > 0 & Hull Geometry & degrees\\
C060 & Hull Bow & Hull & Drift Angle at Deck < 90deg & Hull Geometry & degrees\\
C061 & Hull Bow & Hull & Drift Angle at Deck > 0deg & Hull Geometry & degrees\\
C062 & Hull Bow & Hull & Drift Angle at Drift Vertex > 0deg & Hull Geometry & degrees\\
C063 & Hull Bow & Hull & Drift angle at drift vertex < 90 degrees & Hull Geometry & degrees\\
C064 & Hull Bulbous Bow & Hull & Lower vertical bulbous bow radius is less than Rk & Hull Geometry & meters\\
C065 & Hull Bulbous Bow & Hull & bow bulb beam is less than B@Z = bulb's widest point & Hull Geometry & meters\\
C066 & Hull Bulbous Bow & Hull & bulb is strictly forward of the end of the bow taper (1) & Hull Geometry & meters\\
C067 & Hull Bulbous Bow & Hull & bulb is strictly forward of the end of the bow taper (2) & Hull Geometry & meters\\
C068 & Hull Bulbous Bow & Hull & bulb is strictly forward of the end of the bow taper (3) & Hull Geometry & meters\\
C069 & Hull Bulbous Bow & Hull & longitudinal bulb radius less than Rk & Hull Geometry & meters\\
C070 & Hull Midbody & Hull & Bottom-side end of chine fillet does not exceed chine half-breadth & Hull Geometry & meters\\
C071 & Hull Midbody & Hull & Keel and chine fillets do not overlap along the bottom & Hull Geometry & meters\\
C072 & Hull Midbody & Hull & Keel Radius is not exactly 0 & Hull Geometry & meters\\
C073 & Hull Midbody & Hull & Keel Rise Fwd of Bow Taper End & Hull Geometry & meters\\
C074 & Hull Midbody & Hull & Positive Beam at Chine & Hull Geometry & meters\\
C075 & Hull Midbody & Hull & Positive Chine Height & Hull Geometry & meters\\
C076 & Hull Midbody & Hull & Positive Chine Radius & Hull Geometry & meters\\
C077 & Hull Midbody & Hull & Positive Gunwale-chine Intersection & Hull Geometry & meters\\
C078 & Hull Principal & Hull & Combined bow and stern taper lengths do not exceed overall length & Hull Geometry & meters\\
C079 & Hull Principal & Hull & Reference waterline height does not exceed hull depth & Hull Geometry & meters\\
C080 & Hull Principal & Hydrostatic Numerical Foundation & Consistent full-beam block coefficient & internal numerical method & absolute Cb error\\
C081 & Hull Principal & Hydrostatic Numerical Foundation & Full-waterplane integration at design draft & internal numerical method & Boolean\\
C082 & Hull Principal & Structure & MVR maximum Rule length & ABS MVR 3-1-2/7.i & m\\
C083 & Hull Principal & Structure & MVR maximum breadth from depth & ABS MVR 3-1-2/7.ii-b & m\\
C084 & Hull Principal & Structure & MVR maximum breadth from length & ABS MVR 3-1-2/7.ii-a & m\\
C085 & Hull Stern & Hull & Bc\_trans < B@Dc\_trans & Hull Geometry & meters\\
C086 & Hull Stern & Hull & Positive transom Gunwale-chine Intersection & Hull Geometry & meters\\
C087 & Hull Stern & Hull & Positive transom chine radius & Hull Geometry & meters\\
C088 & Hull Stern & Hull & Stern rise starts aft of stern taper start & Hull Geometry & meters\\
C089 & Hull Stern & Hull & positive transom chine beam & Hull Geometry & meters\\
C090 & Hull Stern & Hull & positive transom chine height & Hull Geometry & meters\\
C091 & Hull Stern & Hull & stern taper is positive at deck & Hull Geometry & meters\\
C092 & Hull Stern & Hull & stern taper is positive at transom-stern rise intersection & Hull Geometry & meters\\
C093 & Hull Stern & Hull & stern taper length is positive & Hull Geometry & meters\\
C094 & Hull Stern & Hull & stern-transom intersect is aft of sternrise start & Hull Geometry & meters\\
C095 & Hull Stern & Hull & transom chine fillet is inboard of chine beam & Hull Geometry & meters\\
C096 & Hull Stern & Hull & transom keel fillet is inboard of chine fillet & Hull Geometry & meters\\
C097 & Hull Stern Bulb & Hull & Stern Bulb lower vertical radius is less than Rk & Hull Geometry & meters\\
C098 & Hull Stern Bulb & Hull & Stern bulb beam is less than B@Z = bulb's widest point & Hull Geometry & meters\\
C099 & Hull Stern Bulb & Hull & Stern bulb height is less than stern-transom intersect & Hull Geometry & meters\\
C100 & Hull Stern Bulb & Hull & Stern bulb is aft of the start of the stern taper (1) & Hull Geometry & meters\\
C101 & Hull Stern Bulb & Hull & Stern bulb is aft of the start of the stern taper (2) & Hull Geometry & meters\\
C102 & Hull Stern Bulb & Hull & Stern bulb is aft of the start of the stern taper (3) & Hull Geometry & meters\\
C103 & Hull Stern Bulb & Hull & Stern bulb longitudinal radius is less than Rk & Hull Geometry & meters\\
C104 & Powering & Fuel Tank Volume & Fuel Tank Not Overfilled & Fuel Tank Volume >= Fuel Volume Filled & m\textasciicircum{}3\\
C105 & Powering & Powering Method Domain & Holtrop low/intermediate-speed Froude boundary & Holtrop (1984), high-speed caveat above Fn 0.5 & dimensionless\\
C106 & Powering & Powering Method Domain & Holtrop prismatic-coefficient equation domain & Holtrop and Mennen (1982), implemented form-factor equation & dimensionless\\
C107 & Powering & Powering Method Domain & Holtrop run-length denominator magnitude & Holtrop and Mennen (1982), |4 Cp - 1| must be nonzero & dimensionless\\
C108 & Powering & Powering Method Domain & Holtrop run-length to waterline-length equation domain & Holtrop and Mennen (1982), implemented form-factor equation; numerical clearance & dimensionless\\
C109 & Powering & Powering Method Domain & Holtrop waterline-length to breadth equation domain & Holtrop (1984), implemented c17 equation & dimensionless\\
C110 & Side Shell & Structure & CSR minimum bulk-carrier double-side width & CSR-H Pt 1 Ch 2 Sec 3 [3.1.2] & m\\
C111 & Side Shell & Structure & MVR side-shell amidship end-rule minimum & ABS MVR 3-2-2/3.9.3-c; 3-2-2/5.1 & mm\\
C112 & Side Shell & Structure & MVR side-shell pressure thickness & ABS MVR 3-2-2/3.9.3 & mm\\
C113 & Side Shell & Structure & Min Inner Side Shell Thickness & ABS 3-2-2/3.9 & mm\\
C114 & Side Shell & Structure & Side-shell end thickness at actual spacing & ABS MVR 3-2-2/5.1 & mm\\
C115 & Structure Longitudinal Bottom & Structure & Double-bottom girder web thickness in deep tank & ABS MVR 3-2-10/3.7.2(iii) & mm\\
C116 & Structure Longitudinal Bottom & Structure & Inner-bottom longitudinal - bottom-strength minimum & ABS MVR 3-2-4/11.5-a; 0.85 of 3-2-4/11.3 & cm\textasciicircum{}3\\
C117 & Structure Longitudinal Bottom & Structure & Inner-bottom longitudinal - optimized cargo loading & ABS MVR 3-2-4/11.5-b & cm\textasciicircum{}3\\
C118 & Structure Longitudinal Bottom & Structure & MVR center-girder plate thickness amidships & ABS MVR 3-2-4/3.1.1(a) & mm\\
C119 & Structure Longitudinal Bottom & Structure & Max Allowable Uniform Bottom Loading & ABS 3-2-8/3.3 & kN\\
C120 & Structure Longitudinal Bottom & Structure & Max Bottom Girder Spacing & ABS 3-2-4/3.7 & mm\\
C121 & Structure Longitudinal Bottom & Structure & Max Bottom Stiffener Spacing & ABS 3-2-4/5 & mm\\
C122 & Structure Longitudinal Bottom & Structure & Min Bottom Girder Height & ABS 3-2-8/5.7 & mm\\
C123 & Structure Longitudinal Bottom & Structure & Min Bottom Girder Section Modulus & ABS 3-2-8/5.3 & cm\textasciicircum{}3\\
C124 & Structure Longitudinal Bottom & Structure & Min Bottom Girder Thickness & ABS 3-2-8/5.13 & mm\\
C125 & Structure Longitudinal Bottom & Structure & Min Bottom Stringer SM & ABS 3-2-4/11.3 & cm\textasciicircum{}3\\
C126 & Structure Longitudinal Bottom & Structure & Min Center Bottom Girder Thickness & ABS 3-2-4/3.1 & mm\\
C127 & Structure Longitudinal Deck & Structure & MVR deck-girder minimum depth & ABS MVR 3-2-8/5.7.i & mm\\
C128 & Structure Longitudinal Deck & Structure & Max Allowable Uniform Deck Loading & ABS 3-2-8/3.3 & kN\\
C129 & Structure Longitudinal Deck & Structure & Min Deck Girder Height & ABS 3-2-8/5.7 & mm\\
C130 & Structure Longitudinal Deck & Structure & Min Deck Girder Thickness & ABS 3-2-8/5.13 & mm\\
C131 & Structure Longitudinal Deck & Structure & Min Relative Deck Stringer Section Modulus & ABS 3-2-7/3.1 & cm\textasciicircum{}3\\
C132 & Structure Longitudinal Deck & Structure & Strength-deck girder section modulus clear of tanks & ABS MVR 3-2-8/5.3 & cm\textasciicircum{}3\\
C133 & Structure Longitudinal Deck & Structure & Strength-deck longitudinal section modulus & ABS MVR 3-2-7/3.1; longitudinal strength deck & cm\textasciicircum{}3\\
C134 & Structure Longitudinal Side & Structure & Min Side Stringer Height & ABS 3-2-6/5.3 & mm\\
C135 & Structure Longitudinal Side & Structure & Side stringer - ordinary side-shell loading & ABS MVR 3-2-6/5.1 & cm\textasciicircum{}3\\
C136 & Structure Transverse Bottom & Structure & Double-bottom floor web thickness in deep tank & ABS MVR 3-2-10/3.7.2(iii) & mm\\
C137 & Structure Transverse Bottom & Structure & MVR solid-floor plate thickness & ABS MVR 3-2-4/5.1-b & mm\\
C138 & Structure Transverse Bottom & Structure & Max Floor Spacing & ABS 3-2-4/5 & mm\\
C139 & Structure Transverse Bottom & Structure & Min Bottom Floor Height & ABS 3-2-8/5.7 & mm\\
C140 & Structure Transverse Bottom & Structure & Min Bottom Floor Section Modulus & ABS 3-2-8/5.5 & cm\textasciicircum{}3\\
C141 & Structure Transverse Bottom & Structure & Min Bottom Floor Thickness & ABS 3-2-8/5.13 & mm\\
C142 & Structure Transverse Bottom & Structure & Min Bottom Frame SM & ABS 3-2-4/7.3 & cm\textasciicircum{}3\\
C143 & Structure Transverse Bottom & Structure & Min Floor Thickness & ABS 3-2-4/5 & mm\\
C144 & Structure Transverse Deck & Structure & MVR deck-transverse minimum depth & ABS MVR 3-2-8/5.7.i & mm\\
C145 & Structure Transverse Deck & Structure & Min Deck Beam Height & ABS 3-2-8/5.7 & mm\\
C146 & Structure Transverse Deck & Structure & Min Deck Beam Thickness & ABS 3-2-8/5.13 & mm\\
C147 & Structure Transverse Deck & Structure & Min Relative Deck Frame Section Modulus & ABS 3-2-7/3.1 & cm\textasciicircum{}3\\
C148 & Structure Transverse Deck & Structure & Ordinary strength-deck transverse-beam section modulus & ABS MVR 3-2-7/3.1; beam between deck girders & cm\textasciicircum{}3\\
C149 & Structure Transverse Deck & Structure & Strength-deck transverse section modulus clear of tanks & ABS MVR 3-2-8/5.5 & cm\textasciicircum{}3\\
C150 & Structure Transverse Deck & Structure Applicability Metric & Deck primary members at modeled deep-tank tops & ABS MVR 3-2-8/5.13-b applicability & --\\
C151 & Structure Transverse Deck & Structure Applicability Metric & Hatch-end beams supported by centerline pillars & ABS MVR 3-2-8/7.1.1 applicability & --\\
C152 & Structure Transverse Side & Structure & MVR maximum transverse-frame spacing & ABS MVR 3-2-5/1.7 & mm\\
C153 & Structure Transverse Side & Structure & MVR transverse side-frame section modulus & ABS MVR 3-2-5/3.1.1 & cm\textasciicircum{}3\\
C154 & Structure Transverse Side & Structure & MVR web-frame section modulus & ABS MVR 3-2-6/3.1.1 & cm\textasciicircum{}3\\
C155 & Structure Transverse Side & Structure & Max Frame Spacing & ABS 3-2-5/1.7 & mm\\
C156 & Structure Transverse Side & Structure & Min Web Frame Height & ABS 3-2-6/3.5 & mm\\
C157 & Structure Transverse Side & Structure & Min Web Frame Thickness & ABS 3-2-6/3.5 & mm\\
C158 & Top Deck Plating & Structure & MVR general strength-deck minimum thickness & ABS MVR 3-2-3/5.1-b & mm\\
C159 & Top Deck Plating & Structure & Strength deck plating - absolute minimum & ABS MVR 3-2-3/5.1 & mm\\
C160 & Top Deck Plating & Structure & Strength deck plating - excavator wheel loading & ABS MVR 3-2-3/5.17 & mm\\
C161 & Top Deck Plating & Structure & Strength deck plating - grab loading & Project extension of ABS MVR 3-2-4/9.5 & mm\\
C162 & Top Deck Plating & Structure & Strength deck plating - unloaded exposed-deck pressure & ABS MVR 3-2-3/5.1 & mm\\
C163 & System & Load Line Applicability & Standard Type-B load-line classification & 46 CFR 42.09-5 & Boolean\\
C164 & System & Load Line Input & Qualifying position-1 hatch-cover branch & 46 CFR 42.20-5(b); 42.15-25(d) or 42.15-30 & Boolean\\
C165 & System & Loading Condition Feasibility & Capacity-feasible cargo allocation reaches longitudinal equilibrium & GenISys physical feasibility: at least one capacity-feasible full-load cargo allocation shall reach displacement and longitudinal-moment equilibrium within the generated hull draft grid & valid candidates\\
C166 & System & Loading Condition Numerical Quality & Full-cargo arrival aft endpoint bending closure & Engineering numerical-quality gate; condition=full\_cargo\_arrival & dimensionless\\
C167 & System & Loading Condition Numerical Quality & Full-cargo arrival aft endpoint shear closure & Engineering numerical-quality gate; condition=full\_cargo\_arrival & dimensionless\\
C168 & System & Loading Condition Numerical Quality & Full-cargo arrival distributed-mass closure & Engineering numerical-quality gate; condition=full\_cargo\_arrival & dimensionless\\
C169 & System & Loading Condition Numerical Quality & Full-cargo arrival hydrostatic equilibrium solved & GenISys loading-condition foundation; condition=full\_cargo\_arrival & Boolean\\
C170 & System & Loading Condition Numerical Quality & Full-cargo arrival longitudinal first-moment closure & Engineering numerical-quality gate; condition=full\_cargo\_arrival & dimensionless\\
C171 & System & Loading Condition Numerical Quality & Full-cargo arrival longitudinal load equilibrium solved & GenISys simultaneous displacement and longitudinal-moment equilibrium; condition=full\_cargo\_arrival & Boolean\\
C172 & System & Loading Condition Numerical Quality & Full-cargo arrival net vertical-force closure & Engineering numerical-quality gate; condition=full\_cargo\_arrival & dimensionless\\
C173 & System & Loading Condition Numerical Quality & Full-cargo departure aft endpoint bending closure & Engineering numerical-quality gate; condition=full\_cargo\_departure & dimensionless\\
C174 & System & Loading Condition Numerical Quality & Full-cargo departure aft endpoint shear closure & Engineering numerical-quality gate; condition=full\_cargo\_departure & dimensionless\\
C175 & System & Loading Condition Numerical Quality & Full-cargo departure distributed-mass closure & Engineering numerical-quality gate; condition=full\_cargo\_departure & dimensionless\\
C176 & System & Loading Condition Numerical Quality & Full-cargo departure hydrostatic equilibrium solved & GenISys loading-condition foundation; condition=full\_cargo\_departure & Boolean\\
C177 & System & Loading Condition Numerical Quality & Full-cargo departure longitudinal first-moment closure & Engineering numerical-quality gate; condition=full\_cargo\_departure & dimensionless\\
C178 & System & Loading Condition Numerical Quality & Full-cargo departure longitudinal load equilibrium solved & GenISys simultaneous displacement and longitudinal-moment equilibrium; condition=full\_cargo\_departure & Boolean\\
C179 & System & Loading Condition Numerical Quality & Full-cargo departure net vertical-force closure & Engineering numerical-quality gate; condition=full\_cargo\_departure & dimensionless\\
C180 & System & Loading Condition Trim & Full-cargo departure one-percent-draft trim & Departure trim tolerance: |T\_aft - T\_forward| <= 0.01*T\_equilibrium from simultaneous load equilibrium; condition=full\_cargo\_departure & m\\
C181 & System & Longitudinal Load Numerical Check & Longitudinal load equilibrium solved & GenISys simultaneous displacement-volume and longitudinal-first-moment equilibrium & Boolean\\
C182 & System & Regulatory Applicability & 2008 IS Code intact-stability route & 46 CFR 170.165; SOLAS II-1/5.1; 2008 IS Code Part A & Boolean\\
C183 & System & Regulatory Applicability & 46 CFR Part 174 dry-cargo tonnage gate & 46 CFR 174.005; 46 CFR 69.57 & 1969 Convention GT\\
C184 & System & Regulatory Applicability & 46 CFR Subpart-J new dry-cargo tonnage gate & 46 CFR 174.350; 174.355; 46 CFR 69.57 & 1969 Convention GT\\
C185 & System & Regulatory Applicability & U.S. Subchapter-S cargo-vessel route & 46 CFR 170.001; 46 CFR 93.01-1 & Boolean\\
C186 & System & Regulatory Applicability & U.S. international-load-line design route & 46 CFR 42.03-5; 42.03-10; 42.03-30 & Boolean\\
C187 & System & Structural Material & CSR carbon-steel elastic constants & CSR-H Pt 1 Ch 3 Sec 1 [2.1.1] & Boolean\\
C188 & System & Structure & Hull-girder absolute minimum section modulus - bottom & ABS MVR 3-2-1/3.7.1(b) & cm\textasciicircum{}3\\
C189 & System & Structure & Hull-girder absolute minimum section modulus - deck & ABS MVR 3-2-1/3.7.1(b) & cm\textasciicircum{}3\\
C190 & System & Structure & Hull-girder minimum moment of inertia & ABS MVR 3-2-1/3.7.2 & cm\textasciicircum{}2-m\textasciicircum{}2\\
C191 & System & Structure & Hull-girder side-shell nominal shear-stress limit & ABS MVR 3-2-1/3.9.1 & kN/cm\textasciicircum{}2\\
C192 & System & Structure & Hull-girder total-load section modulus - bottom & ABS MVR 3-2-1/3.7.1(a) & cm\textasciicircum{}3\\
C193 & System & Structure & Hull-girder total-load section modulus - deck & ABS MVR 3-2-1/3.7.1(a) & cm\textasciicircum{}3\\
C194 & System & Structure Rule Method & MVR plate-thickness rounding method & ABS MVR 3-1-2/5.3 & Boolean\\
C195 & System & System Capacity & Total cargo-hold mass capacity & GenISys physical feasibility: generated cargo-hold capacity shall be at least the selected full-load cargo mass & kg\\
C196 & System & System Mass Feasibility & Modeled lightship fits selected displacement & GenISys physical feasibility: modeled lightship mass shall be less than displacement mass at the selected operating draft & kg\\
C197 & System & System Mass Feasibility & Positive available cargo payload & GenISys physical feasibility: available cargo mass after lightship, voyage fuel, and miscellaneous deadweight shall be positive & kg\\
\end{longtable}
\end{arxivlongtable}
\endgroup
\subsection{Summary of Design Studies}
\label{sec_D4}
Table~\ref{tab_D4} summarizes the design studies, their purposes, and the number of designs requested. It also identifies whether comparisons use sampling batches, independent model-training runs, or redesign starting designs. Every complete subsystem model contains 20 denoisers. The feasible training population contains 11,195 designs; the completed training population contains 119,143. Reusing a trained model across studies does not create another independent training run.

For the fixed-range and redesign rows, ``performance-study contexts'' means ship-global plus hull constraints for DDPM, active-block local constraints for HCDDPM, and local followed by ship-global constraints for SCDDPM. When performance guidance is enabled, DDPM/SCDDPM use full-state performance regression models and HCDDPM uses a predecessor-prefix performance regression model. Other contexts are stated explicitly in their rows. The reverse endpoint applies to each chain in HCDDPM. All studies retain incomplete requests as unsuccessful outcomes.

\begingroup\small
\setlength{\tabcolsep}{2pt}
\renewcommand{\arraystretch}{1.12}
\begin{arxivlongtable}
\begin{longtable}{>{\raggedright\arraybackslash}p{0.13\textwidth}>{\raggedright\arraybackslash}p{0.21\textwidth}>{\raggedright\arraybackslash}p{0.18\textwidth}>{\raggedright\arraybackslash}p{0.25\textwidth}>{\raggedright\arraybackslash}p{0.17\textwidth}}
\caption[Summary of design studies]{Summary of design studies. Counts are requested production designs, including incomplete evaluations. Fits refer to complete generators, not individual subsystem networks. Pilot, refinement, and workflow-validation requests are not pooled with production.}\label{tab_D4}\\
\toprule
Study & Training and fits & Requests and locks & Guidance and reverse steps & Role and statistical unit\\
\midrule
\endfirsthead
\multicolumn{5}{l}{Table \thetable\ continued}\\
\toprule
Study & Training and fits & Requests and locks & Guidance and reverse steps & Role and statistical unit\\
\midrule
\endhead
\midrule\multicolumn{5}{r}{Continued on next page}\\
\endfoot
\bottomrule
\endlastfoot
Feasibility-strength sweep & Feasible or completed training; one saved fit per architecture/population. & 46,000 requests; 46 conditions; all controls editable. & $\alpha=0,0.1,0.3,0.5,0.7$; DDPM system + hull, HCDDPM local, SCDDPM local/system/combined. Guidance $t=T,\ldots,2$; endpoint $\mathbf z_1$. & Feasibility-guidance comparison (Appendix~\ref{app_E}); ten disjoint 100-request subsets per condition, conditional on saved fits.\\
\addlinespace
Noise/guidance ablation & Feasible training; one saved fit per architecture. & 12,000 requests; four arms per architecture; all controls editable. & Ordinary/reduced noise crossed with absent/present guidance ($\alpha=0.7$). DDPM system + hull, HCDDPM local, SCDDPM system. $t=T,\ldots,2$; endpoint $\mathbf z_1$. & Primary constraint analysis (Appendix~\ref{app_E}); ten paired 100-request batches per arm; nine Holm-adjusted contrasts.\\
\addlinespace
Locally feasible training & 22,192--119,143 passing rows per block; one new HCDDPM and one new SCDDPM, compared with saved globally feasible fits. & 18,000 new requests; 18 conditions; all controls editable. & Frozen feasible-only transforms/classifiers; $\alpha=0,0.1,0.3,0.5,0.7$ and supported local/system/combined modes. $t=T,\ldots,2$; endpoint $\mathbf z_1$. & Supporting training study (Appendix~\ref{app_G}); ten disjoint 100-request subsets; 18 matched comparisons per outcome.\\
\addlinespace
Free-range performance & Feasible training; one saved fit per architecture. Performance regression models use completed training rows. & 900 validation requests; nine arms; range, speed, and dimensions editable. Pilot and refinement cohorts separate. & $\alpha=0.7$; DDPM system + hull, HCDDPM/SCDDPM local. Feasibility window $t=T,\ldots,2$; performance $t=101,\ldots,2$ (DDPM/SCDDPM) or $51,\ldots,2$ (HCDDPM); endpoint $\mathbf z_1$. Full-state/prefix performance regression models (Appendix~\ref{app_H}). & Independent sampling-cohort validation after pilots (Appendix~\ref{app_H}); paired requests/descriptive response distributions, not training replication.\\
\addlinespace
Fixed-range strength selection & Feasible training; one saved fit per architecture; performance regression models use completed training rows. & 36,000 requests; 36 cells; range locked at 1,000, 5,000, or 8,000 nmi. & $\alpha=0.7$, $\gamma=0,0.01,0.03,0.10$; performance-study contexts. Guidance $t=800,\ldots,2$; endpoint $\mathbf z_0$. & Select frozen settings for fresh-fit confirmation (Appendix~\ref{app_J}); ten paired 100-request batches; 27 Holm-adjusted contrasts.\\
\addlinespace
Cost Reduction-Guided Sampling Design Study & Feasible training; five fresh fits per architecture; guidance assets and transforms fixed. & 30,000 requests; two arms per fit; 5,000-nmi range locked. & $\alpha=0.7$; $\gamma=0$ versus DDPM/SCDDPM 0.10 or HCDDPM 0.03. Performance-study contexts. Guidance $t=800,\ldots,2$; endpoint $\mathbf z_0$. & Primary repeatability study (Appendix~\ref{app_I}); five complete fits per architecture; paired guidance and Welch architecture tests, Holm correction across five planned tests.\\
\addlinespace
Two-arm fixed-range guidance & Feasible training; one saved fit per architecture. & 18,000 requests; 18 cells; range locked at 1,000, 5,000, or 8,000 nmi. & $\alpha=0.7$; $\gamma=0$ versus DDPM 0.03 or HCDDPM/SCDDPM 0.10. Performance-study contexts. Guidance $t=800,\ldots,2$; endpoint $\mathbf z_1$. & Cost--cargo comparison (Appendix~\ref{app_J}); ten paired 100-request batches per cell; nine within-architecture/range contrasts.\\
\addlinespace
Hopper feasibility restoration & Feasible training; one saved fit per architecture; 100 held-out source-distinct starting designs fail only Hopper. & 6,000 requests; ten attempts per starting design/arm; 18 Hopper controls editable, 115 locked. & Unguided versus $\alpha=0.7$; performance-study feasibility-guidance contexts, no performance gradient. Guidance $t=800,\ldots,2$; endpoint $\mathbf z_0$. & Targeted feasibility restoration (Appendix~\ref{app_L}); 100 paired starting-design fractions; five Holm-adjusted contrasts; parent coverage separately reported.\\
\addlinespace
Exploratory performance-guided redesign & Feasible training; one saved fit per architecture; 100 held-out starting designs (13 shared with pilot). & 18,000 requests; ten attempts per starting design/arm; only 14 bulb, 11 bulkhead, or 7 bottom-longitudinal controls editable. & $\alpha=0.7$ with/without selected performance guidance; performance-study contexts. Guidance $t=800,\ldots,2$; endpoint $\mathbf z_1$. & Exploratory capability study (Appendix~\ref{app_M}); 100 paired starting-design fractions; nine adjusted guidance contrasts; 87-parent sensitivity excludes pilot overlap.\\
\addlinespace
Independent performance-guided redesign & Same saved fits; 55 currently feasible source-distinct starting designs unused in pilot/production, after 3 of 58 exclusions. & 9,900 requests; ten attempts per starting design/arm; same three edit masks; all other independent controls locked. & $\alpha=0.7$; frozen target-specific performance strengths 0.03/0.10 (Appendix~\ref{app_M}); performance-study contexts. Guidance $t=800,\ldots,2$; endpoint $\mathbf z_0$. & Independent parent-cohort confirmation (Appendix~\ref{app_M}); 55 paired starting-design fractions; nine Holm-adjusted guidance contrasts.\\
\end{longtable}
\end{arxivlongtable}
\endgroup

Feasible-set distance calculations and the embedding reuse the identified generation cohorts; they add no generator fits. The 200 subsets of already generated feasible designs measure sensitivity to which designs are included, while the fresh-fit distance analysis averages subset results within each fit before comparing fits. The embedding is qualitative. Appendix~\ref{sec_C7} separately reports engineering numerical verification, including nine finer-grid reevaluations, rather than a generator-performance experiment.

\subsection{Sampling Order and Locked Coordinates}
\label{sec_D5}
Algorithms~\ref{alg_D1}, \ref{alg_D2}, and~\ref{alg_D3} transcribe the frozen fixed-range sampler used by the performance-strength sweep and five-fit confirmation. They use the zero-based schedule index $k=999,\ldots,0$, with mathematical diffusion time $t=k+1$. Let $M$ be the editable-coordinate mask, $r$ the reference physical controls, and $c=\mathcal Q(r)$ their transformed values. All operations are per design; products with masks are coordinatewise. The requested range is a locked reference coordinate, while redesign masks can select other editable subsets.

During intermediate generation steps, the model receives noisy versions of the fixed reference parameters, matching the noise level appropriate to that step. The underlying reference values remain unchanged. One Gaussian vector $\xi$ is drawn per chain for the locked path. The implementation defines $L_k=c$ for $k\leq0$ and $L_k=\sqrt{\bar a_{k+1}}c+\sqrt{1-\bar a_{k+1}}\xi$ otherwise. The locked parameters are already noise-free during the final $k=0$ denoising call, and their physical values are restored exactly after decoding. Each stochastic reverse step draws a fresh innovation. Guidance is active only for $1\leq k\leq799$, with $\alpha=0.7$ and the selected performance strength $\gamma$; the $k=0$ update uses the reverse mean without guidance or innovation.

Here $G$ denotes the gated classifier gradient in Equation~\eqref{eq_13}, with nonfinite entries replaced by zero and root-mean-square magnitudes at most $10^{-12}$ suppressed. Classifier gradients are masked but are not normalized to unit length. Performance descent $U$ masks the negative gradient of the performance regression model before per-design normalization; norms at most $10^{-8}$ produce a zero direction. $C_{k,d}(v)$ caps $v$ at L2 length $\sqrt{\beta_{k+1}}\sqrt d$ as in Equation~\eqref{eq_15}, where $d$ counts editable coordinates. Classifier inputs and the stage at which they are probed differ across architectures.

\begin{algorithm}[H]
\caption{Monolithic fixed-range sampling}
\label{alg_D1}
\small
\begin{arxivtablefit}
\begin{tabular}{r p{0.91\textwidth}}
1 & Transform $r$ to $c$. Draw locked-path noise $\xi$ and initial noise $\eta_0$; set $z=M\eta_0+(1-M)L_{999}$.\\
2 & For $k=999,\ldots,0$, predict noise from $z$ and compute reverse mean $\mu$ and standard deviation $\sigma$ using Equations~\eqref{eq_8}--\eqref{eq_9}.\\
3 & If $k=0$, set $z=M\mu+(1-M)c$ and finish the chain. Otherwise draw innovation $\eta$.\\
4 & Outside the guidance window, set $y=\mu+\sigma\eta$. Inside the window, perform steps 5--7.\\
5 & Probe ship-global and hull classifiers at the current $z$. Mask their gradients to editable coordinates and embed the hull gradient in the full vector; sum them as $g$.\\
6 & If $\gamma>0$, probe the full-state performance regression model at $(z,k)$ and form masked unit descent $u$; otherwise set $u=0$.\\
7 & Set $d=C_{k,\sum M}(\alpha\sigma g+\gamma\sigma u)$ and $y=\mu+(1-\alpha)\sigma\eta+d$.\\
8 & Replace locked coordinates: $z=My+(1-M)L_{k-1}$. Continue step 2.\\
9 & Inverse-transform and apply the 0.5 threshold to two-option controls; restore every locked physical control exactly to $r$.\\
\end{tabular}
\end{arxivtablefit}
\end{algorithm}

Algorithm~\ref{alg_D2} completes each editable block before proceeding. Finished physical predecessors are transformed into the active denoiser's condition. The prefix performance regression model sees completed predecessor latents and the current active block; unobserved future blocks are masked out. A block with no editable coordinates is retained without a reverse chain. This preserves the intentional ordering: earlier hull blocks cannot condition on a requested range in a later block.

\begin{algorithm}[H]
\caption{Hierarchical fixed-range sampling}
\label{alg_D2}
\small
\begin{arxivtablefit}
\begin{tabular}{r p{0.91\textwidth}}
1 & Initialize generated physical controls to $r$ and completed latent controls to $c$. Visit subsystem blocks in their fixed order.\\
2 & Skip blocks with no editable coordinate. For active block $i$, prepare clean predecessor context, block mask $M_i$, clean reference $c_i$, fixed locked noise $\xi_i$, and Gaussian initial state with locks replaced by $L_{999,i}$.\\
3 & For $k=999,\ldots,0$, predict the active block's noise from its current state $z_i$ and predecessor context; compute $\mu_i,\sigma_i$.\\
4 & At $k=0$, set $z_i=M_i\mu_i+(1-M_i)c_i$ and finish this block. Otherwise draw $\eta_i$.\\
5 & Outside the guidance window, set $y_i=\mu_i+\sigma_i\eta_i$. Inside the window, perform steps 6--8.\\
6 & Probe the active local classifier on current $z_i$ concatenated with predecessor context. Keep gradient components only for editable coordinates of block $i$, giving $g_i$.\\
7 & If $\gamma>0$, insert $z_i$ into the completed latent prefix, mask future blocks, and probe the prefix performance regression model at level $i$ and time $k$. Form masked unit descent $u_i$; otherwise set $u_i=0$.\\
8 & Set $d_i=C_{k,\sum M_i}(\alpha\sigma_i g_i+\gamma\sigma_i u_i)$ and $y_i=\mu_i+(1-\alpha)\sigma_i\eta_i+d_i$.\\
9 & Replace the block's locks: $z_i=M_i y_i+(1-M_i)L_{k-1,i}$. Continue step 3.\\
10 & Store finished $z_i$ in the completed latent prefix. Inverse-transform its editable controls and apply the 0.5 threshold to its two-option controls; continue with the next block.\\
11 & After all blocks, restore every locked physical control exactly to $r$.\\
\end{tabular}
\end{arxivtablefit}
\end{algorithm}

Algorithm~\ref{alg_D3} uses one shared copied state for all denoisers and local classifier probes. A hull-local classifier receives the hull subvector; other local classifiers receive the full state. Only each block's owned, editable gradient components enter its move. Ship-global and performance guidance then probe the assembled proposal, before locked-coordinate replacement.

\begin{algorithm}[H]
\caption{Simultaneous fixed-range sampling}
\label{alg_D3}
\small
\begin{arxivtablefit}
\begin{tabular}{r p{0.91\textwidth}}
1 & Transform $r$ to $c$. Draw fixed locked-path noise and Gaussian initial noise; initialize $z$ as in Algorithm~\ref{alg_D1}.\\
2 & For $k=999,\ldots,0$, copy $s=z$. Every subsystem denoiser reads its block and complementary context from this same $s$. Compute all block means and variances; assemble $\mu,\sigma$.\\
3 & At $k=0$, set $z=M\mu+(1-M)c$ and finish the chain. Otherwise draw one innovation $\eta_i$ per block.\\
4 & Outside the guidance window, or for a block with no editable coordinates, set $y_i=\mu_i+\sigma_i\eta_i$. For an active editable block, perform step 5.\\
5 & Probe its local classifier at $s$, mask to owned editable coordinates, and set $d_i=C_{k,\sum M_i}(\alpha\sigma_i g_i)$; set $y_i=\mu_i+(1-\alpha)\sigma_i\eta_i+d_i$.\\
6 & Assemble all $y_i$ as $y$. Outside the guidance window set $v=y$ and proceed to step 10; otherwise perform steps 7--9.\\
7 & Probe the ship-global classifier at $y$, mask its gradient, and cap its move: $d_S=C_{k,\sum M}(\alpha\sigma G_S(y))$.\\
8 & If $\gamma>0$, probe the full-state performance regression model at $(y,k-1)$ and form masked unit descent $u$; otherwise set $u=0$.\\
9 & Form the reduced-noise base $b=\mu+(1-\alpha)\sigma\eta$. Cap the total local, global, and performance displacement together: $v=b+C_{k,\sum M}(y+d_S+\gamma\sigma u-b)$.\\
10 & Replace locks only after guidance: $z=Mv+(1-M)L_{k-1}$. Continue step 2.\\
11 & Inverse-transform and apply the 0.5 threshold to two-option controls; restore every locked physical control exactly to $r$.\\
\end{tabular}
\end{arxivtablefit}
\end{algorithm}

After these algorithms return 133 physical controls, the frozen decoder rounds whole-number controls, restores constants and aliases, reconstructs dependent dimensions, and initializes cargo solver fields. MiDShip-derived structural adjustments and complete engineering evaluation follow. Thus exact independent-control restoration precedes dependent reconstruction and geometry generation; a lock on an independent coordinate does not freeze every derived response.

The feasibility-guidance ablation uses a different, explicitly retained protocol. All coordinates are editable and $k=0$ is omitted. Monolithic and hierarchical classifiers probe the current noisy state. Simultaneous system guidance probes the assembled ordinary-noise proposal $\mu+\sigma\eta$ before replacing its noise multiplier by the selected ablation value. Its gradient is therefore not evaluated at the subsystem-guided proposal in Algorithm~\ref{alg_D3}. Table~\ref{tab_D4} records the sampling windows and endpoints for each study.

\section{Feasibility-Guidance Diagnostics}
\label{app_E}
This appendix provides complete feasibility-guidance ablation results, guidance-strength comparisons, and exploratory failure-family diagnostics.

\subsection{Four-Arm Ablation}
The feasibility-guidance ablation and feasibility-strength sweep use the same saved feasible-only fits but different sampling cohorts. Their results are retained separately. The ablation endpoints reproduce the strength sweep's ordinary and blended sampling operations under matched random inputs; reference workflow checks found zero control differences in the six checked endpoint cases. Table~\ref{tab_E1} records every requested and completed population. Table~\ref{tab_E2} gives the nine planned paired-batch contrasts. For HCDDPM, enabling guidance also changes how model outputs are converted into physical design parameters. Its measured guidance gains therefore cannot be attributed to guidance alone.

\begin{table}[H]
\caption[Four-arm feasibility-guidance ablation counts]{Four-arm ablation counts. Each cell requests 1,000 designs; all requests remain in the feasible-yield denominator. Arms A--D are defined in the main text. Bold marks the highest feasible yield across architectures within each arm.}
\label{tab_E1}
\centering\small
\begin{arxivtablefit}
\begin{tabular}{llrrr}
\toprule
Architecture & Arm & Completed & Feasible & Yield (\%)\\
\midrule
DDPM & A & 976 & 153 & 15.3\\
DDPM & B & 1000 & 547 & 54.7\\
DDPM & C & 986 & 224 & 22.4\\
DDPM & D & 1000 & 581 & 58.1\\
HCDDPM & A & 953 & 196 & 19.6\\
HCDDPM & B & 1000 & 567 & \textbf{56.7}\\
HCDDPM & C & 958 & 346 & 34.6\\
HCDDPM & D & 999 & 597 & \textbf{59.7}\\
SCDDPM & A & 949 & 274 & \textbf{27.4}\\
SCDDPM & B & 993 & 565 & 56.5\\
SCDDPM & C & 963 & 386 & \textbf{38.6}\\
SCDDPM & D & 994 & 595 & 59.5\\
\bottomrule
\end{tabular}
\end{arxivtablefit}
\end{table}
\begin{table}[H]
\caption[Paired feasibility contrasts]{Planned paired feasibility contrasts in percentage points. Intervals are unadjusted 95\% paired $t$ intervals across ten batches; Holm adjustment covers all nine tests. Bold marks the largest observed gain across architectures for each contrast, not a significant difference between architectures.}
\label{tab_E2}
\centering\small
\begin{arxivtablefit}
\begin{tabular}{llrrr}
\toprule
Architecture & Contrast & Gain (pp) & 95\% interval (pp) & $p_{\mathrm{Holm}}$\\
\midrule
DDPM & B--A & \textbf{39.4} & [35.34, 43.46] & $3.61\times10^{-8}$\\
DDPM & C--A & 7.1 & [4.59, 9.61] & 0.000503\\
DDPM & D--B & \textbf{3.4} & [0.09, 6.71] & 0.0451\\
HCDDPM & B--A & 37.1 & [33.22, 40.98] & $3.64\times10^{-8}$\\
HCDDPM & C--A & \textbf{15.0} & [11.59, 18.41] & $2.21\times10^{-5}$\\
HCDDPM & D--B & 3.0 & [1.45, 4.55] & 0.00522\\
SCDDPM & B--A & 29.1 & [25.12, 33.08] & $3.39\times10^{-7}$\\
SCDDPM & C--A & 11.2 & [7.78, 14.62] & 0.000203\\
SCDDPM & D--B & 3.0 & [0.71, 5.29] & 0.0315\\
\bottomrule
\end{tabular}
\end{arxivtablefit}
\end{table}

\subsection{Feasibility-Strength Sweep}
The feasibility-strength sweep crossed feasible-only and full-data fitting with unguided DDPM, HCDDPM, and SCDDPM and architecture-specific guided modes at $\alpha=0.1,0.3,0.5,0.7$. Its 46 conditions submitted 46,000 designs; 44,737 completed and 17,850 were fully feasible. Feasible-only and full-data fitting differ in population size, transform, and label conditioning, so they do not isolate a training-label effect. Table~\ref{tab_E3} retains unguided and strongest-guidance endpoints. Ten disjoint 100-design subsets describe within-fit sampling variation, not independent training replications. Feasibility uses all requested proposals and violations only completed evaluations. Within each training population, SCDDPM is compared with DDPM and HCDDPM at the same feasibility-guidance strength. Unguided models are compared at zero strength. At each nonzero strength, the three SCDDPM guidance configurations (subsystem-level, ship-global, and combined) are compared with ship-global plus hull guidance for DDPM and subsystem-level guidance for HCDDPM. Across both training populations, this produces 52 comparisons per outcome. Two-sided Welch tests compare the saved subset estimates, with Holm correction applied to all 52 tests for each outcome. The paired four-arm ablation provides the main evidence about guidance mechanisms.

\begin{table}[H]
\caption[Feasibility-guidance endpoint outcomes]{Unguided and strongest-guidance conditions. Each row submitted 1,000 designs. Counts are exact full-set totals; rates and violations are mean $\pm$ sample SD across the ten 100-design subsets. Feasibility includes incomplete proposals in each subset denominator, while violations use completed evaluations. Bold marks the highest mean feasibility and fewest mean violations within each training-population and guidance-strength group, including ties; counts are unranked.}
\label{tab_E3}
\centering\small
\setlength{\tabcolsep}{3pt}
\begin{arxivtablefit}
\begin{tabular}{lllrrrrr}
\toprule
Training population & Architecture & Guidance & $\alpha$ & Completed & Feasible $\uparrow$ & Feasibility (\%) $\uparrow$ & Violations $\downarrow$\\
\midrule
Feasible only & DDPM & None & 0.0 & 973 & 143 & $14.3\pm3.2$ & $2.002\pm0.281$\\
Feasible only & DDPM & System + hull & 0.7 & 1,000 & 551 & $55.1\pm4.1$ & $0.525\pm0.061$\\
Feasible only & HCDDPM & None & 0.0 & 958 & 208 & $20.8\pm3.9$ & $1.261\pm0.065$\\
Feasible only & HCDDPM & Local & 0.7 & 999 & 597 & $59.7\pm5.9$ & $0.451\pm0.075$\\
Feasible only & SCDDPM & None & 0.0 & 948 & 279 & $\mathbf{27.9\pm4.4}$ & $\mathbf{1.056\pm0.067}$\\
Feasible only & SCDDPM & Local & 0.7 & 996 & 592 & $59.2\pm3.2$ & $0.436\pm0.034$\\
Feasible only & SCDDPM & System & 0.7 & 991 & 603 & $\mathbf{60.3\pm4.6}$ & $0.429\pm0.047$\\
Feasible only & SCDDPM & Local + system & 0.7 & 997 & 600 & $60.0\pm2.7$ & $\mathbf{0.426\pm0.035}$\\
All known labels & DDPM & None & 0.0 & 920 & 4 & $0.4\pm0.5$ & $13.153\pm1.409$\\
All known labels & DDPM & System + hull & 0.7 & 1,000 & 261 & $26.1\pm3.6$ & $1.187\pm0.086$\\
All known labels & HCDDPM & None & 0.0 & 856 & 37 & $3.7\pm1.9$ & $\mathbf{2.899\pm0.233}$\\
All known labels & HCDDPM & Local & 0.7 & 1,000 & 533 & $53.3\pm4.6$ & $0.563\pm0.052$\\
All known labels & SCDDPM & None & 0.0 & 903 & 40 & $\mathbf{4.0\pm2.2}$ & $3.104\pm0.166$\\
All known labels & SCDDPM & Local & 0.7 & 1,000 & 443 & $44.3\pm3.6$ & $0.634\pm0.049$\\
All known labels & SCDDPM & System & 0.7 & 999 & 561 & $\mathbf{56.1\pm5.5}$ & $\mathbf{0.491\pm0.069}$\\
All known labels & SCDDPM & Local + system & 0.7 & 1,000 & 476 & $47.6\pm5.1$ & $0.581\pm0.062$\\
\bottomrule
\end{tabular}
\end{arxivtablefit}
\end{table}

The strength-sweep distance calculations use the 133 generator-controlled parameters from completed, evaluated designs, including those that remain infeasible. Incomplete evaluations do not supply a complete cleaned parameter record and are excluded from distance calculations. They remain unsuccessful requests in yield calculations. These distances are calculated from completed designs, including those that remain infeasible. They therefore do not describe diversity within the fully feasible subset. For feasible-only system-guided SCDDPM, the mean completed-design subset Chamfer changed from $6.634\pm0.034$ without guidance to $6.085\pm0.022$ at $\alpha=0.7$, and nearest-neighbor distance from $3.352\pm0.028$ to $2.472\pm0.036$. The feasible-only distance analysis in Appendix~\ref{app_F} provides the primary interpretation.

\subsection{Exploratory Failure-Family Diagnostics}
Constraint-family failure rates use designs with known evaluations for that family, including known partial outputs where available; unknown checks are not passes. Table~\ref{tab_E4} shows the structural, hull-girder, and loading families using completed evaluations. Overlapping families cannot be summed into a total failure rate. Constraints assigned to a subsystem can also depend on parameters from other subsystems. For example, the side-shell pressure-thickness check uses shell thickness and support spacing together with the ship's length, depth, and draft. The inner-bottom cargo-pressure check uses inner-bottom thickness and support spacing together with the cargo fill depth determined by hold geometry and cargo allocation. These diagnostics can suggest which engineering constraints account for observed gains, but neither local-label assignment nor an improved marginal failure rate establishes that reciprocal conditioning caused better subsystem coordination.

\begin{table}[H]
\caption[Exploratory failure-family counts]{Exploratory failure-family counts among completed unguided evaluations. Each entry reports designs failing the constraint family divided by designs evaluated against that family, followed by a percentage. The feasibility-strength sweep and four-arm ablation are shown separately. Families overlap; these descriptive counts do not isolate reciprocal conditioning or prove a causal coordination mechanism. Bold marks the lowest failure percentage across architectures within each study and family.}
\label{tab_E4}
\centering\small
\begin{arxivtablefit}
\begin{tabular}{llrrr}
\toprule
Study & Architecture & Structural family & Hull-girder family & Loading/equilibrium\\
\midrule
Strength sweep & DDPM & 789/973 (81.09\%) & 60/973 (6.17\%) & \textbf{13/973 (1.34\%)}\\
Strength sweep & HCDDPM & 645/958 (67.33\%) & 18/958 (1.88\%) & 29/958 (3.03\%)\\
Strength sweep & SCDDPM & \textbf{592/948 (62.45\%)} & \textbf{15/948 (1.58\%)} & 14/948 (1.48\%)\\
Ablation arm A & DDPM & 794/976 (81.35\%) & 43/976 (4.41\%) & \textbf{12/976 (1.23\%)}\\
Ablation arm A & HCDDPM & 638/953 (66.95\%) & 13/953 (1.36\%) & 29/953 (3.04\%)\\
Ablation arm A & SCDDPM & \textbf{569/949 (59.96\%)} & \textbf{5/949 (0.53\%)} & 34/949 (3.58\%)\\
\bottomrule
\end{tabular}
\end{arxivtablefit}
\end{table}

The lower structural and hull-girder failure rates for decomposed generators recur in these two sampling cohorts. Loading/equilibrium does not show a consistent SCDDPM advantage: its ablation arm A rate is 3.58\%, versus 3.04\% for HCDDPM and 1.23\% for DDPM. The structural-family grouping may contain checks that depend on other blocks; it is not equivalent to single-block dependence.

\section{Equal-Sized Fully Feasible Design Sets}
\label{app_F}
This appendix defines distance measures for the 133 generator-controlled parameters and reports feasible-design comparisons and embedding settings.

Distance calculations use the evaluated, cleaned physical-parameter records produced by the completed ship evaluations. Design identifiers link each record to its feasibility result. Only the 133 independent parameters selected by the generators enter the distance calculations. Fixed settings, derived quantities, aliases, inactive fields, and cargo-allocation outputs are excluded. Each comparison family has at least 100 feasible designs per cell. For each cell, 200 subsets of 100 already generated feasible designs are selected. No design appears twice within one subset, but different subsets can overlap; they are not independent observations for significance testing. The scaling extrema come only from 11,195 fully feasible training designs; reference-constant coordinates contribute zero and generated values are not clipped. The held-out feasible test reference contains 2,618 source-parent-separated designs. For the fixed-range comparisons, separate ship populations are sampled with the required voyage range fixed at the values in Table~\ref{tab_F_reference_windows}. Reference designs are drawn from fully feasible training and held-out test designs whose recorded voyage ranges fall within $\pm250$ nmi of that requirement.

\begin{table}[H]
\caption[Feasible reference populations for fixed-range sampling]{Feasible reference designs used to compare ship populations sampled with different fixed voyage-range requirements. Training and held-out test references use the corresponding $\pm250$-nmi voyage-range windows.}
\label{tab_F_reference_windows}
\centering\small
\begin{arxivtablefit}
\begin{tabular}{rrr}
\toprule
\shortstack{Required voyage range\\fixed during sampling (nmi)} & \shortstack{Training reference\\designs} & \shortstack{Held-out test\\reference designs}\\
\midrule
1,000 & 670 & 164\\
5,000 & 603 & 135\\
8,000 & 486 & 131\\
\bottomrule
\end{tabular}
\end{arxivtablefit}
\end{table}

Equation~\eqref{eq_F1} scales a physical coordinate with nonzero training range:
\begin{equation}
 u_j=(x_j-x_{j,min}^{train})/(x_{j,max}^{train}-x_{j,min}^{train}).
 \label{eq_F1}
\end{equation}
Let $G$ be a 100-design feasible generated set and $R$ the specified feasible reference. Equation~\eqref{eq_F2} defines the directional distance:
\begin{equation}
 d(G,R)=|G|^{-1}\sum_{g\in G}\min_{r\in R}\|g-r\|_2,
 \label{eq_F2}
\end{equation}
Equation~\eqref{eq_F3} defines the bidirectional Chamfer distance as the sum of the two directional mean Euclidean distances; distances are not squared:
\begin{equation}
 D_C(G,R)=d(G,R)+d(R,G),
 \label{eq_F3}
\end{equation}
For indexed generated vectors $G=(\mathbf g_i)_{i=1}^{n}$, Equation~\eqref{eq_F4} defines the mean nearest-neighbor distance, excluding each design itself:
\begin{equation}
 D_{NN}(G)=\frac{1}{n}\sum_{i=1}^{n}\min_{1\leq j\leq n,\,j\neq i}\|\mathbf g_i-\mathbf g_j\|_2.
 \label{eq_F4}
\end{equation}
Smaller $d(G,R)$ indicates proximity to the reference, while $d(R,G)$ also reflects how well 100 generated designs represent that reference. Neither direction is an unconditional coverage guarantee. Table~\ref{tab_F1} summarizes feasibility and feasible-set distances. Table~\ref{tab_F2} reports all four noise and feasibility-guidance conditions, and Table~\ref{tab_F3} reports the fixed-range selection-sweep results. All three tables use the same 133 generator-controlled parameters. Nearest-training proximity is compared with held-out-test-to-training proximity rather than treated as a memorization test.

\begin{table}[H]
\caption[Feasible yield and mean nearest-neighbor distance]{Feasibility-guidance ablation summary. Each arm requests 1,000 designs. Mean nearest-neighbor distance and bidirectional Chamfer distance are reported as mean $\pm$ subset SD across 200 subsets, each containing 100 already generated fully feasible designs in 133 independently scaled physical controls. Larger mean nearest-neighbor distance indicates that neighboring designs are farther apart; smaller Chamfer means greater proximity to the stated reference, not necessarily greater utility. These are descriptive saved-fit comparisons. Within each arm, bold marks the highest yield, largest mean nearest-neighbor distance, and smallest reference distances across architectures.}
\label{tab_F1}
\centering\small
\begin{arxivtablefit}
\begin{tabular}{llrrrr}
\toprule
Architecture & Arm & Feasible yield (\%) & \shortstack{Mean nearest-neighbor\\distance $D_{NN}$} & Training $D_C$ & Test $D_C$\\
\midrule
DDPM & A: ordinary noise & 15.3 & $3.086\pm0.020$ & $\mathbf{6.207\pm0.022}$ & $\mathbf{6.921\pm0.013}$\\
DDPM & D: noise + guidance & 58.1 & $2.385\pm0.041$ & $\mathbf{5.656\pm0.036}$ & $\mathbf{6.534\pm0.021}$\\
HCDDPM & A: ordinary noise & 19.6 & $\mathbf{3.516\pm0.025}$ & $6.815\pm0.018$ & $7.284\pm0.017$\\
HCDDPM & D: noise + guidance & \textbf{59.7} & $2.041\pm0.047$ & $6.052\pm0.022$ & $6.573\pm0.018$\\
SCDDPM & A: ordinary noise & \textbf{27.4} & $3.324\pm0.026$ & $6.609\pm0.022$ & $7.078\pm0.022$\\
SCDDPM & D: noise + guidance & 59.5 & $\mathbf{2.455\pm0.040}$ & $6.085\pm0.023$ & $6.606\pm0.020$\\
\bottomrule
\end{tabular}
\end{arxivtablefit}
\end{table}

\begin{table}[H]
\caption[Fully feasible design distances in the four-arm ablation]{Fully feasible design distances across the four noise and feasibility-guidance conditions, using the evaluated, cleaned values of the 133 generator-controlled parameters. Entries are means across 200 subsets of 100 designs. Directional columns use the training reference; test $D_C$ uses the source-parent-separated test reference. Arms A--D are the four ablation conditions. Within each arm, bold marks the smallest reference distances and largest mean nearest-neighbor distance across architectures.}
\label{tab_F2}
\centering\small
\begin{arxivtablefit}
\begin{tabular}{llrrrrr}
\toprule
Architecture & Arm & $d(G,R_{train})$ & $d(R_{train},G)$ & Train $D_C$ & Test $D_C$ & $D_{NN}$\\
\midrule
DDPM & A & $\mathbf{2.797}$ & $\mathbf{3.410}$ & $\mathbf{6.207}$ & $\mathbf{6.921}$ & $3.086$\\
DDPM & B & $\mathbf{2.459}$ & $\mathbf{3.200}$ & $\mathbf{5.658}$ & $\mathbf{6.536}$ & $2.373$\\
DDPM & C & $\mathbf{2.821}$ & $\mathbf{3.404}$ & $\mathbf{6.226}$ & $\mathbf{6.913}$ & $3.087$\\
DDPM & D & $\mathbf{2.458}$ & $\mathbf{3.198}$ & $\mathbf{5.656}$ & $\mathbf{6.534}$ & $2.385$\\
HCDDPM & A & $3.181$ & $3.634$ & $6.815$ & $7.284$ & $\mathbf{3.516}$\\
HCDDPM & B & $2.721$ & $3.328$ & $6.049$ & $6.584$ & $2.030$\\
HCDDPM & C & $3.170$ & $3.625$ & $6.794$ & $7.256$ & $\mathbf{3.501}$\\
HCDDPM & D & $2.724$ & $3.327$ & $6.052$ & $6.573$ & $2.041$\\
SCDDPM & A & $3.077$ & $3.532$ & $6.609$ & $7.078$ & $3.324$\\
SCDDPM & B & $2.771$ & $3.313$ & $6.084$ & $6.604$ & $\mathbf{2.453}$\\
SCDDPM & C & $3.095$ & $3.541$ & $6.637$ & $7.091$ & $3.337$\\
SCDDPM & D & $2.773$ & $3.312$ & $6.085$ & $6.606$ & $\mathbf{2.455}$\\
\bottomrule
\end{tabular}
\end{arxivtablefit}
\end{table}
\begin{table}[H]
\caption[Fixed-range selection-sweep distances]{Distance comparisons among fully feasible designs from the four-strength selection sweep, using RFR performance-guidance strength $\gamma=0.10$. Ships were sampled separately for prescribed voyage ranges of 1,000, 5,000, and 8,000 nmi. All distances use the 133 generator-controlled parameters. Each generated subset contains 100 feasible designs; training and test references use the corresponding $\pm250$-nmi windows. Means are across 200 subsets. Within each required voyage range, bold marks the smallest reference distances and largest mean nearest-neighbor distance across architectures. Results across five independent training runs are reported separately in Table~\ref{tab_I4}.}
\label{tab_F3}
\centering\small
\begin{arxivtablefit}
\begin{tabular}{llrrrrr}
\toprule
Architecture & Range (nmi) & $d(G,R_{train})$ & $d(R_{train},G)$ & Train $D_C$ & Test $D_C$ & $D_{NN}$\\
\midrule
DDPM & 1,000 & $\mathbf{2.911}$ & $\mathbf{3.212}$ & $\mathbf{6.123}$ & $\mathbf{6.747}$ & $2.354$\\
DDPM & 5,000 & $\mathbf{2.960}$ & $\mathbf{3.218}$ & $\mathbf{6.178}$ & $\mathbf{6.785}$ & $2.380$\\
DDPM & 8,000 & $\mathbf{2.966}$ & $\mathbf{3.200}$ & $\mathbf{6.165}$ & $\mathbf{6.774}$ & $2.350$\\
HCDDPM & 1,000 & $3.218$ & $3.543$ & $6.761$ & $7.160$ & $1.932$\\
HCDDPM & 5,000 & $3.259$ & $3.539$ & $6.799$ & $7.182$ & $1.973$\\
HCDDPM & 8,000 & $3.344$ & $3.570$ & $6.914$ & $7.214$ & $1.979$\\
SCDDPM & 1,000 & $3.054$ & $3.331$ & $6.385$ & $6.836$ & $\mathbf{2.358}$\\
SCDDPM & 5,000 & $3.082$ & $3.327$ & $6.409$ & $6.879$ & $\mathbf{2.383}$\\
SCDDPM & 8,000 & $3.105$ & $3.314$ & $6.418$ & $6.881$ & $\mathbf{2.381}$\\
\bottomrule
\end{tabular}
\end{arxivtablefit}
\end{table}

Across all 72 comparison cells, no two fully feasible designs within a cell have exactly equal evaluated 133-control vectors. This rules out exact within-cell duplicates only; it does not establish physical novelty. Generated-to-training distances are generally smaller than generated-to-test distances, which is compatible with training-distribution concentration but is not a sufficient test of memorization.

\subsection{Embedding Settings}
The embedding uses 133 independent physical controls and the evaluated, cleaned vectors. Coordinates are scaled by the range over all 119,143 completed training designs, without clipping generated values. The joint visualization is fitted to 6,049 designs: 1,200 from each of the three training groups and 2,449 fully feasible generated designs across the saved training and guidance conditions. Principal component analysis (PCA) retains 50 components, explaining 78.4\% of variance in this cohort; t-SNE uses perplexity 40, PCA initialization, automatic learning rate, 1,000 iterations, and a fixed seed. The displayed subsets are selected after fitting without moving the points. This embedding scaling differs from feasible-training-only distance scaling.

Figure~\ref{fig_4} displays 512 fully feasible generated designs per model (DDPM, HCDDPM, and SCDDPM), all from feasible-only training: 103 at each of $\alpha=0$ and $0.1$, and 102 at each of $\alpha=0.3,0.5,0.7$. Each training background group displays 400 designs. Guided DDPM uses system-plus-hull classifiers, HCDDPM uses local classifiers, and SCDDPM uses system guidance. Feasible repaired rows occur only in the fully feasible background group.

\section{Locally Feasible Subsystem Training}
\label{app_G}
This appendix compares locally feasible subsystem training with global feasible-only training under matched sampling settings.

\subsection{Side-Experiment Design}
The globally feasible-only HCDDPM and SCDDPM train every denoiser on the same 11,195 globally feasible training designs. A separate completed experiment asks whether each block benefits from a larger, locally feasible training population. For subsystem $i$, its denoiser instead uses the known-label training rows for which that subsystem passes its own checks, even if another block or a system-scope check fails. The subsystem pass/fail label selects the training rows and is not supplied as a conditioning input. Each target and its HCDDPM predecessor context or SCDDPM complementary-block context still come from the same whole design. The 20 block populations range from 22,192 hopper-structure rows to 119,143 rows for each of the six hull blocks; the latter local labels pass throughout the known-label training set. Thus larger sample size is accompanied by varying degrees of cross-subsystem inconsistency.

This experiment holds the feasible-only quantile transforms, shared physical bounds, architecture, optimizer schedule, sampling seed, and pretrained feasibility classifiers fixed. It trains one new HCDDPM and one new SCDDPM with 10,000 updates per denoiser and no pass/fail conditioning input. Unguided generation and the same local, system, or combined classifier modes at $\alpha\in\{0.1,0.3,0.5,0.7\}$ yield 18 matched conditions, each with 1,000 proposals. The exact evaluator completed 17,390 of 18,000 submissions, with 6,801 fully feasible proposals. Feasibility uses the submitted denominator; violations use completed evaluations. Distances use the 133 generator-controlled parameters from completed, evaluated designs and the same 11,195-design feasible training reference. They include infeasible completed designs, unlike the fully feasible populations analyzed in Appendix~\ref{app_F}. Table~\ref{tab_G1} summarizes selected matched conditions using ten disjoint batches of 100 generation requests. Distance calculations use only completed designs within each batch; incomplete evaluations remain unsuccessful requests in the feasibility rates.

\subsection{Matched Outcomes}
Locally feasible subsystem training reduced overall feasible yield in all 18 matched conditions (Table~\ref{tab_G1}). Unguided HCDDPM fell from 20.8\% to 9.8\%, and unguided SCDDPM from 27.9\% to 11.0\%. The locally trained and guided HCDDPM reached the side experiment's highest feasible rate, 56.1\% at $\alpha=0.7$, compared with 59.7\% for its globally feasible-only counterpart. For system-guided SCDDPM at $\alpha=0.7$, the corresponding rates were 46.1\% and 60.3\%. Local-training SCDDPM did not have monotonic feasible yield at the highest strengths: its system-guided rate peaked at 49.9\% at $\alpha=0.5$, and its combined-guidance rate peaked at 45.9\% at $\alpha=0.3$.

\begin{table}[H]
\caption[Training on fully feasible ships or passing subsystem examples]{Comparison of training on fully feasible ships with training each subsystem on examples that pass its own checks. Each row holds the architecture and feasibility-guidance setting fixed while comparing the two training populations. Entries are mean $\pm$ sample SD across ten disjoint batches of 100 generation requests. Feasibility rates include incomplete evaluations as unsuccessful requests. Chamfer uses only completed, evaluated designs within each batch, so batch sizes can differ and these values are not full-population distances. Bold identifies the better observed mean within each matched pair, including the locally trained model where appropriate. Significance is reported in the text.}
\label{tab_G1}
\centering\small
\begin{arxivtablefit}
\begin{tabular}{llcccc}
\toprule
Architecture & \shortstack{Feasibility guidance,\\$\alpha$} & \multicolumn{2}{c}{Feasibility (\%) $\uparrow$} & \multicolumn{2}{c}{Chamfer $\downarrow$}\\
 & & \shortstack{Fully feasible\\ships} & \shortstack{Passing subsystem\\examples} & \shortstack{Fully feasible\\ships} & \shortstack{Passing subsystem\\examples}\\
\midrule
HCDDPM & None, 0 & $\mathbf{20.8\pm3.9}$ & $9.8\pm2.0$ & $\mathbf{6.841\pm0.035}$ & $7.237\pm0.031$\\
HCDDPM & Local, 0.7 & $\mathbf{59.7\pm5.9}$ & $56.1\pm4.0$ & $\mathbf{6.046\pm0.021}$ & $6.095\pm0.015$\\
SCDDPM & None, 0 & $\mathbf{27.9\pm4.4}$ & $11.0\pm3.9$ & $\mathbf{6.634\pm0.034}$ & $7.113\pm0.025$\\
SCDDPM & Local, 0.7 & $\mathbf{59.2\pm3.2}$ & $42.5\pm4.5$ & $6.094\pm0.024$ & $\mathbf{6.016\pm0.016}$\\
SCDDPM & System, 0.5 & $\mathbf{58.0\pm5.4}$ & $49.9\pm4.1$ & $\mathbf{6.216\pm0.034}$ & $6.262\pm0.022$\\
SCDDPM & System, 0.7 & $\mathbf{60.3\pm4.6}$ & $46.1\pm4.9$ & $6.085\pm0.022$ & $\mathbf{6.002\pm0.018}$\\
SCDDPM & Combined, 0.3 & $\mathbf{52.4\pm5.3}$ & $45.9\pm3.8$ & $\mathbf{6.400\pm0.030}$ & $6.611\pm0.025$\\
SCDDPM & Combined, 0.7 & $\mathbf{60.0\pm2.7}$ & $42.6\pm4.9$ & $6.094\pm0.024$ & $\mathbf{6.016\pm0.017}$\\
\bottomrule
\end{tabular}
\end{arxivtablefit}
\end{table}

For each outcome, two-sided Welch tests compare the ten subset estimates of the two training strategies at all 18 matched architecture, guidance, and strength conditions; Holm adjustment covers those 18 tests within that outcome. Local training has significantly lower feasible yield in 16 of 18 matches (adjusted $p<0.05$); the HCDDPM comparisons at $\alpha=0.3$ and $0.7$ are the exceptions. It has more mean counted violations in every match, significant in 17, with HCDDPM at $\alpha=0.7$ the exception. These tests characterize conditional variation across saved subsets, not across independent generator fits.

Completed-design reference distances and mean nearest-neighbor distances do not follow feasible yield identically. Local training has higher subset Chamfer in 15 matches and lower Chamfer in 3; 18 of the 18 differences are significant after adjustment. For nearest-neighbor distance, 11 matches significantly favor local training and 6 significantly favor global feasible-only training, with 1 not significant. These distances include infeasible completed designs and depend on the number of completed designs in each batch. Lower Chamfer does not offset lower feasible yield, and smaller mean nearest-neighbor distance indicates greater concentration. One training and one sampling seed per strategy preclude claims about repeatability across independently trained models.

\section{Exploratory Free-Range Performance Guidance}
\label{app_H}
This appendix reports the exploratory free-range performance-guidance protocol, economic tradeoffs, and selected feasible ship examples.

\subsection{Performance Regression Models, Objectives, and Paired Cohorts}
The performance study reuses the feasible-only generator weights and their transforms. Separate timestep-conditioned performance regression models predict RFR alone or five responses: acquisition cost, annual total cost, RFR, steel weight, and resistance. They train on all 119,143 completed training designs, including infeasible designs; incomplete evaluations supply no regression targets. Each performance response is transformed using $\log(1+y)$, then centered at its training-set median and divided by its training-set interquartile range, the difference between the 75th and 25th percentiles. This compresses large values and places the objectives on comparable numerical scales for regression training. The same fitted transformation is applied to validation and test responses. Full-state performance regression models serve the monolithic DDPM and SCDDPM. HCDDPM uses a level-aware performance regression model whose inputs include clean predecessors, the noisy current block, a visibility mask, the block index, and timestep. Unknown future blocks are not treated as completed geometry.

The performance regression models use a 128-dimensional timestep embedding and hidden widths 384, 384, and 256. Equation~\eqref{eq_H1} defines the preference weights and performance direction for the five-objective arm; one preference vector is sampled per design and held fixed throughout generation:
\begin{equation}
 u_j\sim\mathcal U(0,1),\qquad
 w_j=\frac{u_j}{\sum_{k=1}^{5}u_k},\qquad
 \mathbf g_P=-\sum_{j=1}^{5}w_j\mathcal N(\nabla\widehat h_j).
 \label{eq_H1}
\end{equation}
Here $\mathcal N$ normalizes each objective gradient over the coordinates that can change in that step; a negligible gradient contributes zero. Each design receives a random set of five objective weights that sum to one. These weights stay fixed throughout generation and specify the relative priority of each objective. Some combinations of priorities are more likely than others under this sampling procedure, so it does not explore all trade-offs uniformly. Feasibility and performance strengths are separate, and their combined displacement is capped. SCDDPM uses subsystem-level feasibility guidance in this study and applies a separate full-state performance correction once to the assembled system. HCDDPM changes only the active block using its prefix performance regression model.

Twenty-design pilot arms selected settings for independent 100-design validation cohorts. The validation compares feasibility guidance alone, added RFR performance guidance, and added five-objective performance guidance for each architecture, with paired initial noise within each study. Feasibility strength is 0.7 throughout. The selected performance strengths and windows are: monolithic RFR 0.03 over the final 100 steps; monolithic five-objective 0.10 over the final 100 steps; HCDDPM RFR and five-objective 0.10 over the final 50 steps; and SCDDPM RFR and five-objective 0.10 over the final 100 steps. HCDDPM RFR performance guidance acts at all levels; in its five-objective arm, the RFR contribution begins at the tenth subsystem. A separate 100-design-per-arm HCDDPM refinement examines smaller five-objective strengths. These cohorts share one trained generator per architecture.

Performance regression model quality varies with noise and available subsystem context. At schedule indices $k=0,50,100,200$ (mathematical times $t=1,51,101,201$), the full-state RFR regression model's held-out feasible-design rank correlations are 0.963, 0.809, 0.542, and 0.143, respectively. At $k=50$, the prefix performance regression model's corresponding correlation is approximately 0.006 at the first subsystem and 0.971 at the tenth. These correlations describe prediction on the held-out diagnostic cohort.

\subsection{Validation Results and Mission Tradeoffs}
The nine validation arms submitted 900 designs; 899 completed. The only incomplete evaluation occurred in the HCDDPM feasibility-only baseline. Table~\ref{tab_H1} reports each arm's feasible population separately from paired RFR change. The latter is the median of design-level percentage changes among pairs in which both evaluations completed, including infeasible pairs; it is not the percentage difference between feasible-population medians.

\begin{table}[H]
\caption[Exploratory performance-guidance validation]{Independent sampling-cohort validation of performance guidance. Every arm submitted 100 designs; overall feasibility uses all 100. RFR medians use feasible designs; paired changes use jointly completed pairs. Bold marks the highest feasibility rate and lowest RFR and paired RFR change across architectures within each guidance arm, including ties; counts are unranked.}
\label{tab_H1}
\centering\small
\begin{arxivtablefit}
\begin{tabular}{llrrrrr}
\toprule
Architecture & Performance guidance & Completed & Feasible $\uparrow$ & Feasibility (\%) $\uparrow$ & Median RFR (\pounds/t) $\downarrow$ & Paired RFR change (\%) $\downarrow$\\
\midrule
DDPM & None & 100 & 56 & 56.0 & \textbf{23.68} & \textbf{0.0}\\
DDPM & RFR & 100 & 56 & 56.0 & 20.74 & -11.3\\
DDPM & Five objectives & 100 & 63 & \textbf{63.0} & \textbf{22.53} & \textbf{-3.6}\\
HCDDPM & None & 99 & 58 & 58.0 & 29.79 & \textbf{0.0}\\
HCDDPM & RFR & 100 & 60 & 60.0 & 19.96 & -32.1\\
HCDDPM & Five objectives & 100 & 43 & 43.0 & 26.42 & -2.0\\
SCDDPM & None & 100 & 66 & \textbf{66.0} & 26.73 & \textbf{0.0}\\
SCDDPM & RFR & 100 & 62 & \textbf{62.0} & \textbf{15.29} & \textbf{-40.7}\\
SCDDPM & Five objectives & 100 & 59 & 59.0 & 25.34 & -3.3\\
\bottomrule
\end{tabular}
\end{arxivtablefit}
\end{table}

RFR performance guidance reduced paired median freight rate by 11.3\%, 32.1\%, and 40.7\% for monolithic DDPM, HCDDPM, and SCDDPM. These improvements involved substantial changes to the generated mission (Table~\ref{tab_H2}). Paired annual cargo increased by 13.7\%, 52.6\%, and 76.9\%, while voyage range decreased by 8.6\%, 17.5\%, and 39.2\%. For SCDDPM, annual cost increased by 5.2\% and resistance by 9.2\% despite lower RFR. Equation~\eqref{eq_3} explains why increased transport throughput can reduce the economic ratio without reducing all physical costs. These results do not demonstrate improved performance at fixed speed, range, or cargo service.

\begin{table}[H]
\caption[Free-range paired response changes]{Free-range paired median response changes (\%). Each cell is a median of pairwise percentage changes among jointly completed baseline/guided pairs, including infeasible pairs; 100 pairs are available for DDPM and SCDDPM and 99 for HCDDPM. Changes in mission and throughput preclude a fixed-service interpretation. Within each guidance arm, bold marks the smallest RFR, cost, and resistance changes and the largest cargo change across architectures; voyage range is unranked.}
\label{tab_H2}
\centering\small
\begin{arxivtablefit}
\begin{tabular}{llrrrrrr}
\toprule
Architecture & Guidance & RFR & Acquisition & Annual cost & Cargo & Resistance & Range\\
\midrule
DDPM & RFR & -11.3 & \textbf{+0.3} & \textbf{+1.0} & +13.7 & \textbf{+1.6} & -8.6\\
DDPM & Five objectives & \textbf{-3.6} & -9.4 & -9.9 & \textbf{-7.2} & -17.0 & -1.6\\
HCDDPM & RFR & -32.1 & +2.2 & +4.4 & +52.6 & +7.0 & -17.5\\
HCDDPM & Five objectives & -2.0 & \textbf{-23.1} & \textbf{-22.1} & -21.5 & \textbf{-31.0} & +4.5\\
SCDDPM & RFR & \textbf{-40.7} & +0.7 & +5.2 & \textbf{+76.9} & +9.2 & -39.2\\
SCDDPM & Five objectives & -3.3 & -10.5 & -12.0 & -9.0 & -22.8 & -0.2\\
\bottomrule
\end{tabular}
\end{arxivtablefit}
\end{table}

Five-objective performance guidance reduced paired acquisition cost by 9.4\%, 23.1\%, and 10.5\%, steel weight by 10.2\%, 30.2\%, and 10.1\%, and resistance by 17.0\%, 31.0\%, and 22.8\% for DDPM, HCDDPM, and SCDDPM, respectively. However, annual cargo also decreased by 7.2\%, 21.5\%, and 9.0\%. For HCDDPM, the bootstrap interval for the RFR change spans $-6.9\%$ to $+0.5\%$ and includes zero. These results do not establish a reduction in modeled RFR. Fully feasible yield was 43\%, with a 95\% Wilson lower confidence bound of 33.7\%.

Restricting RFR comparisons to designs feasible in both paired arms leaves 55, 36, and 59 pairs for the three architectures, with median RFR changes of $-11.3\%$, $-32.3\%$, and $-40.8\%$. This sensitivity analysis supports the direction of the RFR result among retained feasible pairs, but guidance also changes which designs become feasible. It cannot replace the full population accounting.

The separate HCDDPM refinement completed 300 evaluations. Feasibility was 52/100 for feasibility guidance alone, 57/100 for five-objective strength 0.03, and 48/100 for strength 0.05. At 0.03, paired median acquisition cost, annual cost, steel weight, and resistance decreased by 7.3\%, 7.4\%, 9.8\%, and 10.8\%; RFR changed by only $-0.4\%$, with a bootstrap interval of $-2.3\%$ to $+0.6\%$. The smaller strength was selected after inspecting this cohort and requires independent confirmation.

The complete exploratory program used 2,000 evaluations, including a superseded 100-design pilot affected by an arithmetic error. The corrected analysis includes 1,900 designs, comprising 700 pilot, 900 validation, and 300 refinement observations. Pilot, validation, and refinement results are not pooled. Pareto coverage and generated-set distance metrics were not measured for this initial free-range validation cohort; the two-arm fixed-range study reports both for its separate production samples.

\subsection{Selected Feasible Design Examples}
\begin{figure}[H]
\centering\includegraphics[width=\textwidth]{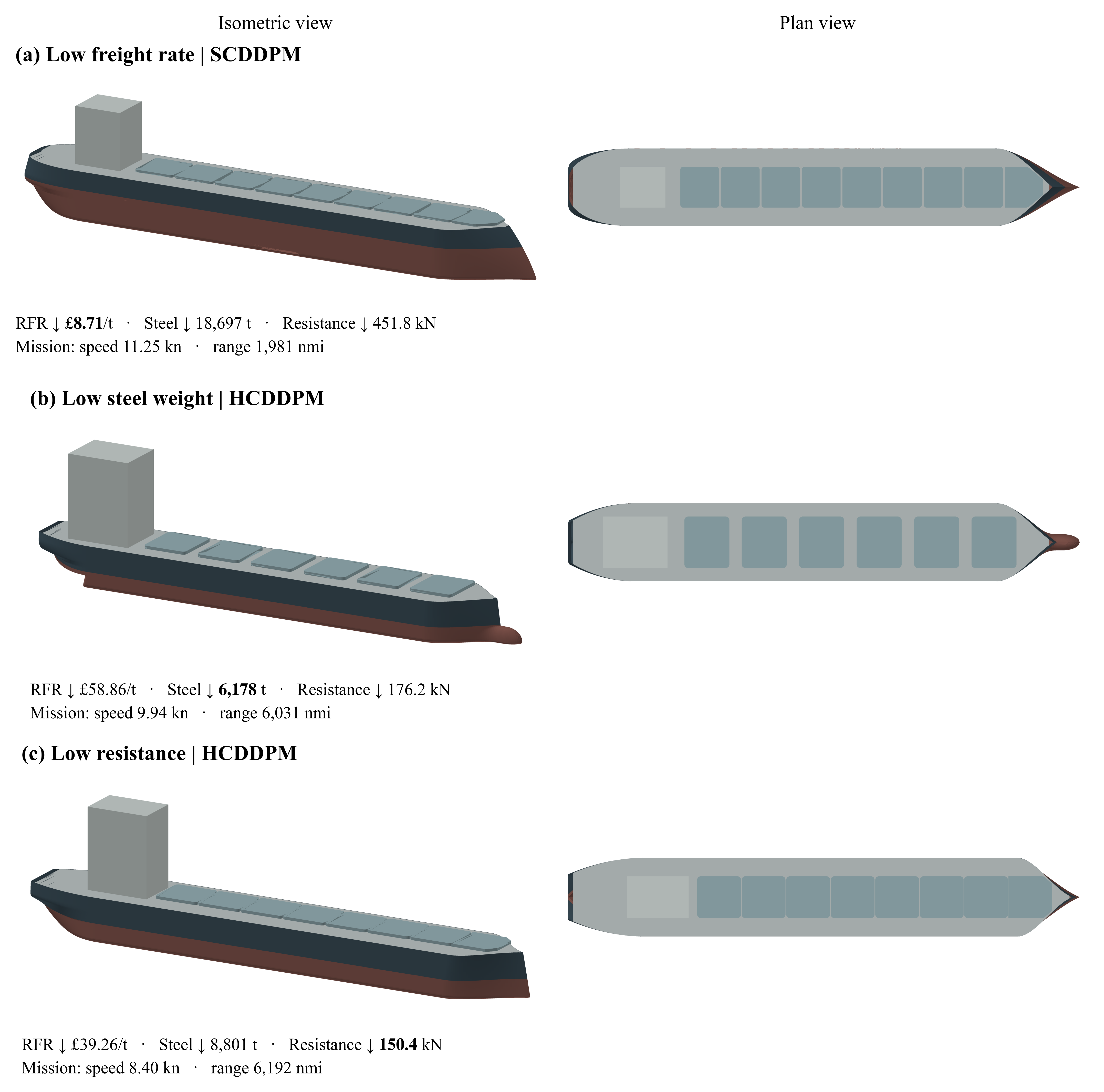}
\caption[Selected bulk-carrier CAD views]{Starboard isometric and plan views of generated examples selected for low freight rate, low steel weight, and low resistance within the feasible performance-guided validation cohort. Evaluated hull, subdivision, and representative hatch-opening parameters inform the scenes; the opening dimensions are copied to each hold for preliminary cover-mass estimation and illustrative cover envelopes, not a designed per-hold hatch layout. The single-block deckhouse is a display-only proxy with no assigned mass; cargo holds are empty. Forward cover envelopes taper for display where needed. This changes neither the assumed cover mass nor the counted feasibility labels and does not verify a hatch-cover design. Images are fitted independently to their panels and do not share a physical scale. Bold response values identify the best among the three illustrated examples; speed and range remain unranked mission context.}
\label{fig_H1}
\end{figure}

Figure~\ref{fig_H1} illustrates three fully feasible generated designs selected to span freight rate, steel weight, and resistance. Candidates are ranked separately by each response among the 343 feasible designs in the six performance-guided validation arms; feasibility-only baselines, pilots, and refinement cohorts are excluded. The freight-rate and steel-weight examples have the lowest respective responses in this cohort. The second-lowest-resistance feasible design is shown in the illustration. The lowest-resistance design was omitted because its representative hatch-cover depiction extends up to 0.124\,m beyond the illustrated hold boundaries. This display rule does not change its engineering feasibility classification. The freight-rate example comes from performance-guided SCDDPM, while the other two come from five-objective HCDDPM. They are selected observations within the evaluated cohort, not established optima over the design space.

Table~\ref{tab_H3} reports their responses and operating conditions. The low-resistance example operates at 8.40\,kn, compared with 11.25\,kn for the low-freight-rate example, and the ranges differ by more than a factor of three. These operating differences motivate the later fixed-range comparison and explicit reporting of speed and cargo throughput. Feasibility here means that the formulated engineering constraints were satisfied in the saved evaluation.

\begin{table}[H]
\caption[Selected feasible generated examples]{Selected feasible generated examples. Bold identifies the lowest displayed response in each objective column. Speed and range are mission context; these examples are not fixed-service comparisons. Labels correspond to Figure~\ref{fig_H1}.}
\label{tab_H3}
\centering\small
\begin{arxivtablefit}
\begin{tabular}{cllrrrrr}
\toprule
Example & Selection criterion & Architecture & RFR (\pounds/t) $\downarrow$ & Steel (t) $\downarrow$ & Resistance (kN) $\downarrow$ & Speed (kn) & Range (nmi)\\
\midrule
(a) & Low freight rate & SCDDPM & \textbf{8.71} & 18,697 & 451.8 & 11.25 & 1,981\\
(b) & Low steel weight & HCDDPM & 58.86 & \textbf{6,178} & 176.2 & 9.94 & 6,031\\
(c) & Low resistance & HCDDPM & 39.26 & 8,801 & \textbf{150.4} & 8.40 & 6,192\\
\bottomrule
\end{tabular}
\end{arxivtablefit}
\end{table}

\section{Independent Retraining Confirmation at 5,000 Nautical Miles}
\label{app_I}
This appendix records the outcomes across five independent training runs, statistical contrasts, RFR-reference sensitivity, feasible-set distances, and computational costs.

The confirmation contains 15 fresh complete generator fits, 30 fit/arm cells, and 300 finalized 100-request batches. All 30,000 attempts remain in the yield denominators, including 30 incomplete evaluations. The older selection fit and workflow-validation cases are excluded. Every requested candidate receives the normal complete ship-system evaluation; 29,970 evaluations complete and 17,110 designs satisfy the formulated engineering constraints. No fully feasible output lacks finite RFR. Shared data and frozen guidance assets make the fitted generators independent training replicates conditional on those common assets; individual ships and subset redraws are not additional training replicates. Table~\ref{tab_I1} shows all fit/arm outcomes, and Table~\ref{tab_I2} shows the five primary contrasts.

\begingroup\small
\begin{arxivlongtable}
\begin{longtable}{llrrrrr}
\caption[Individual outcomes of all fresh generator fits]{Individual fit/arm outcomes at 5,000 nmi. Each row requests 1,000 proposals. F denotes feasibility-only guidance ($\gamma=0$); FR adds the selected RFR strength (DDPM/SCDDPM 0.10, HCDDPM 0.03). Both arms use $\alpha=0.7$. A qualifying design satisfies the formulated engineering constraints and has RFR at most 24.5577\,\pounds/t. RFR is the mean among that row's fully feasible designs, all of which have finite RFR. Within each architecture and fit, bold marks the lower mean RFR of the two guidance arms; counts are unranked.}\label{tab_I1}\\
\toprule
Architecture & Fit & Arm & Completed & Feasible & Qualifying count & Mean RFR (\pounds/t)\\
\midrule
\endfirsthead
\toprule
Architecture & Fit & Arm & Completed & Feasible & Qualifying count & Mean RFR (\pounds/t)\\
\midrule
\endhead
DDPM & 1 & F & 1000 & 566 & 375 & 23.67\\
DDPM & 1 & FR & 1000 & 529 & 527 & \textbf{17.63}\\
DDPM & 2 & F & 1000 & 547 & 322 & 24.75\\
DDPM & 2 & FR & 1000 & 497 & 492 & \textbf{18.01}\\
DDPM & 3 & F & 1000 & 556 & 325 & 24.92\\
DDPM & 3 & FR & 1000 & 525 & 520 & \textbf{18.12}\\
DDPM & 4 & F & 1000 & 568 & 322 & 24.96\\
DDPM & 4 & FR & 1000 & 502 & 493 & \textbf{18.28}\\
DDPM & 5 & F & 1000 & 554 & 348 & 24.15\\
DDPM & 5 & FR & 1000 & 526 & 523 & \textbf{17.85}\\
HCDDPM & 1 & F & 1000 & 575 & 85 & 32.62\\
HCDDPM & 1 & FR & 998 & 621 & 471 & \textbf{22.89}\\
HCDDPM & 2 & F & 997 & 675 & 59 & 33.67\\
HCDDPM & 2 & FR & 999 & 688 & 481 & \textbf{23.32}\\
HCDDPM & 3 & F & 999 & 589 & 69 & 32.37\\
HCDDPM & 3 & FR & 999 & 599 & 454 & \textbf{22.33}\\
HCDDPM & 4 & F & 999 & 567 & 81 & 32.38\\
HCDDPM & 4 & FR & 998 & 573 & 443 & \textbf{22.50}\\
HCDDPM & 5 & F & 999 & 586 & 193 & 28.22\\
HCDDPM & 5 & FR & 999 & 582 & 528 & \textbf{20.46}\\
SCDDPM & 1 & F & 996 & 546 & 219 & 27.81\\
SCDDPM & 1 & FR & 999 & 526 & 525 & \textbf{18.31}\\
SCDDPM & 2 & F & 999 & 545 & 307 & 25.04\\
SCDDPM & 2 & FR & 1000 & 549 & 548 & \textbf{18.52}\\
SCDDPM & 3 & F & 994 & 587 & 217 & 28.31\\
SCDDPM & 3 & FR & 1000 & 574 & 567 & \textbf{19.21}\\
SCDDPM & 4 & F & 998 & 597 & 236 & 27.97\\
SCDDPM & 4 & FR & 1000 & 593 & 582 & \textbf{19.12}\\
SCDDPM & 5 & F & 997 & 587 & 160 & 29.52\\
SCDDPM & 5 & FR & 1000 & 581 & 563 & \textbf{19.14}\\
\bottomrule
\end{longtable}
\end{arxivlongtable}
\endgroup

\begin{table}[H]
\caption[Five prespecified confirmation contrasts]{All five primary comparisons of qualifying yield at the RFR reference value. Within-architecture FR--F contrasts use paired $t$ tests across five complete generator fits (four degrees of freedom); selected-pipeline contrasts use independent-sample Welch tests. Effects and ordinary, unadjusted 95\% intervals are percentage points. Holm correction covers this single family of five tests.}
\label{tab_I2}
\centering\small
\begin{arxivtablefit}
\begin{tabular}{lrrr}
\toprule
Contrast & Gain (pp) & 95\% interval (pp) & $p_{\mathrm{Holm}}$\\
\midrule
DDPM (FR--F) & 17.26 & [15.36, 19.16] & $6.28\times10^{-5}$\\
HCDDPM (FR--F) & 37.80 & [33.80, 41.80] & $6.28\times10^{-5}$\\
SCDDPM (FR--F) & 32.92 & [25.45, 40.39] & $7.69\times10^{-4}$\\
SCDDPM--HCDDPM (FR) & 8.16 & [3.99, 12.33] & $4.92\times10^{-3}$\\
SCDDPM--DDPM (FR) & 4.60 & [1.73, 7.47] & $6.30\times10^{-3}$\\
\bottomrule
\end{tabular}
\end{arxivtablefit}
\end{table}

\subsection{Sensitivity to the RFR Reference Value}
The fixed RFR reference values derived from the feasible training window are 24.5577, 33.3885, and 52.4106\,\pounds/t for the lower quartile, median, and upper quartile, respectively. Table~\ref{tab_I3} reports qualifying yields at all three reference values using the same requests and fits. At the higher median reference value, HCDDPM has the highest qualifying yield (60.10\%, versus 56.46\% for SCDDPM and 51.58\% for DDPM). The preferred method therefore depends on the reference value used for comparison. The median and upper-quartile comparisons are descriptive sensitivity outcomes, not additional prespecified significance tests.

\begin{table}[H]
\caption[Confirmation sensitivity to the RFR reference value]{Qualifying yields (\%), mean $\pm$ between-fit sample SD across five fresh generator fits. The primary, median, and upper-quartile reference values are fixed from the feasible 5,000-nmi training window; the primary reference is the lower quartile. All requests remain in every denominator; F and FR denote the same arms as in Table~\ref{tab_I1}. Bold marks the highest mean yield across architectures within each guidance arm and reference value, including ties.}
\label{tab_I3}
\centering\small
\begin{arxivtablefit}
\begin{tabular}{llrrr}
\toprule
Architecture & Arm & \shortstack{Primary reference\\yield (\%)} & \shortstack{Median reference\\yield (\%)} & \shortstack{Upper-quartile reference\\yield (\%)}\\
\midrule
DDPM & F & $\mathbf{33.84\pm2.32}$ & $\mathbf{52.38\pm1.42}$ & $55.56\pm0.83$\\
DDPM & FR & $51.10\pm1.71$ & $51.58\pm1.51$ & $51.58\pm1.51$\\
HCDDPM & F & $9.74\pm5.44$ & $39.04\pm5.06$ & $\mathbf{58.30\pm4.16}$\\
HCDDPM & FR & $47.54\pm3.29$ & $\mathbf{60.10\pm4.09}$ & $\mathbf{61.18\pm4.58}$\\
SCDDPM & F & $22.78\pm5.28$ & $47.38\pm2.17$ & $56.58\pm2.29$\\
SCDDPM & FR & $\mathbf{55.70\pm2.16}$ & $56.46\pm2.69$ & $56.46\pm2.69$\\
\bottomrule
\end{tabular}
\end{arxivtablefit}
\end{table}

\subsection{Feasible-Set Nearest-Neighbor Distance and Reference Proximity}
Distances use the definitions in Appendix~\ref{app_F}. A common generated subset size of 100 is retained across all 30 cells, with 200 subsets selected from already generated feasible designs in each cell. Scaling uses only the original 11,195 feasible training designs, and reference comparison uses the 5,000-nmi windows (603 training and 135 held-out test designs). Variation across the 200 subsets measures sensitivity to which designs are included; Table~\ref{tab_I4} instead reports between-fit variation after averaging the subset results within each fit. The selected SCDDPM mean nearest-neighbor distance exceeds HCDDPM by 20.3\% using the 133 generator-controlled parameters. Its held-out bidirectional Chamfer distance does not improve on HCDDPM, so a larger mean nearest-neighbor distance should not be interpreted as closer agreement with the held-out reference designs. These secondary comparisons are descriptive and are not among the five planned statistical tests.

\begin{table}[H]
\caption[Confirmation feasible-set distances across fresh fits]{Equal-sized feasible-set distances at a prescribed voyage range of 5,000 nmi, using the evaluated, cleaned values of the 133 generator-controlled parameters. Each fit/arm uses 200 subsets of 100 already generated feasible designs; subset means are calculated within each fit before averaging over the five independent fits. Entries are mean $\pm$ between-fit sample SD. Directional columns use the 603-design feasible training-window reference; test $D_C$ uses the 135-design source-parent-separated test-window reference. F and FR are defined in Table~\ref{tab_I1}. Scaling uses feasible training designs only. Within each guidance arm, bold marks the smallest mean reference distances and largest mean nearest-neighbor distance across architectures.}
\label{tab_I4}
\centering\small
\setlength{\tabcolsep}{4pt}
\begin{arxivtablefit}
\begin{tabular}{llrrrrr}
\toprule
Architecture & Arm & $d(G,R_{train})$ & $d(R_{train},G)$ & Train $D_C$ & Test $D_C$ & $D_{NN}$\\
\midrule
DDPM & F & $\mathbf{2.912\pm0.017}$ & $\mathbf{3.185\pm0.014}$ & $\mathbf{6.097\pm0.029}$ & $\mathbf{6.772\pm0.028}$ & $2.378\pm0.033$\\
DDPM & FR & $\mathbf{2.908\pm0.010}$ & $\mathbf{3.205\pm0.009}$ & $\mathbf{6.113\pm0.017}$ & $\mathbf{6.784\pm0.030}$ & $2.324\pm0.032$\\
HCDDPM & F & $3.039\pm0.029$ & $3.351\pm0.023$ & $6.390\pm0.039$ & $6.829\pm0.032$ & $1.964\pm0.067$\\
HCDDPM & FR & $3.074\pm0.021$ & $3.374\pm0.024$ & $6.449\pm0.041$ & $6.868\pm0.043$ & $1.976\pm0.078$\\
SCDDPM & F & $3.054\pm0.013$ & $3.296\pm0.011$ & $6.350\pm0.023$ & $6.863\pm0.012$ & $\mathbf{2.414\pm0.022}$\\
SCDDPM & FR & $3.059\pm0.013$ & $3.312\pm0.012$ & $6.371\pm0.022$ & $6.874\pm0.024$ & $\mathbf{2.377\pm0.031}$\\
\bottomrule
\end{tabular}
\end{arxivtablefit}
\end{table}

\subsection{Training Work and Failure Families}
Table~\ref{tab_I5} shows generator capacities and recorded training work.
\begin{table}[H]
\caption[Generator capacities and recorded fresh-training work]{Generator capacity and recorded fresh-training work. Each denoiser receives 10,000 optimizer updates; summed updates count all denoisers in a complete fit. Training time is the cumulative time per complete fresh fit, mean $\pm$ between-fit sample SD in seconds. It excludes setup, sampling, and full engineering evaluation. Bold marks the lowest recorded mean training time; network sizes and update counts are unranked.}
\label{tab_I5}
\centering\small
\begin{arxivtablefit}
\begin{tabular}{lrrrr}
\toprule
Architecture & Denoisers & Parameters & \shortstack{Summed\\updates} & Training time (s)\\
\midrule
DDPM & 1 & 3,883,141 & 10,000 & $\mathbf{16.80\pm0.22}$\\
HCDDPM & 20 & 10,888,069 & 200,000 & $365.33\pm18.50$\\
SCDDPM & 20 & 11,303,301 & 200,000 & $376.01\pm3.30$\\
\bottomrule
\end{tabular}
\end{arxivtablefit}
\end{table}

Every output coordinate belongs to a denoiser trained for 10,000 updates. The 200,000 summed updates for a subsystem architecture count 20 separately updated networks; they do not represent 20 times as many updates to each output coordinate. Parameter counts and total computational work nevertheless differ from the monolithic model. All 15 training runs completed. The reported training times cover each complete training run excluding setup, sampling, and engineering evaluation. The recorded training times are hardware- and implementation-dependent observations, not a general computational-efficiency ranking; resumed-run timing measurements are not pooled with them.

\begin{table}[H]
\caption[Recorded cost per qualifying candidate]{Recorded seconds per fully feasible candidate with RFR at or below the fixed reference value at 5,000 nmi. Entries are mean $\pm$ between-fit sample SD across five fresh fits, for the selected feasibility-plus-RFR arms. Bold marks the lowest mean time in each column across architectures.}
\label{tab_I6}
\centering\small
\begin{arxivtablefit}
\begin{tabular}{lrrrr}
\toprule
Architecture & Sampling & Decoding & Evaluation & Total\\
\midrule
DDPM & $\mathbf{0.0303\pm0.0014}$ & $0.0035\pm0.0002$ & $1.849\pm0.083$ & $\mathbf{1.882\pm0.084}$\\
HCDDPM & $0.5179\pm0.0308$ & $0.0036\pm0.0002$ & $2.101\pm0.138$ & $2.622\pm0.169$\\
SCDDPM & $0.3338\pm0.0125$ & $\mathbf{0.0031\pm0.0001}$ & $\mathbf{1.778\pm0.068}$ & $2.115\pm0.081$\\
\bottomrule
\end{tabular}
\end{arxivtablefit}
\end{table}

Recorded online cost is the sum of sampling, decoding, and engineering-evaluation wall time divided by the number of fully feasible outputs with RFR at or below the fixed reference value in each fit. Table~\ref{tab_I6} averages these per-fit ratios equally. Sampling used the recorded NVIDIA RTX~4090 setup, while engineering evaluation used six central processing unit (CPU) workers. Evaluation entries are elapsed wall time, not summed worker time. Generator training, frozen guidance-model development, and setup outside the recorded phases are excluded. Under this accounting, SCDDPM takes 12.4\% more time per qualifying output than DDPM and 19.3\% less than HCDDPM; these are descriptive hardware-specific observations. The SD of the total is calculated from each fit's total, rather than by adding phase SDs.

Local structural checks remain the dominant failure family at the selected performance settings. Mean failure rates among designs evaluated against each constraint family are 46.36\% for DDPM, 35.15\% for HCDDPM, and 42.33\% for SCDDPM; the corresponding hull-geometry rates are 3.04\%, 5.20\%, and 1.90\%. A design can fail multiple families, and each family uses the designs evaluated against its constraints as the denominator, so these rates must not be added or read as a decomposition of total infeasibility. These diagnostics identify where the implemented pipelines fail without isolating a causal architectural mechanism.

\subsection{Engineering Properties of Fully Feasible Designs}
Table~\ref{tab_I7} compares the engineering properties of all 17,110 fully feasible confirmation outputs across architectures and guidance arms. Each fit first supplies a mean over its feasible designs; five equally weighted fit means then supply the reported mean and between-fit sample SD. Dimension ratios are formed per design before averaging. No RFR reference-value filter is applied, so the table describes the feasible populations accompanying the per-request yields rather than only the qualifying subset.

\begin{table}[H]
\caption[Engineering properties across architectures and guidance arms]{Engineering properties of all fully feasible confirmation outputs at 5,000 nmi, with no RFR filter. Entries are means $\pm$ sample SD of five fit-level means. F denotes feasibility guidance with reduced noise; FR adds the selected RFR performance guidance. Each architecture/arm requested 5,000 designs. Bold marks the lowest mean RFR across architectures within each guidance arm; other rows describe the generated vessel populations.}
\label{tab_I7}
\centering\footnotesize
\setlength{\tabcolsep}{2.5pt}
\begin{arxivtablefit}
\begin{tabular}{lrrrrrr}
\toprule
& \multicolumn{2}{c}{DDPM} & \multicolumn{2}{c}{HCDDPM} & \multicolumn{2}{c}{SCDDPM}\\
\cmidrule(lr){2-3}\cmidrule(lr){4-5}\cmidrule(lr){6-7}
Property & F & FR & F & FR & F & FR\\
\midrule
Length, LOA (m) & $176.2\,\pm\,1.6$ & $168.3\,\pm\,1.1$ & $172.9\,\pm\,0.6$ & $174.6\,\pm\,0.6$ & $181.0\,\pm\,1.2$ & $173.3\,\pm\,1.5$\\[2pt]
Beam (m) & $27.1\,\pm\,0.1$ & $25.9\,\pm\,0.1$ & $26.7\,\pm\,0.1$ & $27.1\,\pm\,0.1$ & $27.4\,\pm\,0.1$ & $26.2\,\pm\,0.1$\\[2pt]
Draft (m) & $7.94\,\pm\,0.09$ & $8.88\,\pm\,0.08$ & $7.01\,\pm\,0.18$ & $8.49\,\pm\,0.18$ & $7.88\,\pm\,0.22$ & $9.22\,\pm\,0.10$\\[2pt]
Depth (m) & $15.54\,\pm\,0.10$ & $14.80\,\pm\,0.07$ & $15.33\,\pm\,0.05$ & $15.76\,\pm\,0.07$ & $15.82\,\pm\,0.07$ & $15.11\,\pm\,0.10$\\[2pt]
Length / beam & $6.51\,\pm\,0.05$ & $6.51\,\pm\,0.04$ & $6.49\,\pm\,0.01$ & $6.44\,\pm\,0.01$ & $6.62\,\pm\,0.06$ & $6.63\,\pm\,0.07$\\[2pt]
Length / depth & $11.37\,\pm\,0.04$ & $11.40\,\pm\,0.03$ & $11.30\,\pm\,0.02$ & $11.09\,\pm\,0.02$ & $11.47\,\pm\,0.12$ & $11.49\,\pm\,0.14$\\[2pt]
Beam / draft & $3.45\,\pm\,0.03$ & $2.93\,\pm\,0.03$ & $3.85\,\pm\,0.10$ & $3.23\,\pm\,0.08$ & $3.52\,\pm\,0.09$ & $2.85\,\pm\,0.02$\\[2pt]
\midrule
Displacement (t) & $35{,}894\,\pm\,799$ & $36{,}608\,\pm\,641$ & $29{,}849\,\pm\,612$ & $37{,}193\,\pm\,840$ & $36{,}687\,\pm\,1{,}181$ & $39{,}396\,\pm\,573$\\[2pt]
Structural weight (t) & $18{,}263\,\pm\,533$ & $14{,}726\,\pm\,372$ & $16{,}802\,\pm\,264$ & $17{,}949\,\pm\,413$ & $19{,}855\,\pm\,273$ & $16{,}971\,\pm\,239$\\[2pt]
Deadweight (t) & $15{,}817\,\pm\,401$ & $20{,}225\,\pm\,387$ & $11{,}263\,\pm\,817$ & $17{,}393\,\pm\,904$ & $14{,}895\,\pm\,994$ & $20{,}648\,\pm\,533$\\[2pt]
Outfit weight (t) & $986\,\pm\,10$ & $912\,\pm\,7$ & $950\,\pm\,5$ & $975\,\pm\,6$ & $1{,}019\,\pm\,4$ & $945\,\pm\,6$\\[2pt]
\midrule
Block coefficient & $0.903\,\pm\,0.002$ & $0.904\,\pm\,0.002$ & $0.892\,\pm\,0.001$ & $0.891\,\pm\,0.001$ & $0.906\,\pm\,0.002$ & $0.908\,\pm\,0.001$\\[2pt]
Speed (kn) & $11.37\,\pm\,0.07$ & $10.96\,\pm\,0.06$ & $11.69\,\pm\,0.20$ & $11.67\,\pm\,0.23$ & $11.86\,\pm\,0.12$ & $11.46\,\pm\,0.05$\\[2pt]
RFR (\pounds{}/t) & $\mathbf{24.49\,\pm\,0.56}$ & $\mathbf{17.98\,\pm\,0.25}$ & $31.85\,\pm\,2.10$ & $22.30\,\pm\,1.10$ & $27.73\,\pm\,1.65$ & $18.86\,\pm\,0.41$\\[2pt]
\midrule
Fully feasible designs & 2,791 & 2,579 & 2,992 & 3,063 & 2,862 & 2,823\\
\bottomrule
\end{tabular}
\end{arxivtablefit}
\begin{minipage}{\textwidth}\footnotesize
Length is LOA; beam and depth are modeled dimensions; draft and speed are the selected operating values. Displacement and deadweight use full-cargo departure. All masses are metric tonnes. Structural steel excludes separately modeled hatch covers; outfit is separate from machinery. The recorded block coefficient uses LOA, full design beam, and operating draft. Ratios are calculated per ship before averaging. Feasible populations differ between arms; SD describes variation across fit means.
\end{minipage}
\end{table}

RFR performance guidance accompanies higher mean draft and deadweight in every architecture. Mean steel weight falls for DDPM and SCDDPM and rises for HCDDPM; mean outfit weight falls slightly for DDPM and SCDDPM and rises for HCDDPM. These are comparisons of different feasible generated populations, not controlled changes to individual vessels. Across the six configurations, the fit-averaged block coefficients are 0.891--0.908 and steel masses are 14.7--19.9 thousand tonnes. Their engineering interpretation should retain the synthetic, repair-shaped training population and the evaluator's preliminary mass and geometry assumptions. Appendix~\ref{app_K} compares selected generated particulars with published bulk carriers.

\section{Fixed-Range Supporting Results}
\label{app_J}
This appendix reports fixed-range guidance sweeps, paired qualifying-yield contrasts, and the cost--cargo tradeoff analysis.

\subsection{Four-Strength Performance Sweep}
Separate sets of ship designs are sampled with the required voyage range fixed at 1,000, 5,000, or 8,000 nautical miles. The selection sweep requests 1,000 proposals per architecture, required voyage range, and strength in ten independently seeded 100-request batches. Within each architecture and required voyage range, all four strengths share initial noise and corresponding reverse innovations. The primary comparisons use paired batch values of $Y(c)$ against $\gamma=0$, with Holm adjustment over 27 tests. Feasible RFR is pooled within each cell rather than averaged across batch means. The reverse process includes $\mathbf z_1\to\mathbf z_0$ and guides at $t=800,\ldots,2$ (code indices $k=799,\ldots,1$); architecture-specific feasibility-guidance contexts and RFR regression models are as described in the main text. The selection sweep, single-fit feasibility-guidance ablation, and five-fit confirmation remain separate cohorts.

At common $\gamma=0.10$, SCDDPM has 57.8--60.3\% qualifying yield across the three ranges, 15.8--18.0\% higher than DDPM and 18.3--36.0\% higher than HCDDPM. HCDDPM has the lowest mean feasible RFR at that common strength but retains substantially more feasibility at $\gamma=0.03$. These sweep observations motivate separate confirmation of the selected settings and do not establish architecture superiority across training randomness or ranges.

Tables~\ref{tab_J1} and~\ref{tab_J2} report the 36 configurations and all 27 planned comparisons. Each cell has ten paired 100-request batches. Each set of ships is assessed against cost targets derived from fully feasible training designs within the corresponding voyage-range window, with no selection using generated outcomes. Table~\ref{tab_J_reference_values} lists these targets. The lower-quartile value defines the primary low-cost target for ships generated for that voyage-range requirement. The median and upper-quartile values provide less restrictive targets for the sensitivity analysis.

\begin{table}[H]
\caption[RFR targets for each required voyage range]{RFR targets for each fixed voyage-range sampling requirement, in pounds per tonne of cargo transported (\pounds/t). Targets are derived from fully feasible training designs in the corresponding voyage-range window.}
\label{tab_J_reference_values}
\centering\small
\begin{arxivtablefit}
\begin{tabular}{rrrr}
\toprule
\shortstack{Required voyage\\range (nmi)} & \shortstack{Lower-quartile\\RFR} & \shortstack{Median\\RFR} & \shortstack{Upper-quartile\\RFR}\\
\midrule
1,000 & 9.6246 & 13.2215 & 19.3101\\
5,000 & 24.5577 & 33.3885 & 52.4106\\
8,000 & 33.4776 & 46.7062 & 77.7864\\
\bottomrule
\end{tabular}
\end{arxivtablefit}
\end{table}

\begingroup\small
\begin{arxivlongtable}
\begin{longtable}{llrrrrrrr}
\caption[Fixed-range strength sweep. Each row requests 1,000 designs. Feasible and qualifying yields use all requests; mean RFR (\pounds/t) conditions on full feasibility. The reference columns report qualifying yields at the primary, median, and upper-quartile RFR reference values derived from feasible training designs.]{Fixed-range strength sweep. Each row requests 1,000 designs. Feasible and qualifying yields use all requests; mean RFR (\pounds/t) conditions on full feasibility. The reference columns report qualifying yields at the primary, median, and upper-quartile RFR reference values derived from feasible training designs. Within each architecture and voyage range, bold marks the highest yields and lowest RFR across the four guidance strengths, including ties.}\label{tab_J1}\\
\toprule
Architecture & Range (nmi) & $\gamma$ & Completed & Feasible (\%) & RFR & \shortstack{Primary\\reference\\yield (\%)} & \shortstack{Median\\reference\\yield (\%)} & \shortstack{Upper-quartile\\reference\\yield (\%)}\\
\midrule
\endfirsthead
\toprule
Architecture & Range (nmi) & $\gamma$ & Completed & Feasible (\%) & RFR & \shortstack{Primary\\reference\\yield (\%)} & \shortstack{Median\\reference\\yield (\%)} & \shortstack{Upper-quartile\\reference\\yield (\%)}\\
\midrule
\endhead
DDPM & 1,000 & 0.00 & 1000 & 56.9 & 9.18 & 40.2 & 55.6 & 56.8\\
DDPM & 1,000 & 0.01 & 1000 & \textbf{57.3} & 8.88 & 44.2 & \textbf{56.7} & \textbf{57.3}\\
DDPM & 1,000 & 0.03 & 999 & 55.8 & 8.43 & 47.9 & 55.3 & 55.8\\
DDPM & 1,000 & 0.10 & 1000 & 51.8 & \textbf{7.29} & \textbf{51.1} & 51.8 & 51.8\\
DDPM & 5,000 & 0.00 & 999 & 52.2 & 24.07 & 33.7 & 49.1 & 52.0\\
DDPM & 5,000 & 0.01 & 1000 & \textbf{54.1} & 23.20 & 37.9 & 52.3 & \textbf{54.0}\\
DDPM & 5,000 & 0.03 & 1000 & 53.8 & 21.48 & 45.1 & \textbf{53.4} & 53.8\\
DDPM & 5,000 & 0.10 & 1000 & 50.7 & \textbf{17.84} & \textbf{50.4} & 50.7 & 50.7\\
DDPM & 8,000 & 0.00 & 1000 & \textbf{57.7} & 38.13 & 21.2 & 50.2 & \textbf{56.9}\\
DDPM & 8,000 & 0.01 & 1000 & 56.7 & 36.07 & 25.6 & 51.6 & 56.2\\
DDPM & 8,000 & 0.03 & 1000 & 56.1 & 32.95 & 34.8 & \textbf{54.0} & 56.0\\
DDPM & 8,000 & 0.10 & 1000 & 52.7 & \textbf{26.56} & \textbf{49.9} & 52.7 & 52.7\\
HCDDPM & 1,000 & 0.00 & 997 & 55.9 & 11.61 & 10.8 & 44.1 & 55.2\\
HCDDPM & 1,000 & 0.01 & 998 & \textbf{58.5} & 10.77 & 20.6 & 52.3 & \textbf{57.8}\\
HCDDPM & 1,000 & 0.03 & 998 & 56.0 & 9.39 & 36.7 & \textbf{54.6} & 55.7\\
HCDDPM & 1,000 & 0.10 & 999 & 45.6 & \textbf{7.13} & \textbf{45.3} & 45.5 & 45.6\\
HCDDPM & 5,000 & 0.00 & 999 & 58.4 & 31.08 & 10.5 & 41.2 & 57.3\\
HCDDPM & 5,000 & 0.01 & 997 & 58.2 & 27.20 & 22.4 & 50.5 & 58.0\\
HCDDPM & 5,000 & 0.03 & 995 & \textbf{60.9} & 21.83 & \textbf{49.9} & \textbf{60.1} & \textbf{60.9}\\
HCDDPM & 5,000 & 0.10 & 1000 & 49.8 & \textbf{15.22} & 49.8 & 49.8 & 49.8\\
HCDDPM & 8,000 & 0.00 & 999 & \textbf{59.8} & 45.79 & 4.5 & 38.9 & 58.3\\
HCDDPM & 8,000 & 0.01 & 998 & 58.1 & 39.85 & 13.1 & 47.2 & 57.6\\
HCDDPM & 8,000 & 0.03 & 998 & 59.1 & 31.95 & 41.9 & \textbf{57.7} & \textbf{59.1}\\
HCDDPM & 8,000 & 0.10 & 995 & 42.5 & \textbf{21.50} & \textbf{42.5} & 42.5 & 42.5\\
SCDDPM & 1,000 & 0.00 & 995 & 61.3 & 10.51 & 24.3 & 56.3 & 61.2\\
SCDDPM & 1,000 & 0.01 & 997 & 61.7 & 10.08 & 30.3 & 57.8 & 61.7\\
SCDDPM & 1,000 & 0.03 & 999 & \textbf{62.7} & 9.45 & 40.3 & 61.1 & \textbf{62.7}\\
SCDDPM & 1,000 & 0.10 & 999 & 62.6 & \textbf{8.11} & \textbf{60.3} & \textbf{62.6} & 62.6\\
SCDDPM & 5,000 & 0.00 & 997 & 60.5 & 26.67 & 26.3 & 52.8 & 60.2\\
SCDDPM & 5,000 & 0.01 & 998 & 59.9 & 25.05 & 32.9 & 55.4 & 59.7\\
SCDDPM & 5,000 & 0.03 & 998 & \textbf{60.9} & 22.74 & 45.4 & \textbf{59.4} & \textbf{60.9}\\
SCDDPM & 5,000 & 0.10 & 1000 & 59.4 & \textbf{18.74} & \textbf{58.9} & \textbf{59.4} & 59.4\\
SCDDPM & 8,000 & 0.00 & 998 & \textbf{62.8} & 40.09 & 17.5 & 50.7 & 62.1\\
SCDDPM & 8,000 & 0.01 & 999 & 62.7 & 37.70 & 24.6 & 54.4 & 61.9\\
SCDDPM & 8,000 & 0.03 & 999 & 62.5 & 33.86 & 35.2 & 59.6 & \textbf{62.3}\\
SCDDPM & 8,000 & 0.10 & 999 & 60.4 & \textbf{27.40} & \textbf{57.8} & \textbf{60.3} & 60.4\\
\bottomrule
\end{longtable}
\end{arxivlongtable}
\endgroup
\begingroup\small
\begin{arxivlongtable}
\begin{longtable}{llrrrr}
\caption[Fixed-range qualifying-yield contrasts at the RFR reference value against $\gamma=0$, in percentage points. Unadjusted 95\% paired $t$ intervals use ten matched batches. Holm correction covers all 27 planned comparisons.]{Fixed-range qualifying-yield contrasts at the RFR reference value against $\gamma=0$, in percentage points. Unadjusted 95\% paired $t$ intervals use ten matched batches. Holm correction covers all 27 planned comparisons. Within each architecture and voyage range, bold marks the largest observed gain across guidance strengths; p-values are unranked.}\label{tab_J2}\\
\toprule
Architecture & Range (nmi) & $\gamma$ & Gain (pp) & 95\% interval (pp) & $p_{\mathrm{Holm}}$\\
\midrule
\endfirsthead
\toprule
Architecture & Range (nmi) & $\gamma$ & Gain (pp) & 95\% interval (pp) & $p_{\mathrm{Holm}}$\\
\midrule
\endhead
DDPM & 1,000 & 0.01 & 4.0 & [2.18, 5.82] & 0.00151\\
DDPM & 1,000 & 0.03 & 7.7 & [5.82, 9.58] & $7.13\times10^{-5}$\\
DDPM & 1,000 & 0.10 & \textbf{10.9} & [8.60, 13.20] & $2.58\times10^{-5}$\\
DDPM & 5,000 & 0.01 & 4.2 & [2.63, 5.77] & 0.000777\\
DDPM & 5,000 & 0.03 & 11.4 & [9.06, 13.74] & $2.23\times10^{-5}$\\
DDPM & 5,000 & 0.10 & \textbf{16.7} & [8.64, 24.76] & 0.00151\\
DDPM & 8,000 & 0.01 & 4.4 & [2.96, 5.84] & 0.000484\\
DDPM & 8,000 & 0.03 & 13.6 & [11.63, 15.57] & $1.36\times10^{-6}$\\
DDPM & 8,000 & 0.10 & \textbf{28.7} & [25.88, 31.52] & $6.59\times10^{-8}$\\
HCDDPM & 1,000 & 0.01 & 9.8 & [6.36, 13.24] & 0.000589\\
HCDDPM & 1,000 & 0.03 & 25.9 & [23.24, 28.56] & $9.05\times10^{-8}$\\
HCDDPM & 1,000 & 0.10 & \textbf{34.5} & [30.71, 38.29] & $1.54\times10^{-7}$\\
HCDDPM & 5,000 & 0.01 & 11.9 & [8.73, 15.07] & 0.000123\\
HCDDPM & 5,000 & 0.03 & \textbf{39.4} & [35.41, 43.39] & $8.25\times10^{-8}$\\
HCDDPM & 5,000 & 0.10 & 39.3 & [34.94, 43.66] & $1.6\times10^{-7}$\\
HCDDPM & 8,000 & 0.01 & 8.6 & [6.57, 10.63] & $6.09\times10^{-5}$\\
HCDDPM & 8,000 & 0.03 & 37.4 & [32.56, 42.24] & $5.66\times10^{-7}$\\
HCDDPM & 8,000 & 0.10 & \textbf{38.0} & [35.57, 40.43] & $1.55\times10^{-9}$\\
SCDDPM & 1,000 & 0.01 & 6.0 & [3.95, 8.05] & 0.000584\\
SCDDPM & 1,000 & 0.03 & 16.0 & [12.11, 19.89] & $7.13\times10^{-5}$\\
SCDDPM & 1,000 & 0.10 & \textbf{36.0} & [32.82, 39.18] & $2.65\times10^{-8}$\\
SCDDPM & 5,000 & 0.01 & 6.6 & [3.88, 9.32] & 0.00117\\
SCDDPM & 5,000 & 0.03 & 19.1 & [15.86, 22.34] & $4.71\times10^{-6}$\\
SCDDPM & 5,000 & 0.10 & \textbf{32.6} & [27.35, 37.85] & $3.18\times10^{-6}$\\
SCDDPM & 8,000 & 0.01 & 7.1 & [5.21, 8.99] & 0.000123\\
SCDDPM & 8,000 & 0.03 & 17.7 & [15.34, 20.06] & $6.98\times10^{-7}$\\
SCDDPM & 8,000 & 0.10 & \textbf{40.3} & [35.66, 44.94] & $2.11\times10^{-7}$\\
\bottomrule
\end{longtable}
\end{arxivlongtable}
\endgroup

\subsection{Two-Arm Fixed-Range Study}
The two-arm fixed-range experiment submitted 18,000 requests and compared feasibility-only sampling against one performance setting per architecture. Separate ship populations were sampled for required voyage ranges fixed at 1,000, 5,000, or 8,000 nmi. It completed 17,978 evaluations, of which 10,145 were fully feasible. DDPM used RFR strength 0.03, while HCDDPM and SCDDPM used 0.10. The sampler uses 1,000 steps, with guidance active at $t=800,\ldots,2$ (schedule indices $k=799,\ldots,1$). Table~\ref{tab_J3} reports the outcomes of this comparison separately from the four-strength sweep. Appendix~\ref{app_K} uses DDPM, HCDDPM, and SCDDPM designs from this comparison.

\begin{table}[H]
\caption[Fixed-range freight rate and feasible yield]{Two-arm fixed-range RFR study. Every arm requests 1,000 designs. Each RFR entry reports the mean across all fully feasible designs in that sampling arm, followed by $\pm$ the sample standard deviation of the ten batch mean RFRs, in \pounds/t. The standard deviation describes variation between batches. Feasibility uses all requests. Paired comparisons of guided and baseline batch mean RFRs give Holm-adjusted $p<10^{-8}$ for all nine combinations of model and required voyage range. The DDPM performance strength is 0.03, while HCDDPM and SCDDPM use 0.10. Within each required voyage range, bold marks the lowest mean baseline and guided RFR, the largest RFR reduction, and the highest feasibility rate in each arm across architectures.}
\label{tab_J3}
\centering\small
\begin{arxivtablefit}
\begin{tabular}{llrrrr}
\toprule
Architecture & Range (nmi) & Baseline RFR & Guided RFR & $\Delta$RFR (\%) & Feasible (\%) base$\rightarrow$guided\\
\midrule
DDPM & 1,000 & $\mathbf{9.30\pm0.15}$ & $8.47\pm0.13$ & $-8.9$ & $57.2\rightarrow53.7$\\
DDPM & 5,000 & $\mathbf{24.22\pm0.47}$ & $21.46\pm0.44$ & $-11.4$ & $58.9\rightarrow57.2$\\
DDPM & 8,000 & $\mathbf{36.48\pm0.88}$ & $32.34\pm0.94$ & $-11.4$ & $53.6\rightarrow55.4$\\
HCDDPM & 1,000 & $11.87\pm0.35$ & $\mathbf{7.16\pm0.10}$ & $\mathbf{-39.7}$ & $58.7\rightarrow45.3$\\
HCDDPM & 5,000 & $31.21\pm1.36$ & $\mathbf{15.36\pm0.13}$ & $\mathbf{-50.8}$ & $59.6\rightarrow52.1$\\
HCDDPM & 8,000 & $44.88\pm1.19$ & $\mathbf{21.35\pm0.16}$ & $\mathbf{-52.4}$ & $58.2\rightarrow42.3$\\
SCDDPM & 1,000 & $10.70\pm0.33$ & $8.16\pm0.11$ & $-23.7$ & $\mathbf{61.8}\rightarrow\mathbf{61.2}$\\
SCDDPM & 5,000 & $26.42\pm1.15$ & $18.81\pm0.24$ & $-28.8$ & $\mathbf{61.2}\rightarrow\mathbf{59.3}$\\
SCDDPM & 8,000 & $39.60\pm0.87$ & $27.28\pm0.32$ & $-31.1$ & $\mathbf{60.0}\rightarrow\mathbf{58.8}$\\
\bottomrule
\end{tabular}
\end{arxivtablefit}
\end{table}

Equation~\eqref{eq_J1} defines the economic coverage $\mathcal C(A,B)$ among fully feasible designs: the fraction of set $B$ dominated by at least one member of set $A$ when annual total cost is minimized and annual cargo transported is maximized:
\begin{equation}
\begin{aligned}
\mathcal C(A,B)&=\frac{1}{|B|}\sum_{b\in B}
\mathbf 1\!\left[\exists a\in A:\right.\\
&\quad C_{yr}(a)\leq C_{yr}(b),\\
&\quad \left.Q_{yr}(a)\geq Q_{yr}(b),\
\text{at least one strict}\right].
\end{aligned}
\label{eq_J1}
\end{equation}
Because RFR is $C_{yr}/Q_{yr}$, these two outcomes expose the cost--throughput tradeoff behind the ratio. For each model and required voyage range, the analysis retains only completed, fully feasible designs with finite outcomes. The comparison includes ten baseline batches and ten performance-guided batches, each with 100 design requests. The same number of feasible designs is sampled from every batch, using the smallest feasible-design count among those 20 batches. This number ranges from 32 to 53 across the model and required voyage range combinations. One hundred repeated subsamples stabilize each matched batch pair's two directional coverage estimates; the ten independent batch pairs, not the redraws, provide the reported SD and paired comparison. Holm adjustment covers the nine within-architecture/range net-coverage tests.

Table~\ref{tab_J4} shows that RFR performance guidance improves the cost--cargo tradeoff among feasible designs across all three architectures and voyage ranges: guided sets cover a larger share of the baseline sets than the reverse. Guided-to-baseline coverage is 0.920--0.941 for DDPM, 0.894--0.952 for HCDDPM, and 0.967--0.976 for SCDDPM; reverse coverage is 0.553--0.587, zero, and 0.150--0.179, respectively. All nine net differences have adjusted $p<1.4\times10^{-7}$ across the ten batches. HCDDPM's zero reverse coverage occurs in these equal-sized within-batch comparisons, where its guided designs move strongly toward larger annual cargo even when annual cost rises. Coverage summarizes the feasible sets within each architecture and range; qualifying yield per request is evaluated separately in the five-fit confirmation.

\begin{table}[H]
\caption[Cost--cargo Pareto coverage]{Cost--cargo Pareto coverage. $G$ denotes performance-guided feasible designs and $B$ the feasibility-only baseline. Equal subset size $n$ is fixed for each model and required voyage range. Entries are mean $\pm$ SD over ten matched batch pairs after 100 equal-size redraws per pair. A design outperforms another when it has no higher annual cost and no lower annual cargo transported, with at least one strict improvement. For each model and required voyage range, the test compares the fraction of baseline designs outperformed by at least one guided design against the fraction of guided designs outperformed by at least one baseline design. Holm-adjusted tests evaluate the difference between these fractions. Within each row, bold marks the larger of the two directional coverage means; bold does not rank architectures.}
\label{tab_J4}
\centering\small
\begin{arxivtablefit}
\begin{tabular}{llrrrr}
\toprule
Architecture & Range (nmi) & $n$ & $\mathcal C(G,B)$ & $\mathcal C(B,G)$ & $p_{\mathrm{Holm}}$\\
\midrule
DDPM & 1,000 & 48 & $\mathbf{0.941\pm0.014}$ & $0.587\pm0.074$ & $1.31\times10^{-7}$\\
DDPM & 5,000 & 45 & $\mathbf{0.938\pm0.025}$ & $0.572\pm0.056$ & $1\times10^{-7}$\\
DDPM & 8,000 & 43 & $\mathbf{0.920\pm0.028}$ & $0.553\pm0.058$ & $1.31\times10^{-7}$\\
HCDDPM & 1,000 & 33 & $\mathbf{0.952\pm0.030}$ & $0.000\pm0.000$ & $4.24\times10^{-14}$\\
HCDDPM & 5,000 & 45 & $\mathbf{0.942\pm0.042}$ & $0.000\pm0.000$ & $8.06\times10^{-13}$\\
HCDDPM & 8,000 & 32 & $\mathbf{0.894\pm0.058}$ & $0.000\pm0.000$ & $2.15\times10^{-11}$\\
SCDDPM & 1,000 & 53 & $\mathbf{0.976\pm0.020}$ & $0.175\pm0.067$ & $2.88\times10^{-10}$\\
SCDDPM & 5,000 & 50 & $\mathbf{0.970\pm0.027}$ & $0.179\pm0.080$ & $2.95\times10^{-9}$\\
SCDDPM & 8,000 & 53 & $\mathbf{0.967\pm0.023}$ & $0.150\pm0.077$ & $2.38\times10^{-9}$\\
\bottomrule
\end{tabular}
\end{arxivtablefit}
\end{table}

\section{Dimensions of Feasible Designs With Low Required Freight Rates}
\label{app_K}
This appendix compares the dimensions of low-RFR generations with other feasible samples and selected published bulk carriers.

The two-arm fixed-range study contains 18,000 proposals and 10,145 completed designs satisfying the formulated engineering constraints. Within each prescribed voyage range, the lowest-RFR group contains the first $\lceil0.10n\rceil$ feasible designs when ranked by exact-evaluator RFR; the remaining feasible designs form the comparison group. Table~\ref{tab_K1} reports medians of evaluator dimensions and design-level ratios. Ratios are computed for each design before taking their medians, so a median ratio need not equal the ratio of the two displayed dimension medians. $L$ is length overall, $B$ is beam, $D$ is depth, and $T$ is draft.

\begin{table}[!h]
\caption[Dimensions in low-RFR feasible generations]{Post hoc comparison of the lowest 10\% of exact-evaluator RFR values within each fixed voyage range against the other fully feasible generated designs. Each range pools DDPM, HCDDPM, and SCDDPM, with feasibility-only and performance-guided arms. All entries are medians except sample counts. These groups are selected by RFR and should not be interpreted as a controlled effect of any one hull dimension.}
\label{tab_K1}
\centering\small
\setlength{\tabcolsep}{4pt}
\begin{arxivtablefit}
\begin{tabular}{llrrrrrr}
\toprule
Range (nmi) & Group ($n$) & $L/B$ & $L$ (m) & $B$ (m) & $D$ (m) & $T$ (m) & $B/T$\\
\midrule
1,000 & Other feasible (3,041) & 6.46 & 175.7 & 27.0 & 15.4 & 8.20 & 3.27\\
      & Low RFR (338) & 6.36 & 168.8 & 26.5 & 15.6 & 9.84 & 2.69\\
5,000 & Other feasible (3,134) & 6.47 & 175.3 & 26.9 & 15.4 & 8.20 & 3.29\\
      & Low RFR (349) & 6.32 & 180.1 & 28.2 & 16.7 & 10.79 & 2.61\\
8,000 & Other feasible (2,954) & 6.46 & 174.4 & 27.0 & 15.4 & 8.07 & 3.34\\
      & Low RFR (329) & 6.34 & 180.3 & 28.4 & 16.9 & 10.89 & 2.59\\
\bottomrule
\end{tabular}
\end{arxivtablefit}
\end{table}

The pooled low-RFR deciles contain 251/338 (74\%), 336/349 (96\%), and 317/329 (96\%) performance-guided HCDDPM designs at the three ranges. To check this composition effect, the bottom decile was also formed separately within each architecture/range/arm cell. Compared with the remaining feasible designs in the same cell, median draft is higher in 17 of 18 cells, beam-over-draft is lower in 18 of 18, cargo per voyage is higher in 17 of 18, and annual cargo is higher in 16 of 18. Length-over-beam is lower in only 11 of 18. The observed geometry therefore suggests an association between low RFR and deeper, cargo-carrying designs in this model; it does not identify a preferred slenderness or a causal hull-design rule. Speed and operating economics also enter RFR, and these feasible-only comparisons do not account for the different feasible yields of the sampling arms.

\subsection{Illustrative Comparison With Published Bulk Carriers}
Publicly available bulk-carrier particulars permit comparisons of principal dimensions, displacement, deadweight, block coefficient, and operating speed. Table~\ref{tab_K2} compares selected low-RFR generations with four built bulk carriers to identify similarities and differences in these characteristics. The published comparators span 8,009--82,108 DWT. For each required voyage range, fully feasible designs sampled for that requirement are pooled across all three architectures and both sampling arms. The 10\% with the lowest cost per tonne transported are selected first. The illustrative vessel is then chosen as the design closest to that group's median RFR. These are individual designs, separate from the five-fit means in Appendix~\ref{app_I}.

\begin{table}[!h]
\caption[Generated and published bulk-carrier particulars]{Principal particulars of three selected low-RFR feasible generations and four published bulk carriers. Generated values are model estimates at full-cargo departure; published dimensions and masses use summer loading. Speeds are model-selected versus operator-reported service values, except the published full-laden speed for Cemix.}
\label{tab_K2}
\centering\footnotesize
\setlength{\tabcolsep}{2.5pt}
\begin{arxivtablefit}
\begin{tabular}{lrrrrrrrrr}
\toprule
Design or vessel & $L$ (m) & $B$ (m) & $T$ (m) & $D$ (m) & DWT (t) & Disp. (t) & Speed (kn) & $C_{B,\mathrm{OA}}$ & DWT/$\Delta$\\
\midrule
GenISys, 1,000 nmi & 162.9 & 24.5 & 8.97 & 13.73 & 19,571 & 33,381 & 10.61 & 0.911 & 0.586\\
GenISys, 5,000 nmi & 190.6 & 29.3 & 10.86 & 16.82 & 33,534 & 56,892 & 11.23 & 0.914 & 0.589\\
GenISys, 8,000 nmi & 179.6 & 25.9 & 10.06 & 15.59 & 26,235 & 42,607 & 10.54 & 0.888 & 0.616\\
\midrule
\emph{Cemix} & 112.8 & 17.2 & 7.28 & 9.10 & 8,009 & 10,622 & 10.50 & 0.734 & 0.754\\
\emph{Aeriko} & 200.0 & 32.3 & 13.30 & 18.50 & 63,351 & 74,922 & 13.00 & 0.852 & 0.846\\
\emph{Anna Smile} & 225.0 & 32.3 & 14.25 & 19.60 & 74,823 & 87,337 & 13.00 & 0.824 & 0.857\\
\emph{Diamanti} & 229.0 & 32.3 & 14.43 & 20.05 & 82,108 & 94,350 & 14.50 & 0.864 & 0.870\\
\bottomrule
\end{tabular}
\end{arxivtablefit}
\end{table}

The operator's \emph{Aeriko} fleet page gives LOA 199.99 m, summer draft 13.300 m, DWT 63,351 t, and lightship mass 11,571 t \cite{prosperity2026aeriko}. The deadweight and lightship masses give a derived summer displacement of 74,922 t. For \emph{Cemix}, \emph{Anna Smile}, and \emph{Diamanti}, summer displacement is explicitly derived as published DWT plus lightship \cite{aykop2026cemix,prosperity2026anna,prosperity2026diamanti}. Speeds are the operators' approximate service values, except \emph{Cemix}'s published 10.5-kn full-laden speed; they are not matched design-speed trials.

Every row uses $C_{B,\mathrm{OA}}=\Delta/(1.025L_{\mathrm{OA}}BT)$, with mass in tonnes and dimensions in metres, matching the evaluator's definition of block coefficient. Block coefficients and DWT/displacement ratios are calculated from unrounded values. The generated lengths, beams, depths, drafts, and speeds lie within the span of the four published vessels. The generated examples have higher block coefficients (0.888--0.914 versus 0.734--0.864) and lower deadweight-to-displacement ratios (0.586--0.616 versus 0.754--0.870). The deadweight-to-displacement ratio gives the share of total ship mass available for cargo, fuel, and other deadweight. The lower ratios therefore indicate a larger lightship share. GenISys uses a representative structural model and repair-derived designs \cite{bagazinski2026midship}, and its operating drafts differ from the published summer-loading conditions. 

The published particulars used here do not provide the detailed structural arrangements, loading information, and measured performance needed to assess the generated structural arrangements or the predictive accuracy of the complete evaluator. The comparison therefore identifies similarities and differences at the level supported by the available data. Further comparisons will require richer reference data, higher-fidelity design representations, improved cost models, and additional engineering and economic metrics.

\section{Feasibility-Restoring Hopper Subsystem Redesign}
\label{app_L}
This appendix describes the Hopper-only redesign experiment and reports feasible-child yield and starting-design coverage.

A starting design is an existing ship selected for subsystem redesign. This differs from a structural parent, the starting set of structural parameters shared by dataset variants, and from a repair source, the evaluated design from which repair alternatives are generated. The Hopper experiment uses 100 held-out test designs as starting points. Each fails the Hopper checks but passes the other 19 subsystem labels and the system-level checks. Only 18 Hopper controls may change; the other 115 independent controls, including all representative hatch controls, remain locked. Each architecture requests ten unguided and ten feasibility-guided children per starting design at strength 0.7, giving 6,000 requests; all completed evaluation. Success requires the lock audit and whole-system counted feasibility. Starting-design-level ten-attempt fractions are paired across arms, and Holm correction covers three within-architecture guidance contrasts plus guided SCDDPM versus each guided baseline. Table~\ref{tab_L1} shows child yields, guidance gains, and starting-design coverage.

\begin{table}[H]
\caption[Hopper-only feasibility restoration]{Hopper-only feasibility restoration. Yield uses all 1,000 requested children per architecture/arm. Guidance gains and unadjusted 95\% intervals use 100 paired parent fractions. Holm correction covers the five planned contrasts. Starting-design counts give at least one fully feasible child among ten attempts and are descriptive. Bold marks the highest feasible yield in each arm and largest observed guidance gain across architectures; parent counts and p-values are unranked.}
\label{tab_L1}
\centering\small
\begin{arxivtablefit}
\begin{tabular}{llrrrr}
\toprule
Architecture & Feasible yield (\%) & Gain (pp) & 95\% interval (pp) & $p_{\mathrm{Holm}}$ & Parents reached\\
\midrule
DDPM & $10.6\rightarrow17.7$ & 7.1 & [3.49, 10.71] & 0.000708 & $38\rightarrow38$\\
HCDDPM & $\mathbf{17.4}\rightarrow\mathbf{25.3}$ & 7.9 & [3.11, 12.69] & 0.00439 & $51\rightarrow38$\\
SCDDPM & $15.4\rightarrow23.9$ & \textbf{8.5} & [5.18, 11.82] & 8.76e-06 & $51\rightarrow55$\\
\bottomrule
\end{tabular}
\end{arxivtablefit}
\end{table}

Guided SCDDPM exceeds guided DDPM by 6.2 percentage points (unadjusted interval 1.66--10.74; $p_{\mathrm{Holm}}=0.0158$), but its difference from HCDDPM is $-1.4$ points (interval $-7.77$--4.97; $p_{\mathrm{Holm}}=0.664$). HCDDPM's higher guided attempt yield accompanies a fall in starting designs reached from 51 to 38; SCDDPM reaches 55. Attempt yield is the fraction of all 1,000 redesign requests that produce fully feasible ships while preserving the locked parameters. The number of successful starting ships counts how many of the 100 starting designs produce at least one successful redesign in ten attempts. A higher attempt yield can therefore accompany fewer starting ships with successful redesigns. This task does not validate a general algebraic repair algorithm.

\section{Performance-Guided Subsystem Redesign}
\label{app_M}
This appendix presents the exploratory and independent confirmation studies of performance-guided subsystem redesign.

\subsection{Exploratory Production Matrix}
The original objective-redesign matrix uses the same 100 generator-training-held-out, fully feasible starting designs for every architecture and target, with ten attempts per starting design per arm. Only the 14 bow/stern bulb controls, 11 bulkhead-structure controls, or seven longitudinal-bottom-structure controls may change. The bulb objective is total resistance at the starting design's locked operating point; each structural objective is total steel weight. Every other independent control is restored exactly before evaluation, although dependent reconstructed values may change. DDPM resamples selected coordinates in its full-system denoiser; HCDDPM resamples a block using predecessor context, processing bow and stern bulbs in sequence; SCDDPM updates the selected block or both bulb blocks using the other blocks as context. No downstream HCDDPM blocks are regenerated.

Feasibility guidance alone is compared with feasibility plus a selected single-objective gradient. The 480 pilot children were excluded from production; the pilot tested performance strengths 0.03 and 0.10 for bulbs and bulkheads, and the selected steel setting was also applied to longitudinal bottom without a separate bottom-specific pilot. Pilot and production parent selection nevertheless overlapped for 13 of the 100 production starting designs. Each 1,000-attempt cell is judged primarily by the fraction of all attempts that are both fully feasible and better than their starting design. The 100 starting-design-level ten-attempt fractions are the units for paired $t$ tests, with Holm adjustment over nine architecture/target comparisons; a sensitivity analysis excluding the 13 shared starting designs uses the remaining 87 parent units and the same adjustment.

The exploratory matrix completed 17,818 of 18,000 evaluations, with 14,719 fully feasible attempts. Performance guidance raises the observed feasible-and-improved fraction in all nine architecture/target cells (Table~\ref{tab_M1}). In the exploratory analysis of 100 starting designs, added performance guidance produced gains with Holm-adjusted $p<0.05$ in six cases: bulb redesign with DDPM, HCDDPM, and SCDDPM; bulkhead redesign with HCDDPM; and longitudinal-bottom redesign with DDPM and HCDDPM. Gains for DDPM bulkhead redesign and SCDDPM bulkhead and longitudinal-bottom redesign do not meet this adjusted significance threshold. Excluding the 13 starting designs shared with the pilot leaves five adjusted-significant gains; HCDDPM longitudinal bottom changes from $+11.0$ percentage points in all starting designs to $+8.6$ points among 87 never-used-in-pilot starting designs, with adjusted $p=0.117$. No SCDDPM-versus-DDPM or SCDDPM-versus-HCDDPM difference in joint-success rate remains significant after adjustment over the 12 matched target/arm comparisons. The reported mean reductions in resistance or steel weight condition on completed, feasible children and cannot be applied to failures or infeasible attempts. The joint-success measure retains those attempts in its denominator.

\begin{table}[H]
\caption[Exploratory locked-subsystem redesign outcomes]{Exploratory locked-subsystem performance-guided redesign. Each cell requests 1,000 attempts from the same 100 generator-training-held-out starting designs; arrows show feasibility-only to performance-guided values. ``Joint success'' means fully feasible and better than the starting design, divided by all attempts. The gain is the paired difference across starting-design-level ten-attempt fractions, with an unadjusted 95\% $t$ interval in percentage points (pp). Six gains have Holm-adjusted $p<0.05$ across nine full-cohort tests; a sensitivity analysis excluding 13 starting designs shared with the tuning pilot retains five of these six. The final column is the guided arm's mean objective change from parent among feasible children only; negative is better. Within each target, bold marks the highest feasibility and joint-success rate in each arm, largest observed gain, and lowest guided objective change across architectures; it does not indicate significance.}
\label{tab_M1}
\centering\small
\begin{arxivtablefit}
\begin{tabular}{llrrrr}
\toprule
Target & Architecture & Feasible (\%) & Joint success (\%) & Gain (pp; 95\% CI) & Guided objective change\\
\midrule
Bulbs & DDPM & $86.8\rightarrow88.5$ & $46.5\rightarrow49.5$ & $3.0\ [0.9,5.1]$ & $-41.0$ kN\\
Bulbs & HCDDPM & $\mathbf{87.7}\rightarrow\mathbf{89.1}$ & $\mathbf{50.4}\rightarrow\mathbf{56.0}$ & $\mathbf{5.6\ [3.3,7.9]}$ & $\mathbf{-51.0}$ kN\\
Bulbs & SCDDPM & $85.5\rightarrow88.1$ & $45.2\rightarrow49.7$ & $4.5\ [2.0,7.0]$ & $-43.0$ kN\\
Bulkhead & DDPM & $75.9\rightarrow72.7$ & $34.8\rightarrow35.8$ & $1.0\ [-1.5,3.5]$ & $-409$ t\\
Bulkhead & HCDDPM & $\mathbf{85.2}\rightarrow\mathbf{83.3}$ & $\mathbf{43.1}\rightarrow\mathbf{50.5}$ & $\mathbf{7.4\ [2.7,12.1]}$ & $\mathbf{-447}$ t\\
Bulkhead & SCDDPM & $82.6\rightarrow81.6$ & $38.4\rightarrow39.1$ & $0.7\ [-0.5,1.9]$ & $-159$ t\\
Longitudinal bottom & DDPM & $75.8\rightarrow73.7$ & $42.7\rightarrow47.2$ & $4.5\ [2.1,6.9]$ & $\mathbf{-683}$ t\\
Longitudinal bottom & HCDDPM & $79.4\rightarrow75.8$ & $37.0\rightarrow\mathbf{48.0}$ & $\mathbf{11.0\ [4.0,18.0]}$ & $-556$ t\\
Longitudinal bottom & SCDDPM & $\mathbf{80.4}\rightarrow\mathbf{79.8}$ & $\mathbf{43.6}\rightarrow44.0$ & $0.4\ [-0.7,1.5]$ & $-317$ t\\
\bottomrule
\end{tabular}
\end{arxivtablefit}
\end{table}

\subsection{Independent Starting-Design Confirmation}
Before redesign generation, 58 candidate starting designs from source groups unused in either pilot or production were rebuilt and evaluated. Structural cleanup adjusted bracket endpoints, and three rebuilt designs failed the study's $40^\circ$--$50^\circ$ lower-hopper bracket-angle requirement. These three were excluded before generation, leaving 55 starting designs. Generated designs were compared with the corresponding reevaluated starting design's resistance or steel weight. The training-set labels and RFR reference value retain their recorded definitions. All 55 starting designs that passed reevaluation were used in the confirmation study. The three excluded designs were not replaced. The same three edit masks and objectives apply. Every architecture/target pair has feasibility-only and performance-guided arms with ten attempts per starting design, for 9,900 requested children. Feasibility strength is 0.7 in both arms. Performance-guidance strengths are 0.03 for DDPM on all targets; 0.03 for HCDDPM bulbs and 0.10 for its two structural targets; and 0.10 for SCDDPM bulbs and 0.03 for its two structural targets. These settings were frozen before confirmation. All 9,900 raw-control and decoded-parameter lock audits passed. The exact evaluator completed 9,786 children; failures remain unsuccessful requests. There were 8,345 fully feasible children and 4,082 that were both feasible and better than the starting design's current objective.

Each starting design's success fraction over its ten requested children is paired across arms. Two-sided paired $t$ tests use 55 starting-design differences per architecture/target; Holm adjustment covers the nine planned comparisons. Table~\ref{tab_M2} reports all nine effects. Although every point estimate is positive, only the HCDDPM bulkhead effect survives the adjusted test. The displayed 95\% intervals are unadjusted and should not be read as nine simultaneous intervals. In particular, several intervals exclude zero while their multiplicity-adjusted $p$ values exceed 0.05. Both sampling arms produced fully feasible designs that improved on their starting ships. Adding performance guidance produced a statistically supported increase in this success rate only for HCDDPM bulkhead redesign. The observed increases for the other eight model and redesign-target combinations were not statistically supported after adjustment for the nine comparisons.

\begin{table}[H]
\caption[Independent objective-redesign confirmation]{Independent confirmation on 55 currently feasible, source-distinct starting designs with ten requests per starting design and arm. Joint success requires full counted feasibility and a strictly lower target objective than the starting design's current exact baseline. Rates use all 550 requests in each arm; the paired gain and unadjusted 95\% $t$ interval use starting-design-level ten-attempt fractions. Holm-adjusted $p$ values cover the nine comparisons. Only the HCDDPM bulkhead result meets adjusted $p<0.05$. Within each target, bold marks the highest joint-success rate in each arm and largest observed gain across architectures, including ties; bold does not indicate significance.}
\label{tab_M2}
\centering\small
\begin{arxivtablefit}
\begin{tabular}{llrrr}
\toprule
Target & Architecture & Joint success (\%) base$\rightarrow$guided & Paired gain (pp; 95\% CI) & $p_{\mathrm{Holm}}$\\
\midrule
Bulbs & DDPM & $\mathbf{35.82}\rightarrow37.45$ & $1.64\ [-0.36,3.64]$ & 0.320\\
Bulbs & HCDDPM & $32.91\rightarrow\mathbf{37.64}$ & $\mathbf{4.73\ [0.60,8.85]}$ & 0.168\\
Bulbs & SCDDPM & $34.73\rightarrow36.18$ & $1.45\ [-0.78,3.69]$ & 0.341\\
Bulkhead & DDPM & $34.18\rightarrow37.27$ & $3.09\ [-0.15,6.33]$ & 0.246\\
Bulkhead & HCDDPM & $\mathbf{50.00}\rightarrow\mathbf{59.82}$ & $\mathbf{9.82\ [3.35,16.28]}$ & $0.032$\\
Bulkhead & SCDDPM & $46.36\rightarrow48.18$ & $1.82\ [0.25,3.39]$ & 0.168\\
Longitudinal bottom & DDPM & $\mathbf{42.55}\rightarrow\mathbf{44.91}$ & $2.36\ [-1.05,5.77]$ & 0.341\\
Longitudinal bottom & HCDDPM & $33.82\rightarrow\mathbf{44.91}$ & $\mathbf{11.09\ [1.32,20.86]}$ & 0.168\\
Longitudinal bottom & SCDDPM & $42.18\rightarrow43.27$ & $1.09\ [0.24,1.94]$ & 0.103\\
\bottomrule
\end{tabular}
\end{arxivtablefit}
\end{table}

\clearpage
\bibliographystyle{asmejour}
\bibliography{biblio}

\end{document}